\documentclass[final]{cvpr} % review, rebuttal, final

\usepackage{times}
\usepackage{epsfig}
\usepackage{graphicx}
\usepackage{amsmath}
\usepackage{amssymb}

\usepackage{color}
\usepackage{gensymb}
\usepackage{adjustbox}
\usepackage[sort,nocompress]{cite}
\usepackage{multirow}
\usepackage{caption} 
\usepackage{microtype} 
\usepackage{booktabs}

\usepackage[pagebackref=true,breaklinks=true,colorlinks,bookmarks=false]{hyperref}

\DeclareGraphicsExtensions{.pdf,.jpg,.png}
\graphicspath{{images/}}

\DeclareMathOperator*{\argmin}{arg\,min}

\renewcommand{\figref}[1]{Figure~\ref{fig:#1}}
\newcommand{\tblref}[1]{Table~\ref{tab:#1}}
\newcommand{\ra}[1]{\renewcommand{\arraystretch}{#1}}
\newcommand{\rt}[1]{\textcolor{red}{#1}}

\title{A Structure-Aware Method for Direct Pose Estimation}
\author{
  \centering
  \begin{minipage}{.9\linewidth}
    \centering
    \begin{minipage}{1.5in}
      \centering
      Hunter Blanton$^1$ 
    \end{minipage}
    \begin{minipage}{1.5in}
      \centering
      Scott Workman$^2$ 
    \end{minipage}
    \begin{minipage}{1.5in}
      \centering
      Nathan Jacobs$^1$ 
    \end{minipage}
    \\[.2cm]
    \begin{minipage}{2.0in}
      \centering
      $^1$University of Kentucky
    \end{minipage}
    \begin{minipage}{2.0in}
      \centering
      $^2$DZYNE Technologies
    \end{minipage}
  \end{minipage}
}

\begin{document}

\maketitle

\begin{abstract}

Estimating camera pose from a single image is a fundamental problem in computer vision. Existing methods for solving this task fall into two distinct categories, which we refer to as direct and indirect. Direct methods, such as PoseNet, regress pose from the image as a fixed function, for example using a feed-forward convolutional network. Such methods are desirable because they are deterministic and run in constant time. Indirect methods for pose regression are often non-deterministic, with various external dependencies such as image retrieval and hypothesis sampling. We propose a direct method that takes inspiration from structure-based approaches to incorporate explicit 3D constraints into the network. Our approach maintains the desirable qualities of other direct methods while achieving much lower error in general.
    
\end{abstract}

\section{Introduction}

Camera pose estimation is a fundamental task in computer vision. Often this research is referred to as visual localization, or camera relocalization, and refers to estimating the position and orientation of a camera with respect to some predetermined reference frame. Recently, much work has explored learning-based absolute pose regression which directly regresses camera pose using a single forward pass through a neural network. These methods typically use the same basic pipeline: predict a feature embedding using a convolutional neural network (CNN), and then use features from this embedding to regress the camera pose. This works because the weights of the network implicitly capture understanding of the scene. The main differences between methods of this type are choices of feature extraction architecture or loss function. However, absolute pose regression typically performs much worse than more sophisticated alternatives. A recent study by Sattler et al.~\cite{sattler2019understanding} found that absolute pose regression essentially interpolates between a set of learned basis poses. 

In this work we partition methods into one of two classes, direct or indirect. Direct methods determine pose as a fixed pipeline in a deterministic way with no dependency on external steps such as hypothesis sampling, database querying, pose averaging, correspondence matching, or refinement. These methods are typically a single CNN architecture, such as PoseNet~\cite{posenet}, but we show that it is possible to design a more explicit architecture that still resides in the direct pose estimation category. Indirect methods are any method that fall outside of this definition. Examples include structure-based methods such as Active Search~\cite{activesearch} and DSAC++~\cite{dsac2} which compute 2D-3D correspondences and require RANSAC and refinement for final pose determination, and retrieval-based methods such as DenseVLAD~\cite{densevlad} which require a database query. 

An alternate way of thinking about the difference between direct and indirect methods is that direct methods generate pose hypotheses while indirect methods use one or more pose hypotheses for further processing. We believe this is an important distinction because improvements to direct methods can result in improvements in future indirect methods. While the more accurate methods typically fall into the indirect pose estimation category, we believe direct methods are an exciting area and wish to provide a direction for future improvement. Direct pose estimation methods have many practical benefits. Specifically, they perform relocalization in a way that is fast, deterministic, and provide a constant runtime guarantee regardless of scene complexity. 

We propose an approach for solving the direct pose estimation problem using a CNN that bridges the gap between absolute pose regression and structure-based methods. Our approach has several key components. First, given a single RGB image we simultaneously perform scene coordinate regression and monocular depth estimation to extract geometric information from the image. Then, we use the known camera intrinsics to convert the estimated per-pixel depth to 3D camera-frame coordinates, resulting in a set of 3D-3D correspondences. Our approach takes advantage of known geometric constraints to frame the problem as 3D-3D alignment of point clouds, integrating a differentiable singular value decomposition (SVD) layer with a learned per-pixel weighting scheme to compute the final camera pose. 

We evaluate our method both quantitatively and qualitatively through a variety of experiments using well-known benchmark datasets. Compared to existing direct pose estimation methods, we significantly advance the state-of-the-art and decrease the gap to indirect methods.

\section{Related Work}

In this section we provide an overview of related work in pose regression using CNNs, scene coordinate regression, monocular depth estimation, and point cloud registration.

\subsection{Pose Regression with CNNs}

\begin{figure*}
  \includegraphics[width=\linewidth]{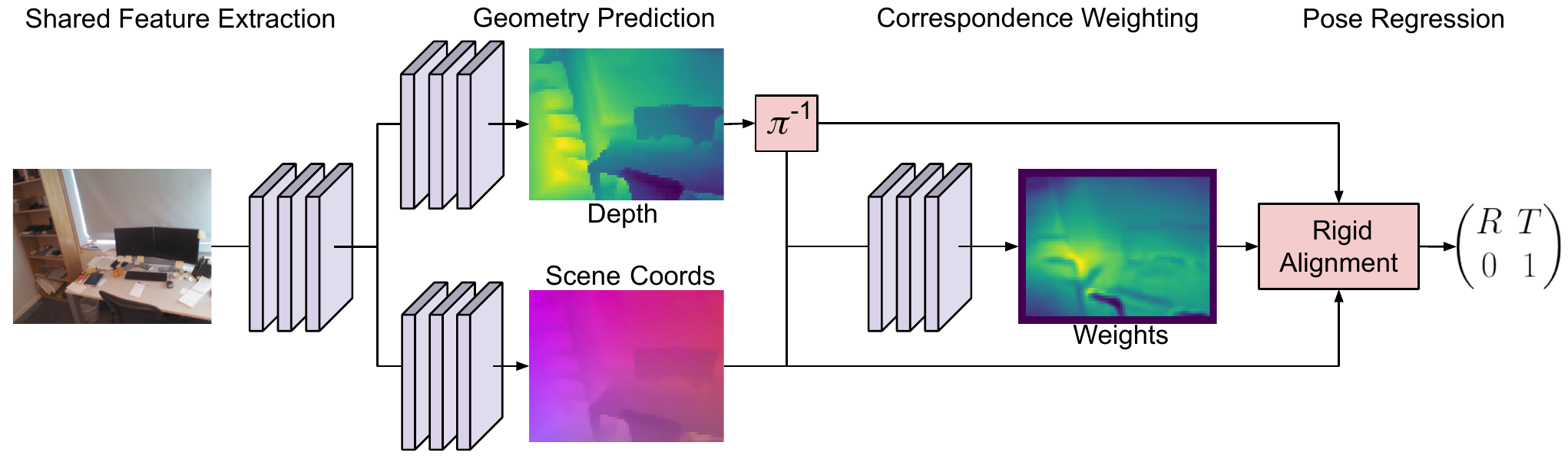}
  \caption{Our direct pose estimation pipeline. Given a single RGB input, we jointly estimate scene coordinates and depth. Using the known camera focal length, depth is converted to camera frame coordinates by unprojecting using $\pi^{-1}$. The second stage of our network computes a correspondence weight for each point pair. These weights are used for point normalization and final pose regression using weighted rigid alignment to estimate camera pose.}
  \label{fig:architecture}
\end{figure*}

Directly estimating camera pose from an image using convolutional neural networks began with the introduction of PoseNet~\cite{posenet}. This style of approach which computes pose as the output of a CNN is known as absolute pose regression. Absolute pose regression differs significantly from traditional methods in that it does not require explicit image-level or pixel-level correspondences.  Many works have explored the application of CNNs to this problem, but differences lie largely in changes to the underlying feature extraction architecture or modified objective functions~\cite{baysposenet,posenet2, hourglasspose,lstmposenet}. Another common approach is to add relative pose constraints on pairs of images during training instead of only absolute pose~\cite{mapnet,localglobal}. In general, this class of methods is attractive due to their simplicity and moderate performance, but are far from the most accurate approaches in terms of pose accuracy. Sattler et al.~\cite{sattler2019understanding} provide an in depth review of existing work in absolute pose regression, with comparisons to more accurate structure-based and image retrieval methods. 

Relative pose estimation approaches predict the pose of a query image relative to a set of database images with known pose. RelocNet~\cite{relocnet} uses a method for camera pose retrieval using nearest neighbors with learned features. Similarly, CamNet~\cite{canmet} uses coarse-to-fine retrieval with two rounds of retrieval and pose refinement. AnchorNet~\cite{anchornet} uses a set of spatial anchors, and the final pose estimate is computed by a weighted average of predicted offsets from the anchors. This method is different from others of this style in that reference poses for relative pose estimation are not based on retrieval, but rather embedded into the network either prior to or during training. The analogue to retrieval in this scenario is a set of anchor point scores, which define the weights for averaging all pose hypotheses. 

\subsection{Scene Coordinate Regression}

Recent work has explored the problem of estimating the 3D location of image pixels in the scene's coordinate frame, known as scene coordinate regression. Unlike camera pose, scene coordinate information is often explicit in the image content, as points in space often look similar from varying viewpoints.  Early work showed that scene coordinates could accurately be computed from RGB-D imagery in small indoor scenes using random forests~\cite{scenecoordforest}. Later, random forests were replaced with CNNs that only require RGB imagery without depth information~\cite{dsac,dsac2,esac}. 

Other methods combine image retrieval and scene coordinate regression. SANet~\cite{sanet} retrieves a set of images to construct a local 3D point cloud that is used for scene coordinate regression. Perspective-n-Learned-Point~\cite{pnlp} uses image retrieval to find a single nearby image for stereo depth estimation. CNN features are used for computing pixel correspondences which, along with the stereo depth and known pose of the retrieved image, result in a set of 2D-3D correspondences. 

From scene coordinates, the camera pose can be computed using perspective-n-point~\cite{ransac} (PnP) methods that use the 2D pixel and 3D scene coordinate correspondences to triangulate the position and orientation of the camera. While these methods produce highly accurate results, they are sensitive to noise and require the use of a robust estimator, such as RANSAC~\cite{ransac}, and hypothesis refinement to achieve high accuracy. Though there is active research exploring the use of neural networks for generating samples for pose hypotheses, e.g., NG-RANSAC~\cite{ngransac}, our goal is to avoid this step altogether. Dang et al.~\cite{eigenfree} proposed an accurate alternative to RANSAC that predicts correspondence weighting using a CNN. However, the method of Dang et al. only works on a very specific type of problem that is solved by finding the minimum eigenvector of a matrix. There has been work on estimating point weights specifically for scene coordinate regression and pose estimation~\cite{scenecoordconfidence}, but accurate results still require RANSAC. 

\subsection{Monocular Depth Estimation}

Much progress has been made using CNNs for single image depth estimation~\cite{godard2017unsupervised,fu2018deep}. Importantly, recent work has shown that a learned representation for depth is sufficient for visual odometry in simultaneous localization and mapping (SLAM) systems~\cite{cnnslam,codeslam,dvso}. Other work uses stereo matching to improve depth estimation~\cite{tosi2019learning}. Ranftl et al.~\cite{ranftl2020towards} propose tools for mixing datasets during training, including a robust objective function that is invariant to changes in depth range and scale. Our work takes advantage of the success of monocular depth estimation to frame the absolute pose regression problem as an end-to-end learned alignment of corresponding point clouds.

\paragraph{Estimating Depth using Scene Coordinates} It is possible to compute depth explicitly from scene coordinates and camera calibration parameters using PnP algorithms. A detailed overview of solutions to this problem is presented by Xiao~\cite{pnpoverview}. From $N$ 2D-3D correspondences, a common approach involves solving $N$ systems of $\frac{(N-1)(N-2)}{2}$ equations to find the distance from each 3D point to the camera center. With five or more points, there exists a direct solution to the linear system~\cite{linearpnp}. While it is theoretically possible to perform the PnP depth estimation step in a differentiable way, it is extremely sensitive to noise. This is why the final solution in practice is found using robust estimation schemes, such as RANSAC, for accurate pose regression. Our proposed approach avoids the need for PnP and sampling altogether.

\subsection{Point Cloud Registration}

Point cloud registration is fundamental to feature-matching based image localization. The final step of pose retrieval in many PnP algorithms is absolute alignment of the 3D scene coordinates and recovered 3D camera coordinates~\cite{pnpoverview}. Modern approaches rely on features extracted from neural networks for high accuracy and robustness to noise. While recent work is largely focused on synthetic or single model alignment~\cite{pointnetlk, pcrnet, deepclosestpoint}, it has also been demonstrated that large point cloud scans can be accurately aligned using CNN architectures~\cite{deepvcp}. While we rely on point cloud registration for our method, it is a simpler case where we have explicit correspondences. Wang and Solomon~\cite{deepclosestpoint} explore registering single object point clouds. They predict correspondences from 3D input and use the Kabsch~\cite{kabsh} method for final alignment. We also use the Kabsch algorithm, but tackle the problem of camera pose estimation and extract 3D correspondences from an RBG image only.

\section{Method}

We present a direct pose estimation pipeline that combines scene coordinate regression, monocular depth estimation, and point cloud registration to estimate camera pose from corresponding point clouds extracted from an image.

\subsection{Problem Statement and Formulation}

We address the problem of single-image pose regression, in which we must estimate the camera pose with respect to a scene coordinate frame from a single image. The pose consists of two components, the camera orientation, $R$, and position, $~\vec{T}$. Together, these can be used to transform 3D positions in the camera frame, $A$, to the scene coordinate frame, $B$, using $B = RA+\vec{T}$. Given corresponding points between $A$ and $B$, it is possible to estimate the pose using various point cloud registration approaches. The challenge is that this estimation problem is sensitive to noise and it is difficult to find noise-free corresponding points.

We formulate our approach such that it reduces to the problem of point cloud registration. We directly regress points in the camera and scene coordinate frames for every pixel in the image. Each pixel thus defines a pair of corresponding points between the two coordinate frames. To address the problem of noise, we score each point correspondence using a per-pixel weighting mechanism prior to point cloud registration. Our approach is implemented as a sequence of differentiable neural network layers, enabling end-to-end optimization, deterministic inference, and a simple implementation. Our network architecture is shown in \figref{architecture}. We provide a detailed description of each component in the following subsections.

\subsection{Estimating Depth and Scene Coordinates}

We train a CNN to simultaneously estimate depth and scene coordinates directly from image content. We use a shared CNN backbone for initial feature map extraction and then the intermediate feature is passed to two separate sub-networks for independent depth and scene coordinate regression. 
For scene coordinate regression, we follow DSAC++~\cite{dsac2} and use a fully convolutional approach without upsampling layers. For this, we use the features from the shared feature extractor and pass them through a series of stride 1 convolutions such that the output size is $1/8$th the input image resolution. Smaller resolutions have shown to perform adequately~\cite{dsac2} and a smaller output size is more efficient, as computation of the cross-covariance matrix used for the Kabsch algorithm requires a multiplication of two $N \times 3$ matrices. 

For estimating monocular depth, we follow recent work~\cite{diggingdepth, codeslam} and perform both coarse and fine-scale depth estimation. We use the features from the shared feature extractor as input to an encoder/decoder network. We predict depth at two different scales in the network, optimizing each individually for accuracy. The largest resolution depth output is $1/8$th the size of the input such that it matches the scene coordinate resolution.

\subsection{Estimating Camera Pose}

We use the estimated depth and known camera geometry to compute 3D camera frame coordinates. For pixel $i$ with homogeneous pixel coordinate $\vec{u_i}$ and depth $d_i$, the camera frame point is computed using the known camera intrinsic matrix K as follows: $\vec{p_i} = d_iK^{-1}\vec{u_i}$.

Given scene coordinates and 3D camera frame coordinates, each pixel defines a pair of corresponding points in different reference frames. Thus, the problem now becomes pose estimation from corresponding point clouds. We solve this using the Kabsch method~\cite{kabsh} which we describe here. 

Finding the optimal pose between two point sets A and B amounts to solving the following minimization problem:
\begin{equation*}
    \argmin_{R,\vec{T}}\sum_i{(\vec{b_i} - R\vec{a_i}-\vec{T})^2}.
\end{equation*}
In the ideal case of noise-free data, we would assume all points have uniform weighting. However, directly optimizing for this results in low accuracy due to noise in the point positions, so we apply a per-point weighting term. The optimal rotation and translation can then be found by solving:
\begin{equation*}
    \argmin_{R,\vec{T}}\sum_i{w_i(\vec{b_i} - R\vec{a_i}-\vec{T})^2},
\end{equation*}
where $w_i$ is a weight assigned to the point pair $i$. 

To solve for $R$ and $~\vec{T}$, the translation component is first removed by centering both point clouds:
\begin{equation*}
    \begin{split}
        \vec{\mu_B} = \frac{\sum_i{w_ib_i}}{\sum_i{w_i}} \ \ \ \ \ \ \ \vec{\mu_A} = \frac{\sum_i{w_ia_i}}{\sum_i{w_i}} \\
        \bar{B} = B - \vec{\mu_B} \ \ \ \ \ \ \ \ \ \ \ \bar{A} = A - \vec{\mu_A}
    \end{split}.
\end{equation*}
We recover $R$ and $~\vec{T}$ as follows:
\begin{align*}
    USV^\textsf{T} & =  svd(\bar{A}^\textsf{T}W \bar{B}) \\
    d & = det(VU^\textsf{T})  \\
    R & = V \begin{pmatrix} 1 & 0 & 0 \\ 0 & 1 & 0 \\ 0 & 0 & d \end{pmatrix} U^\textsf{T} \\
    \vec{T} & = -R\vec{\mu_A} + \vec{\mu_B}.
\end{align*}
Our output representation for rotation is unique among direct pose estimation methods. Levinson et al.~\cite{svdforrot} show how this use of SVD to construct a orthonormal matrix is the most accurate among a large variety of rotation representations for deep networks.  

Intuitively, the point weights move the point towards ($w_i<1$) or away ($w_i>1$) from the origin. Motion towards the camera reduces the contribution of that point to the singular vectors. Likewise, motion away from the camera increases the contribution of that point. This allows for more robustness to noise without the need for hard thresholding or sampling. To generate the weights, we pass the correspondences through a scoring CNN that produces a per-pixel weighting. The input to this network is the concatenation of the output scene coordinates and unprojected camera frame coordinates computed from the output depth. The weighting is vital because the accuracy of the point cloud centroid depends on the accuracy of the points themselves. Since the estimated rotation is dependent on accurate centroid subtraction, and the estimated translation is dependent on \textit{both} the centroid and rotation estimate, it is important to reduce the impact of the noisy points. 

\subsection{Implementation Details}

\begin{table*}[t]
  \centering
  \ra{1.3}
  \resizebox{\linewidth}{!}{
    \begin{tabular}{lllllllll}
    \toprule
    Method & Chess & Fire & Heads & Office & Pumpkin & Kitchen & Stairs & Avg. \\ \midrule
    PoseNet~\cite{posenet}                  & 0.32/8.12 & 0.47/14.4 & 0.29/12.0 & 0.48/7.68 & 0.47/8.42 & 0.59/8.64 & 0.47/13.8 & 0.44/10.44\\
    PoseNet Learned Weights~\cite{posenet2} & 0.14/4.50 & 0.27/11.8 & 0.18/12.1 & 0.20/5.77 & 0.25/4.82 & 0.24/5.52 & 0.37/10.6 & 0.24/7.87\\
    Geo PoseNet~\cite{posenet2}             & 0.13/4.48 & 0.27/11.3 & 0.17/13.0 & 0.19/5.55 & 0.26/4.75 & 0.23/5.35 & 0.35/12.4 & 0.23/8.12\\
    LSTM PoseNet~\cite{lstmposenet}         & 0.24/5.77 & 0.34/11.9 & 0.21/13.7 & 0.30/8.08 & 0.33/7.00 & 0.37/8.83 & 0.40/13.7 & 0.31/9.85\\
    GPoseNet~\cite{gposenet}                & 0.20/7.11 & 0.38/12.3 & 0.21/13.8 & 0.28/8.83 & 0.37/6.94 & 0.35/8.15 & 0.37/12.5 & 0.31/9.95\\
    Hourglass PN~\cite{hourglasspose}       & 0.15/6.17 & 0.27/10.8 & 0.19/11.6 & 0.21/8.48 & 0.25/7.01 & 0.27/10.2 & 0.29/12.5 & 0.23/9.54\\
    BranchNet~\cite{branchnet}              & 0.18/5.17 & 0.34/8.99 & 0.20/14.2 & 0.30/7.05 & 0.27/5.10 & 0.33/7.40 & 0.38/10.3 & 0.29/8.32\\
    MapNet~\cite{mapnet}                    & 0.08/3.25 & 0.27/11.7 & 0.18/13.3 & 0.17/5.15 & 0.22/4.02 & 0.23/4.93 & 0.30/12.1 & 0.21/7.78\\
    MapNet++~\cite{mapnet}                  & 0.10/3.17 & \textbf{0.20}/9.04 & 0.13/11.1 & 0.18/5.38 & 0.19/3.92 & 0.20/5.01 & 0.30/13.4 & 0.19/7.29\\
    Sequence Enhancement~\cite{localglobal} & 0.09/3.28 & 0.26/10.92 & 0.17/12.70 & 0.18/ 5.45 & 0.20/3.66 & 0.23/4.92 & \textbf{0.23}/11.3 & 0.19/7.46\\
    ESAC no RANSAC & 0.12/2.96 & 0.28/7.58  & 1.04/60.68 & 0.48/9.05 & 0.21/4.32 & 0.31/6.88 & 0.58/10.25 & 0.43/14.53
 \\
    SC-conf no RANSAC~\cite{scenecoordconfidence} & - & - & 0.18/10.6 & - & - & - & - & - \\
    \midrule
    Ours                                    & \textbf{0.08}/\textbf{2.17} & 0.21/\textbf{6.14} & \textbf{0.13}/\textbf{7.93} & \textbf{0.11}/\textbf{2.65} & \textbf{0.14}/\textbf{3.34} & \textbf{0.12}/\textbf{2.75} & 0.29/\textbf{6.88} & \textbf{0.15}/\textbf{4.55}  \\
    \bottomrule
    
    \end{tabular}
  }
  \caption{Direct pose estimation results on 7Scenes compared to other methods (median position in meters/median rotation in degrees). We outperform all methods in rotation accuracy, typically by several degrees. Our method is only outperformed in position error for 2 scenes, and the difference is minor relative to the improvements in performance overall.}
  \label{tab:apr}
 
\end{table*}

We train our method in two stages, denoted as geometry and pose optimization respectively.

In the first stage, we train the depth estimation and scene coordinate regression networks jointly for accuracy. Given the scene coordinates, $C$, depth, $D$, and half resolution depth, $D_{1/2}$, we minimize the following loss function:
\begin{equation}
L_{geom} = ||C-\hat C||_1 + ||D-\hat D||_1 + ||D_{1/2}-\hat D_{1/2}||_1. \nonumber
\end{equation}

Notably, we do not use a validity mask during training which is common in depth and scene coordinate regression work. This encourages the network to explicitly learn areas of the image which do not have valid depth data in the training set. We train this stage for 50 epochs with the Adam optimizer~\cite{adam} using an initial learning rate of $1e^{-4}$. The learning rate is reduced by a factor of $0.1$ every 20 epochs. The output depth is mapped to the range $[0,1]$ using the \emph{sigmoid} activation and we normalize the target depth about the mean as in CodeSLAM~\cite{codeslam}. For each scene, we determine the mean scene coordinate and subtract this value from the target scene coordinates for optimization. At inference, the mean is added to the scene coordinate outputs. We use $\alpha=0.5$ for weighting the loss on the half resolution depth prediction.

In the second stage, we train the weighting network with scene coordinates and camera coordinates from the previous step. We apply a \emph{sigmoid} activation at the end of our weighting network such that weight values are in the range $[0,1]$.  We train this stage for 10 epochs with the Adam optimizer using a learning rate of $1e^{-3}$. While we do not explicitly have loss terms on the depth and scene coordinate outputs in this stage, we do allow the relevant network weights to update by using a learning rate of $1e^{-6}$. This stage is purely optimized for the accuracy of the final pose rotation, $R$, and translation, $\vec{T}$ with the following function:
\begin{equation*}
    L_{pose} = ||R-\hat R||_1+ ||\vec{T}-\vec{\hat T}||_1.
\end{equation*}
All input images are resized to $640 \times 480$, resulting in an output depth and scene coordinate resolution of $80 \times 60$.

The backbone of our network is ResNet34~\cite{resnet}. The depth and scene coordinate sub-networks share the first half of the backbone and split after the second residual block. The scene coordinate regression sub-network consists of a series of $3 \times 3$ stride 1 convolutions with ReLU activations. The depth network is based on the LinkNet~\cite{linknet} segmentation network. The weighting network is a series of $3 \times 3$ stride 1 convolutions with ReLU activations. Please see the supplemental material for more details about the CNN architectures used.

\section{Evaluation}

We evaluate our method both quantitatively and qualitatively through a variety of experiments using well-known benchmark datasets.

\subsection{Datasets}

We report results on two common benchmark datasets. The 7Scenes dataset is a collection of 7 unique indoor scenes of varying size and localization difficulty. Each scene has a set of sequences containing 500 or 1000 frames of RGB, depth, and pose information resulting in 1000 to 7000 frames for training. The 12Scenes~\cite{12scenes} dataset is a more difficult indoor dataset for absolute pose regression. It contains few training images relative to the size of the space for each scene which makes training accurate absolute pose regression models difficult, whereas structure-based methods perform very well on this dataset. For each scene in these datasets, ground-truth depth labels are found by ray-casting into structure-from-motion models using the ground-truth pose information. The pose and depth labels are used for computing ground-truth scene coordinates. Please see the supplemental material for additional results on these datasets as well as results on outdoor scenes.

\subsection{Quantitative Evaluation}

\tblref{apr} shows how our method compares to several RGB only direct pose estimation methods on the 7Scenes dataset. For ``SC-conf" we report the 2D-3D, single hypothesis metric which uses only RGB input. We only report for the \emph{heads} scene because this is all that is reported in the paper and a public implementation is not provided by the authors. For ``ESAC no RANSAC" we used the scene coordinate regressors from ESAC and used the EPnP~\cite{epnp} algorithm for final pose estimation without RANSAC or refinement. Our method dominates all other methods in rotation accuracy. Our rotation error is often less than $70\%$ of the next best method, and is better by at least $7\%$ in all cases. While not as dominant for position error, we match or outperform the next best method in all but two scenarios, one of which we obtain very similar results. The only scene where our method is significantly outperformed is \emph{stairs}. This scene is very challenging even for indirect methods\cite{scenecoordforest}. Note that while our method does require depth ground-truth for training, we still show significant improvements over other methods that make use of depth for training, such as Geo-PoseNet~\cite{posenet2}. Like all of these approaches, we make use of only the input RGB imagery without depth input at test time.

While it is clear that our approach outperforms all direct absolute pose regression methods, it is important to also compare to indirect methods that rely on techniques such as image retrieval, RANSAC, correspondence matching, etc. \tblref{morecomparisons} shows how  our method compares to several recent indirect methods. Our method outperforms several of these methods. It typically takes several additional techniques in order for a method to significantly outperform our direct method. 

We show the cumulative histogram of errors for all images from the 7Scenes dataset in \figref{histograms} comparing our method to PoseNet, MapNet, and the scene oracle approach from ESAC~\cite{esac}. Additionally, we compare to the same methods for overall accuracy and median error on the complete dataset in \tblref{accuracytab}. While the improvement from PoseNet to MapNet is minor, our method shows a significant improvement over MapNet. These comparisons better illustrate that while our method is still less accurate than the most effective single image localization methods, it provides a sizable improvement over the next best direct method.

Additional results on 12Scenes are shown in \tblref{12scenes}. These scenes are very difficult for PoseNet-style approaches, leading to very poor performance. However, our method is able to achieve accuracy very similar to PnLP, a method that performs retrieval, correspondence matching, RANSAC, and refinement.

\begin{table}[t]
\centering
\ra{1.3}
\begin{adjustbox}{max width=\textwidth}
\begin{tabular}{lll}\toprule
Method & Error & Techniques  \\ \midrule
Ours                   & 0.15/4.55 \\
\midrule
Bayesian PN~\cite{baysposenet} & \rt{0.47}/\rt{9.81} & Avg \\
MapNet+PGO~\cite{mapnet}                & \rt{0.18}/\rt{6.56} & Ref\\
DenseVLAD~\cite{densevlad} &  \rt{0.26}/\rt{13.11} & Ret  \\
AnchorNet~\cite{anchornet} & 0.10/\rt{6.74} & Avg\\
RelocNet~\cite{relocnet} & \rt{0.21}/\rt{6.73} & Ret, Avg\\
PnLP~\cite{pnlp} & 0.12/3.93 &  Ret, Cor, RAN, Ref\\
Active Search~\cite{activesearch} & 0.05/2.46 & Cor, RAN, Ref \\
SCoRe~\cite{scenecoordforest} & 0.08/1.60 & RAN, Ref\\
CamNet~\cite{canmet}   & 0.04/1.69 & Ret, Avg \\
SANet~\cite{sanet} & 0.05/1.68 & Ret, Cor, RAN, Ref \\
SC-conf~\cite{scenecoordconfidence} & 0.059/3.057 & RAN, Ref \\
DSAC++~\cite{dsac2} & 0.03/1.10 & RAN, Ref\\

\bottomrule
\end{tabular}
\end{adjustbox}
\caption{We compare to indirect pose estimation methods that rely on some combination of retrieval (Ret), image correspondence (Cor), hypothesis averaging (Avg), RANSAC (RAN), and refinement (Ref). We highlight in red instances where a method is outperformed by our approach. Results are shown as the average median error across all scenes from 7Scenes.}
\label{tab:morecomparisons}
\end{table}

\begin{table}
\centering
\begin{tabular}{llcc}\toprule
        & Acc. & Pos Err (m) & Rot Err (deg)   \\ \midrule
PoseNet   & 0.024 & 0.212  & 6.54  \\
MapNet & 0.052  & 0.199 & 5.794       \\
Ours & 0.116 & 0.131 &  3.287  \\
ESAC & 0.752 & 0.030 & 0.946   \\
\bottomrule
\end{tabular}
\caption{We report accuracy, as well as median and mean position and orientation error for all images from 7Scenes across several methods. Accuracy is defined as the ratio of images correctly localized within 5 cm and 5 degrees. We also report the median position and rotation error.}
\label{tab:accuracytab}
\end{table}

\begin{figure}
  \includegraphics[trim=60 0 50 30,clip,width=\linewidth]{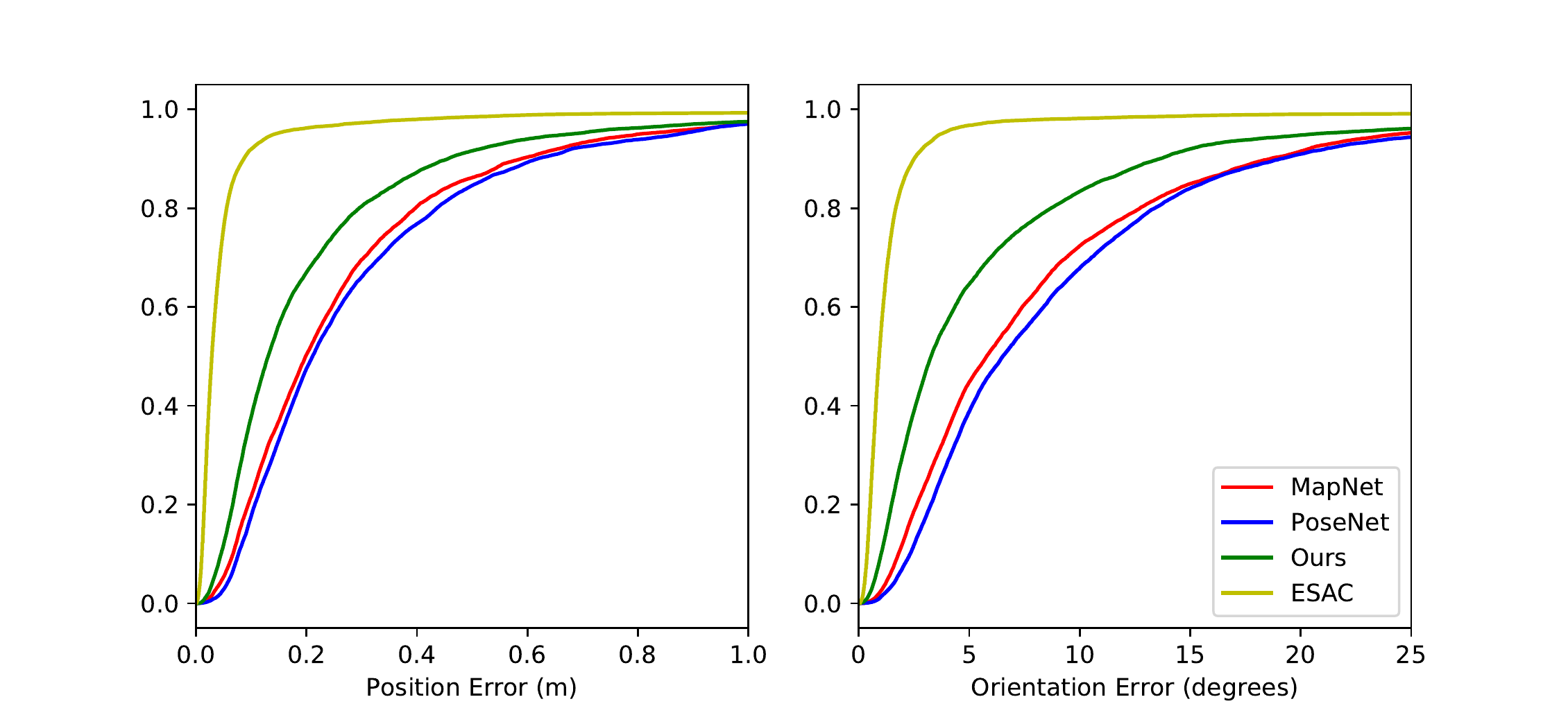}
  \caption{Cumulative histograms of error for all scenes from 7Scenes for several methods, truncated to 1 meter and 25 degrees error.}
  
  \label{fig:histograms}
\end{figure}

\begin{table}
\centering
\ra{1.1}
\begin{tabular}{lllll}\toprule
        & PoseNet & Ours  & PnLP   & ESAC \\ \midrule
Kitchen1   & 0.29/15.48   & 0.08/4.45    & 0.09/4.1     & 0.01/0.44  \\
Living1    & 0.29/15.31   & 0.08/2.50    & 0.08/2.9     & 0.01/0.43  \\
Kitchen2   & 0.21/18.18   & 0.08/2.96    & 0.10/3.7     & 0.01/0.46  \\
Living2    & 0.31/23.58   & 0.09/3.13    & 0.10/4.7     & 0.01/0.40  \\
Bed       & 0.57/17.85   & 0.07/3.87    & 0.12/5.7     & 0.01/0.46  \\
Luke      & 0.35/20.07   & 0.12/4.82    & 0.14/5.5     & 0.01/0.59  \\
Office 5a      & 0.57/14.55   & 0.10/5.08    & 0.09/3.6     & 0.01/0.59  \\
Office 5b      & 0.47/15.49   & 0.09/2.57    & 0.10/3.7     & 0.02/0.59  \\
Lounge         & 0.29/18.42   & 0.07/2.53    & 0.10/3.5     & 0.02/0.61  \\
Manolis        & 0.22/17.45   & 0.09/3.52    & 0.09/3.7     & 0.01/0.53  \\
Gates362       & 0.27/16.71   & 0.06/2.04    & 0.10/4.7     & 0.01/0.46  \\
Gates381       & 0.37/20.52   & 0.12/5.02    & 0.11/4.4     & 0.01/0.67  \\
\bottomrule
\end{tabular}
\caption{Comparison of median position and orientation error for several methods on the 12Scenes dataset.}
\label{tab:12scenes}
\end{table}

\begin{table*}[t]
  \centering
  \ra{1.3}
  \begin{adjustbox}{max width=\textwidth}
   \begin{tabular}{llllllllll}\toprule
      Pose & Cor& Loss  \\
       Opt &  Weighting & Mask & Chess & Fire & Heads & Office & Pumpkin & Kitchen & Stairs \\ \midrule
      &  &  & 0.36/10.72 & 0.29/9.06 & 0.23/12.69 & 0.15/3.81 & 0.21/4.92 & 0.17/4.10 & 0.35/8.15 \\

      \checkmark & & & 0.36/10.70 & 0.26/9.10 & 0.22/12.4 & 0.14/3.68 & 0.20/4.86 & 0.16/4.08 & 0.34/7.89 \\
       & \checkmark &  & 0.08/2.27 & 0.23/6.01 & 0.14/\textbf{7.75} & 0.11/2.70 & 0.15/3.47 & 0.13/2.93 & 0.31/7.14 \\
       \checkmark & \checkmark & & \textbf{0.08}/\textbf{2.17} & \textbf{0.21}/6.14 & \textbf{0.13}/7.93 &  \textbf{0.11}/\textbf{2.65} & \textbf{0.14}/\textbf{3.34} & \textbf{0.12}/\textbf{2.75} & \textbf{0.29}/6.88  \\
       \midrule
        & & \checkmark & 0.14/3.70 & 0.25/8.41 & 0.23/15.52 & 0.14/3.59 & 0.17/3.86 & 0.16/4.21 & 0.30/7.89 \\
      \checkmark & \checkmark & \checkmark & 0.08/2.46 & 0.21/\textbf{5.51} & 0.16/10.67 &  0.11/2.75 & 0.18/4.17 & 0.14/2.99 & 0.29/\textbf{6.36} \\

    \bottomrule
   \end{tabular}
  \end{adjustbox}
  \caption{We compare the effect of each component of our system for all scenes of 7Scenes. Pose opt indicates that the pose optimization stage that optimizes $L_{pose}$ has been performed. Cor weighting indicates that the learned correspondence weights are being applied. Loss mask indicates that a validity mask was used with $L_{geom}$ during training.}
  \label{tab:ablation}
\end{table*}

\subsection{Ablation Study}
We evaluate several different configurations of our approach. First we evaluate the pose performance of the output depth and scene coordinates before the pose optimization step has occurred. Next, we evaluate the effect of our correspondence weighting network by evaluating pose both with and without the predicted weighting. Additionally, we show results which use a masked version of $L_{geom}$ which only considers valid regions of the image in loss computation. Ablation study results are given in \tblref{ablation}. \newline

\newcommand{\ti}[1]{\includegraphics[width=27mm]{images/7scenes/#1}}

\begin{figure*}[t]
  \centering
  \begin{tabular}{c|ccc|c}
  & \multicolumn{3}{c|}{Network Outputs} & Computed \\
    Input &  Depth &  Scene Coords &  Weights & Endpoint Error\\ 
   
    \ti{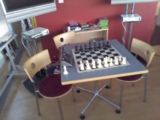} & \ti{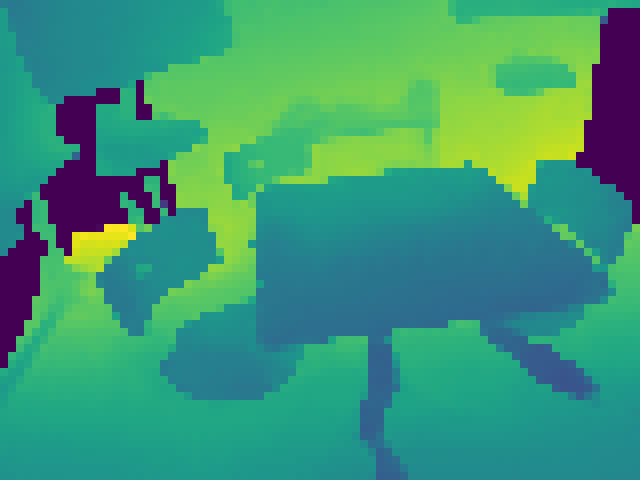} & \ti{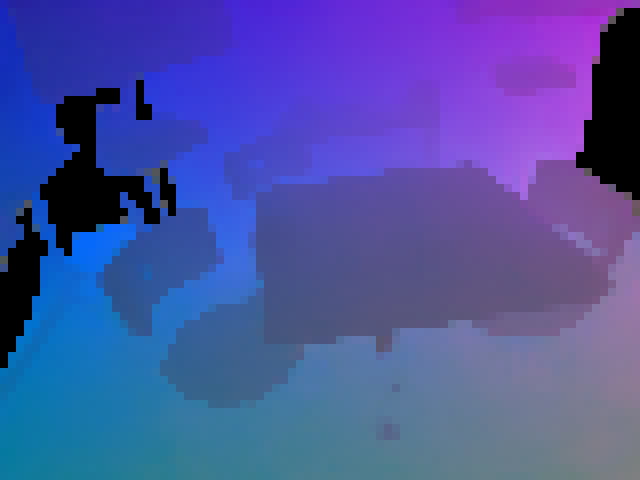} & \ti{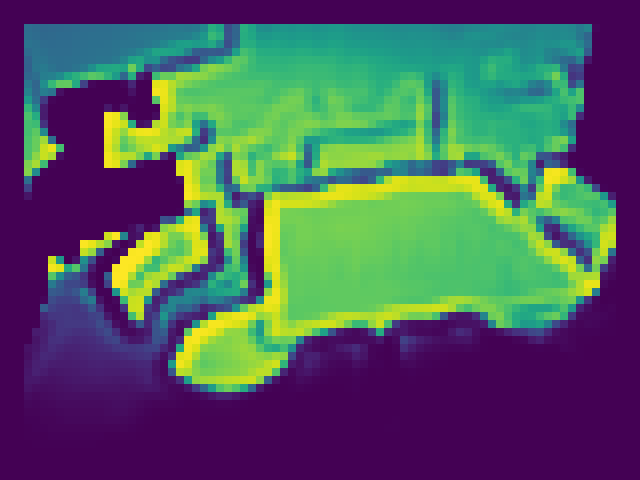} & \ti{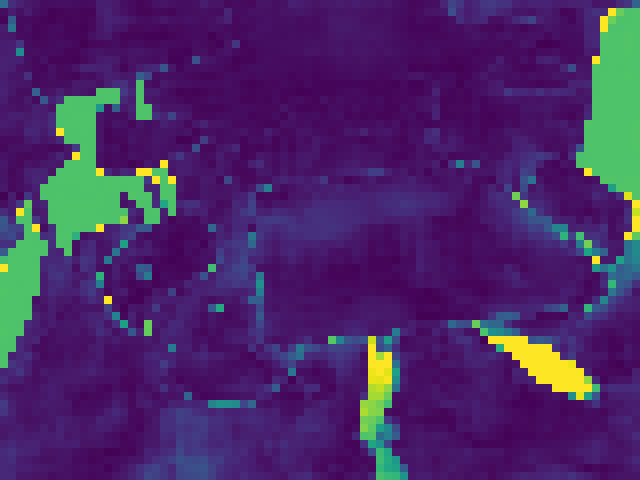}  \\
    
    \ti{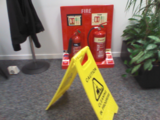} & \ti{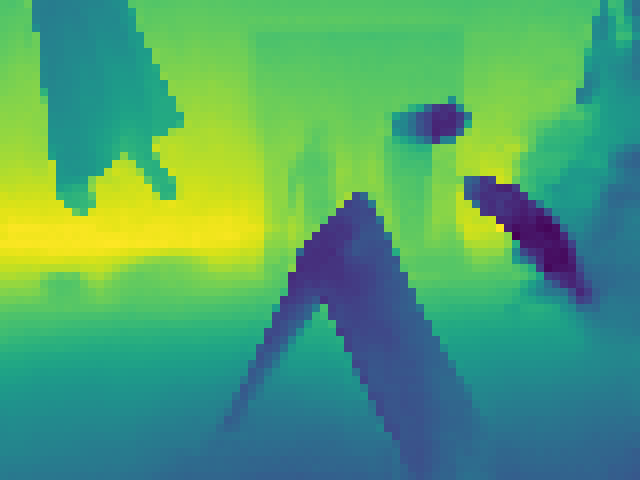} & \ti{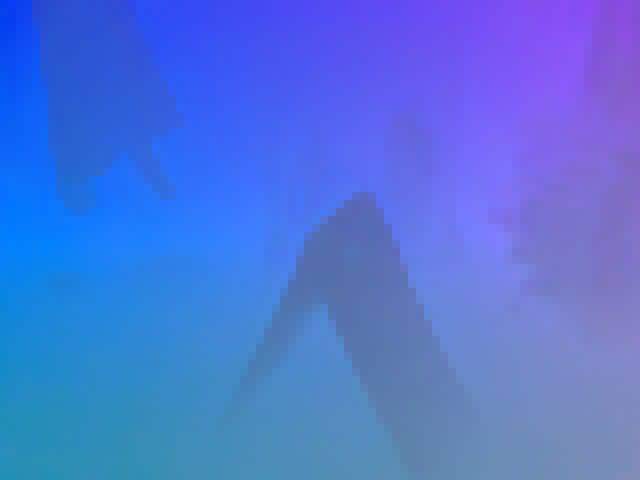} & \ti{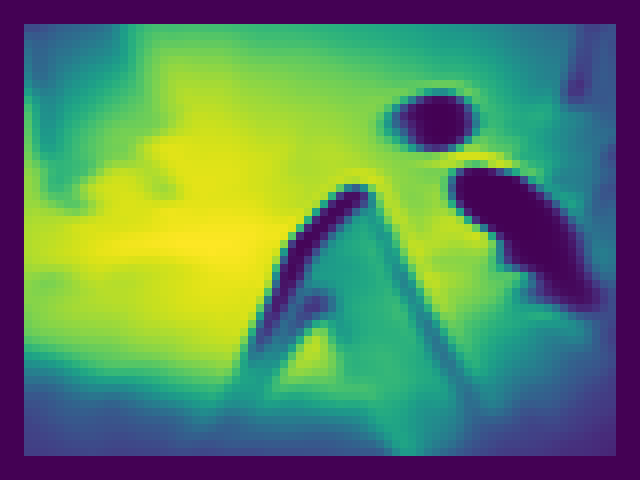} & \ti{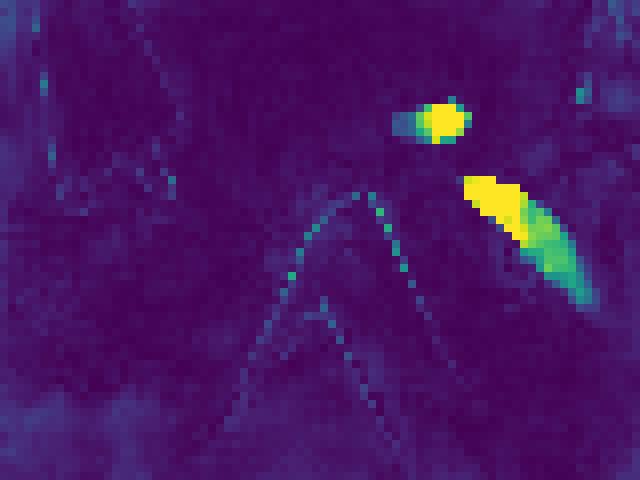} \\

    \ti{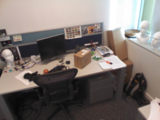} & \ti{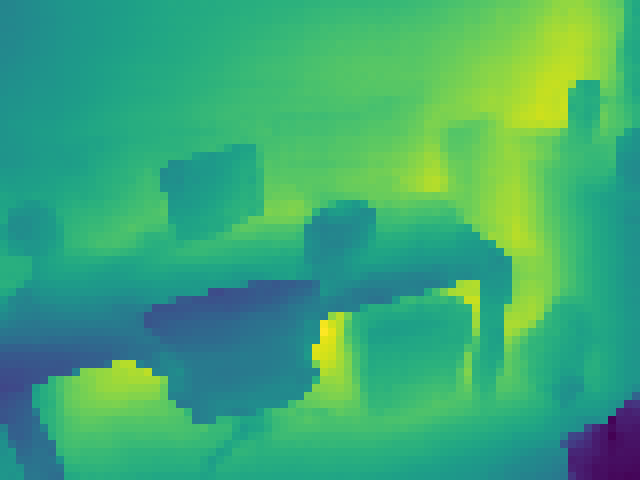} & \ti{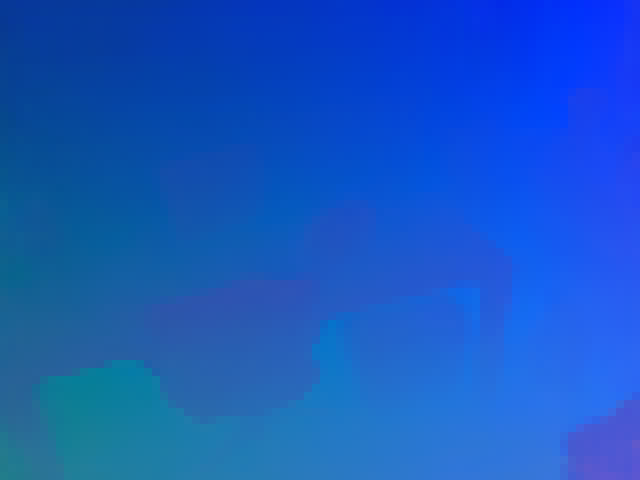} & \ti{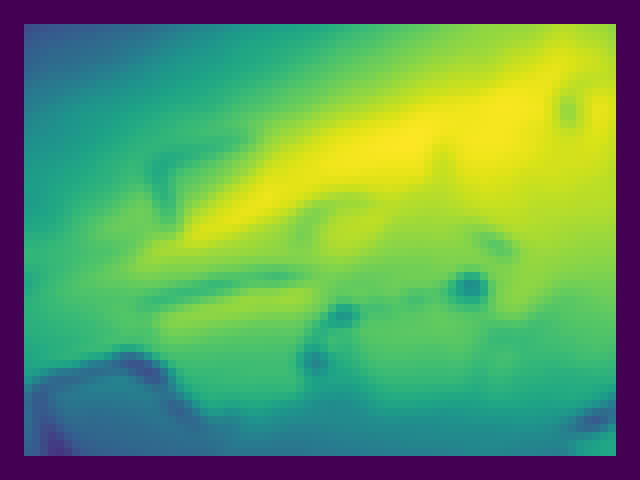} & \ti{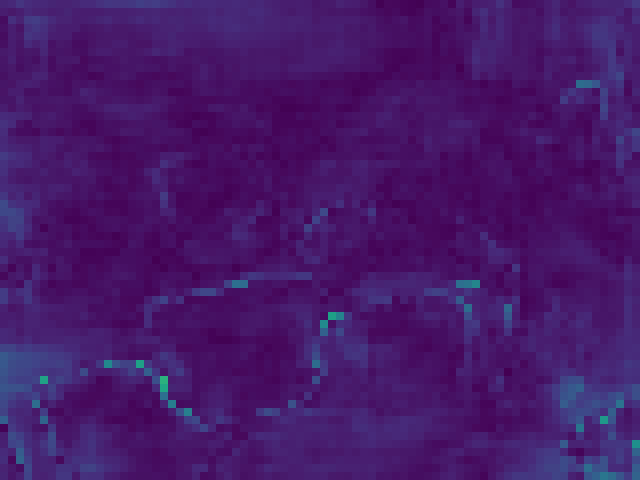} \\

    \ti{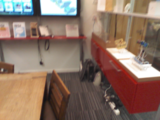} & \ti{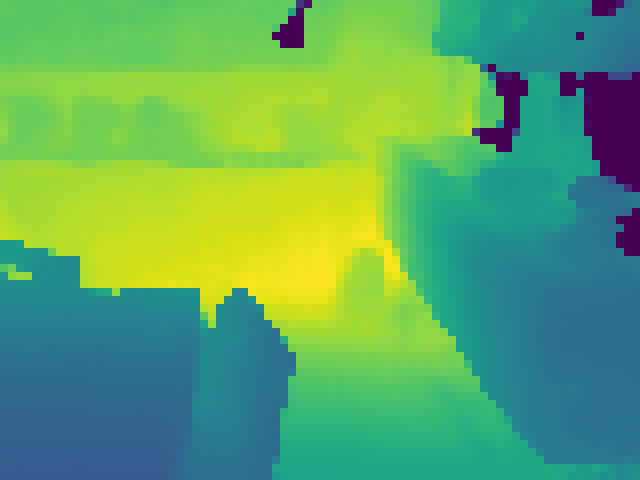} & \ti{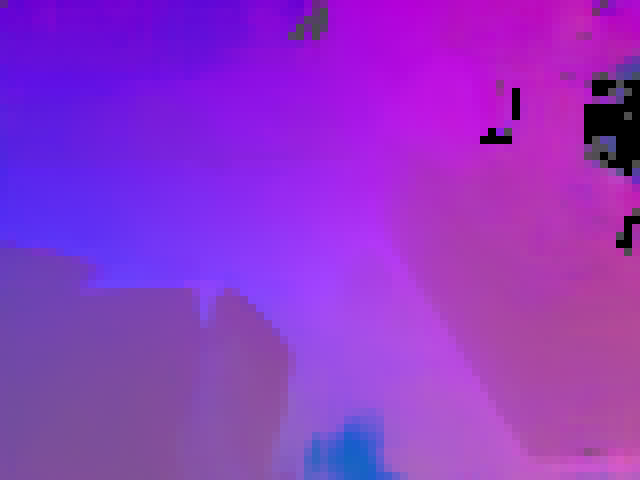} & \ti{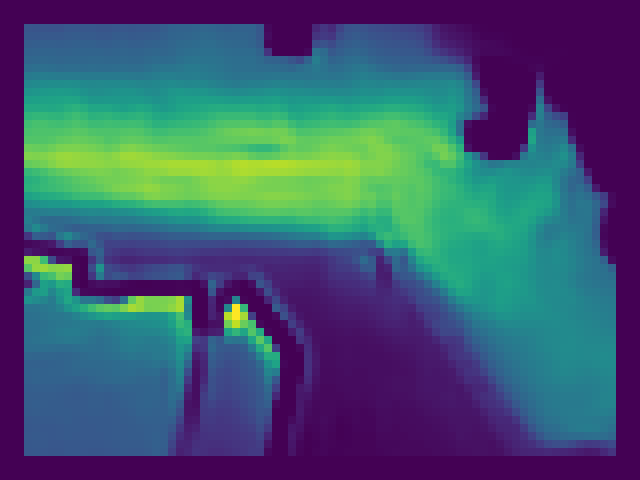} &  \ti{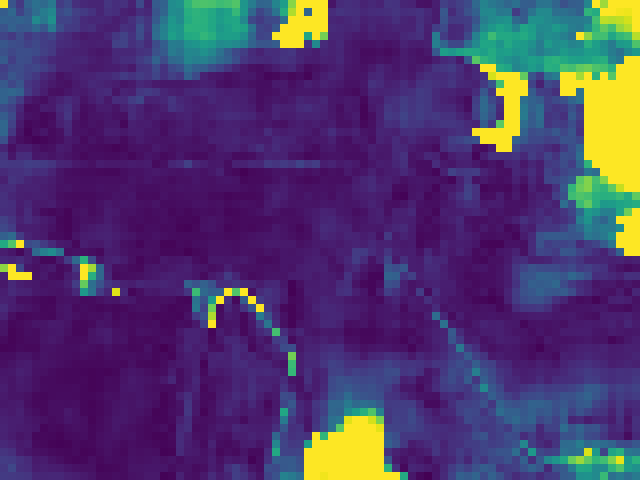} \\

    \ti{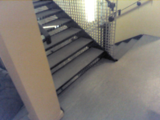} & \ti{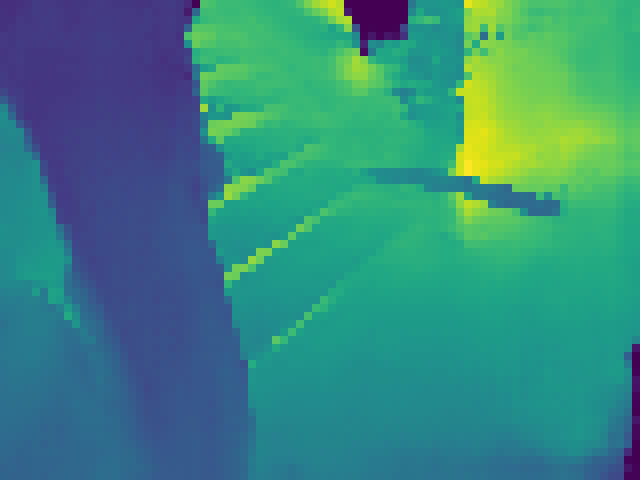} & \ti{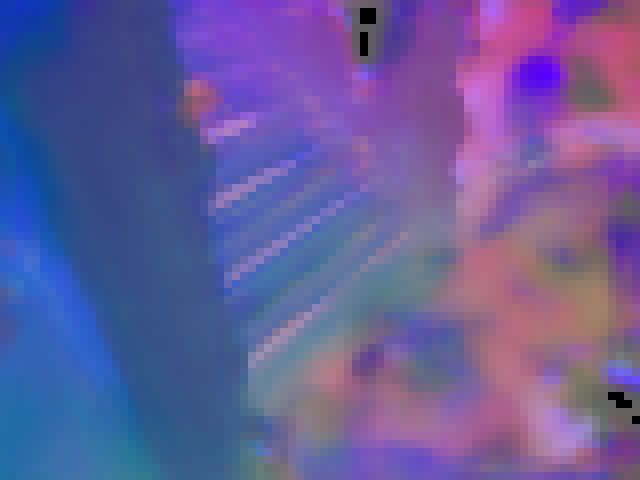} & \ti{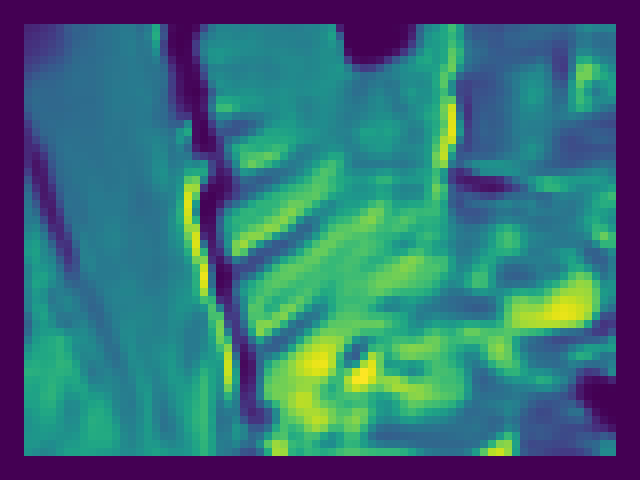} & \ti{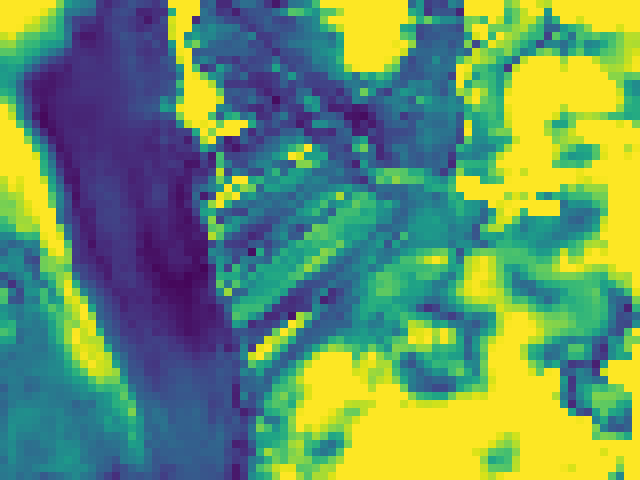} 
  \end{tabular}
  \caption{Example network outputs. Depth, scene coordinates, and weights are direct outputs of our network. The endpoint error is computed by applying the regressed pose to the estimated camera coordinates and comparing to the estimated scene coordinates. The bottom row shows a failure case with a bad final pose result caused by poor scene coordinate estimates.}
  \label{fig:examples}
\end{figure*}

\noindent\textbf{Effect of the Pose Optimization Stage} First we test the pose optimization stage. This is the second stage of training that optimizes directly for pose accuracy through $L_{pose}$. Surprisingly, the outputs prior to this stage are reasonable. However, in many cases we see a slight decrease in median error from the pose optimization stage even when not applying the correspondence weighting, indicating that depth and scene coordinate estimates were improved. \newline

\noindent\textbf{Effect of Correspondence Weighting} Next, we see the effect of using the predicted correspondence weighting. As expected, the application of the weights drastically reduces the error. This shows that the weighting network is learning to accurately segment incorrect correspondences.  \newline

\noindent\textbf{Effect of Validity Mask Training} It is common in dense prediction tasks such as depth estimation and scene coordinate regression to train only against ``valid" data, that is, pixels where ground-truth labels are available. Our final test shows the results of applying a validity mask to $L_{geom}$. While performance is similar, we found that training without the validity mask results in improved performance in most cases. An interesting result to note is that when training with the loss mask it is possible to get results on par with, and in many cases much better than, many other direct pose estimation methods from \tblref{apr} even without correspondence weighting. Performance in this case is better than in the case of not using a validity mask during training due to the mean subtraction of the point clouds in the alignment phase. The unmasked loss results in zeros being predicted, which while they can be easily detected and ignored by the correspondence weighting CNN, negatively effect the point cloud centroids. On the other hand, loss masking prevents zero values from being predicted. Though error still decreases in this case with the use of the predicted weights. 

\subsection{Qualitative Results}

Qualitative outputs of depth, scene coordinates, and correspondence weights are shown in \figref{examples}. To assess the quality of the estimated scene geometry and final pose estimate, the endpoint error after applying the regressed pose to the predicted point clouds is also given. The endpoint error is clamped to $1m$ for visualization purposes, with dark colors representing lower values and bright colors representing high values. Overall, it appears that the weighting is learning to recognize regions which are inconsistent with the rest of the point clouds. This is apparent as the output weights are often similar and opposite to the endpoint error after applying the pose. For example, the second row shows an example where the predicted scene coordinates are accurate, but the depth prediction is incorrect. This is correctly captured in the weighting network. The bottom row shows a difficult failure case. The depth estimates are reasonable, but the high noise present in the scene coordinates confuses the weighting network, leading to an incorrect pose.

\section{Discussion}

While it is uncertain in general how exactly CNNs see depth in monocular images~\cite{howdepth}, it is clear in our case that both the depth and scene coordinate networks are effectively memorizing scene layout and geometry. Luckily, in the case of single image localization, this is precisely what is desired. This type of memorization, unlike PoseNet, works well for pose estimation because the memorized objects have semantic and structural meaning. Instead of interpolating between a set of learned poses, we are able to exploit the high capacity of the network to effectively store two 3D copies of the scene which can then be recalled and used for pose estimation. 

We believe this is advantageous over the less explicit PoseNet approaches for several reasons. First, as we have shown, this allows for much lower error in general across a wide variety of scenes. Second, this allows for explainable pose estimation. For PoseNet-style methods, it is difficult to quantify what about an image makes it difficult for pose regression. On the other hand, the intermediate outputs of our method allow for explicit evaluation of geometric consistency. We believe this added benefit of explainability will be useful for investigating the shortcomings of direct pose estimation, leading to better architecture and optimization design choices.

\section{Conclusion}

We presented a pose estimation approach that has many of the desirable properties of PoseNet-style approaches in that it is fully differentiable, uses only feed-forward processing, and has a constant runtime, but significantly improved accuracy. Unlike PoseNet, which uses a generic CNN to perform pose estimation, our approach is composed of modules with specific geometric functions. This means that our method is not only more accurate, but it is also easier to understand when it fails. While our approach has not achieved the accuracy of the state-of-the-art indirect pose estimation methods, it begins to close the performance gap.

\section*{Acknowledgements}

We gratefully acknowledge the support of the National Science Foundation (IIS-1553116). We thank Torsten Sattler for the valuable feedback.

{\small
\bibliographystyle{ieee_fullname}
\bibliography{biblio}

\begin{thebibliography}{10}\itemsep=-1pt

\bibitem{relocnet}
Vassileios Balntas, Shuda Li, and Victor Prisacariu.
\newblock Relocnet: Continuous metric learning relocalisation using neural
  nets.
\newblock In {\em European Conference on Computer Vision}, 2018.

\bibitem{codeslam}
Michael Bloesch, Jan Czarnowski, Ronald Clark, Stefan Leutenegger, and
  Andrew~J. Davison.
\newblock Codeslam — learning a compact, optimisable representation for dense
  visual slam.
\newblock In {\em IEEE Conference on Computer Vision and Pattern Recognition},
  2018.

\bibitem{dsac}
Eric Brachmann, Alexander Krull, Sebastian Nowozin, Jamie Shotton, Frank
  Michel, Stefan Gumhold, and Carsten Rother.
\newblock Dsac - differentiable ransac for camera localization.
\newblock In {\em IEEE Conference on Computer Vision and Pattern Recognition},
  2017.

\bibitem{dsac2}
Eric Brachmann and Carsten Rother.
\newblock Learning less is more - 6d camera localization via 3d surface
  regression.
\newblock In {\em IEEE Conference on Computer Vision and Pattern Recognition},
  2018.

\bibitem{esac}
Eric Brachmann and Carsten Rother.
\newblock Expert sample consensus applied to camera re-localization.
\newblock In {\em International Conference on Computer Vision}, 2019.

\bibitem{ngransac}
Eric Brachmann and Carsten Rother.
\newblock {N}eural- {G}uided {RANSAC}: {L}earning where to sample model
  hypotheses.
\newblock In {\em International Conference on Computer Vision}, 2019.

\bibitem{mapnet}
Samarth Brahmbhatt, Jinwei Gu, Kihwan Kim, James Hays, and Jan Kautz.
\newblock Geometry-aware learning of maps for camera localization.
\newblock In {\em IEEE Conference on Computer Vision and Pattern Recognition},
  2018.

\bibitem{scenecoordconfidence}
Mai Bui, Shadi Albarqouni, Slobodan Ilic, and Nassir Navab.
\newblock Scene coordinate and correspondence learning for image-based
  localization.
\newblock In {\em British Machine Vision Conference}, 2018.

\bibitem{gposenet}
Ming Cai, Chunhua Shen, and Ian~D. Reid.
\newblock A hybrid probabilistic model for camera relocalization.
\newblock In {\em British Machine Vision Conference}, 2018.

\bibitem{linknet}
Abhishek Chaurasia and Eugenio Culurciello.
\newblock Linknet: Exploiting encoder representations for efficient semantic
  segmentation.
\newblock In {\em IEEE Visual Communications and Image Processing}, 2017.

\bibitem{eigenfree}
Zheng Dang, Kwang~Moo Yi, Yinlin Hu, Fei Wang, Pascal Fua, and Mathieu
  Salzmann.
\newblock Eigendecomposition-free training of deep networks with zero
  eigenvalue-based losses.
\newblock In {\em European Conference on Computer Vision}, 2018.

\bibitem{howdepth}
Tom~van Dijk and Guido~de Croon.
\newblock How do neural networks see depth in single images?
\newblock In {\em International Conference on Computer Vision}, 2019.

\bibitem{canmet}
Mingyu Ding, Zhe Wang, Jiankai Sun, Jianping Shi, and Ping Luo.
\newblock Camnet: Coarse-to-fine retrieval for camera re-localization.
\newblock In {\em International Conference on Computer Vision}, 2019.

\bibitem{ransac}
Martin~A. Fischler and Robert~C. Bolles.
\newblock Random sample consensus: A paradigm for model fitting with
  applications to image analysis and automated cartography.
\newblock {\em Communications of the ACM}, 24(6):381–395, 1981.

\bibitem{fu2018deep}
Huan Fu, Mingming Gong, Chaohui Wang, Kayhan Batmanghelich, and Dacheng Tao.
\newblock Deep ordinal regression network for monocular depth estimation.
\newblock In {\em IEEE Conference on Computer Vision and Pattern Recognition},
  2018.

\bibitem{godard2017unsupervised}
Cl{\'e}ment Godard, Oisin Mac~Aodha, and Gabriel~J Brostow.
\newblock Unsupervised monocular depth estimation with left-right consistency.
\newblock In {\em IEEE Conference on Computer Vision and Pattern Recognition},
  2017.

\bibitem{diggingdepth}
Clement Godard, Oisin Mac~Aodha, Michael Firman, and Gabriel~J. Brostow.
\newblock Digging into self-supervised monocular depth estimation.
\newblock In {\em International Conference on Computer Vision}, 2019.

\bibitem{pointnetlk}
Hunter Goforth, Yasuhiro Aoki, Arun~Srivatsan Rangaprasad, and Simon Lucey.
\newblock Pointnetlk: Robust \& efficient point cloud registration using
  pointnet.
\newblock In {\em IEEE Conference on Computer Vision and Pattern Recognition},
  2019.

\bibitem{resnet}
Kaiming He, Xiangyu Zhang, Shaoqing Ren, and Jian Sun.
\newblock Deep residual learning for image recognition.
\newblock In {\em IEEE Conference on Computer Vision and Pattern Recognition},
  2016.

\bibitem{kabsh}
Wolfgang Kabsch.
\newblock A solution for the best rotation to relate two sets of vectors.
\newblock {\em Acta Crystallographica Section A: Crystal Physics, Diffraction,
  Theoretical and General Crystallography}, 32(5):922--923, 1976.

\bibitem{baysposenet}
Alex Kendall and Roberto Cipolla.
\newblock Modelling uncertainty in deep learning for camera relocalization.
\newblock In {\em IEEE International Conference on Robotics and Automation},
  2016.

\bibitem{posenet2}
Alex Kendall and Roberto Cipolla.
\newblock Geometric loss functions for camera pose regression with deep
  learning.
\newblock {\em IEEE Conference on Computer Vision and Pattern Recognition},
  2017.

\bibitem{posenet}
Alex Kendall, Matthew Grimes, and Roberto Cipolla.
\newblock Posenet: A convolutional network for real-time 6-dof camera
  relocalization.
\newblock 2015.

\bibitem{adam}
Diederik~P Kingma and Jimmy Ba.
\newblock Adam: A method for stochastic optimization.
\newblock {\em arXiv preprint arXiv:1412.6980}, 2014.

\bibitem{epnp}
Vincent Lepetit, Francesc Moreno-Noguer, and Pascal Fua.
\newblock Epnp: An accurate o (n) solution to the pnp problem.
\newblock {\em International Journal of Computer Vision}, 81(2):155, 2009.

\bibitem{svdforrot}
Jake Levinson, Carlos Esteves, Kefan Chen, Noah Snavely, Angjoo Kanazawa,
  Afshin Rostamizadeh, and Ameesh Makadia.
\newblock An analysis of svd for deep rotation estimation.
\newblock {\em arXiv preprint arXiv:2006.14616}, 2020.

\bibitem{deepvcp}
Weixin Lu, Guowei Wan, Yao Zhou, Xiangyu Fu, Pengfei Yuan, and Shiyu Song.
\newblock Deepvcp: An end-to-end deep neural network for point cloud
  registration.
\newblock In {\em International Conference on Computer Vision}, 2019.

\bibitem{pnpoverview}
Xiao~Xin Lu.
\newblock A review of solutions for perspective-n-point problem in camera pose
  estimation.
\newblock In {\em Journal of Physics: Conference Series}, volume 1087, page
  052009, 2018.

\bibitem{hourglasspose}
Iaroslav Melekhov, Juha Ylioinas, Juho Kannala, and Esa Rahtu.
\newblock Image-based localization using hourglass networks.
\newblock In {\em Proceedings of the IEEE International Conference on Computer
  Vision Workshops}, 2017.

\bibitem{svspose}
Tayyab Naseer and Wolfram Burgard.
\newblock Deep regression for monocular camera-based 6-dof global localization
  in outdoor environments.
\newblock In {\em IEEE/RSJ International Conference on Intelligent Robots and
  Systems}, 2017.

\bibitem{pytorch}
Adam Paszke, Sam Gross, Francisco Massa, Adam Lerer, James Bradbury, Gregory
  Chanan, Trevor Killeen, Zeming Lin, Natalia Gimelshein, Luca Antiga, et~al.
\newblock Pytorch: An imperative style, high-performance deep learning library.
\newblock In {\em Advances in Neural Information Processing Systems}, 2019.

\bibitem{pnlp}
Nathan Piasco, D{\'e}sir{\'e} Sidib{\'e}, C{\'e}dric Demonceaux, and
  Val{\'e}rie Gouet-Brunet.
\newblock Perspective-n-learned-point: Pose estimation from relative depth.
\newblock In {\em British Machine Vision Conference}, 2019.

\bibitem{linearpnp}
Long Quan and Zhongdan Lan.
\newblock Linear n-point camera pose determination.
\newblock {\em IEEE Transactions on Pattern Analysis and Machine Intelligence},
  21(8):774--780, 1999.

\bibitem{ranftl2020towards}
Ren\'{e} Ranftl, Katrin Lasinger, David Hafner, Konrad Schindler, and Vladlen
  Koltun.
\newblock Towards robust monocular depth estimation: Mixing datasets for
  zero-shot cross-dataset transfer.
\newblock {\em IEEE Transactions on Pattern Analysis and Machine Intelligence},
  2020.

\bibitem{anchornet}
Soham Saha, Girish Varma, and CV Jawahar.
\newblock Improved visual relocalization by discovering anchor points.
\newblock {\em arXiv preprint arXiv:1811.04370}, 2018.

\bibitem{pcrnet}
Vinit Sarode, Xueqian Li, Hunter Goforth, Yasuhiro Aoki, Rangaprasad~Arun
  Srivatsan, Simon Lucey, and Howie Choset.
\newblock Pcrnet: Point cloud registration network using pointnet encoding.
\newblock {\em arXiv preprint arXiv:1908.07906}, 2019.

\bibitem{activesearch}
Torsten Sattler, Bastian Leibe, and Leif Kobbelt.
\newblock Efficient \& effective prioritized matching for large-scale
  image-based localization.
\newblock {\em IEEE Transactions on Pattern Analysis and Machine Intelligence},
  39(9):1744--1756, 2016.

\bibitem{sattler2019understanding}
Torsten Sattler, Qunjie Zhou, Marc Pollefeys, and Laura Leal-Taixe.
\newblock Understanding the limitations of cnn-based absolute camera pose
  regression.
\newblock In {\em IEEE Conference on Computer Vision and Pattern Recognition},
  2019.

\bibitem{scenecoordforest}
Jamie Shotton, Ben Glocker, Christopher Zach, Shahram Izadi, Antonio Criminisi,
  and Andrew Fitzgibbon.
\newblock Scene coordinate regression forests for camera relocalization in
  rgb-d images.
\newblock In {\em IEEE Conference on Computer Vision and Pattern Recognition},
  2013.

\bibitem{cnnslam}
Keisuke Tateno, Federico Tombari, Iro Laina, and Nassir Navab.
\newblock Cnn-slam: Real-time dense monocular slam with learned depth
  prediction.
\newblock In {\em IEEE Conference on Computer Vision and Pattern Recognition},
  2017.

\bibitem{densevlad}
A. {Torii}, R. {Arandjelović}, J. {Sivic}, M. {Okutomi}, and T. {Pajdla}.
\newblock 24/7 place recognition by view synthesis.
\newblock In {\em IEEE Conference on Computer Vision and Pattern Recognition},
  2015.

\bibitem{tosi2019learning}
Fabio Tosi, Filippo Aleotti, Matteo Poggi, and Stefano Mattoccia.
\newblock Learning monocular depth estimation infusing traditional stereo
  knowledge.
\newblock In {\em IEEE Conference on Computer Vision and Pattern Recognition},
  2019.

\bibitem{12scenes}
Julien Valentin, Angela Dai, Matthias Nie{\ss}ner, Pushmeet Kohli, Philip Torr,
  Shahram Izadi, and Cem Keskin.
\newblock Learning to navigate the energy landscape.
\newblock In {\em International Conference on 3D Vision}, 2016.

\bibitem{lstmposenet}
Florian Walch, Caner Hazirbas, Laura Leal-Taixe, Torsten Sattler, Sebastian
  Hilsenbeck, and Daniel Cremers.
\newblock Image-based localization using lstms for structured feature
  correlation.
\newblock In {\em International Conference on Computer Vision}, 2017.

\bibitem{deepclosestpoint}
Yue Wang and Justin~M. Solomon.
\newblock Deep closest point: Learning representations for point cloud
  registration.
\newblock In {\em International Conference on Computer Vision}, 2019.

\bibitem{branchnet}
Jian Wu, Liwei Ma, and Xiaolin Hu.
\newblock Delving deeper into convolutional neural networks for camera
  relocalization.
\newblock In {\em IEEE International Conference on Robotics and Automation},
  2017.

\bibitem{localglobal}
Fei Xue, Xin Wang, Zike Yan, Qiuyuan Wang, Junqiu Wang, and Hongbin Zha.
\newblock Local supports global: Deep camera relocalization with sequence
  enhancement.
\newblock In {\em International Conference on Computer Vision}, 2019.

\bibitem{sanet}
Luwei Yang, Ziqian Bai, Chengzhou Tang, Honghua Li, Yasutaka Furukawa, and Ping
  Tan.
\newblock Sanet: Scene agnostic network for camera localization.
\newblock In {\em International Conference on Computer Vision}, 2019.

\bibitem{dvso}
Nan Yang, Rui Wang, Jorg Stuckler, and Daniel Cremers.
\newblock Deep virtual stereo odometry: Leveraging deep depth prediction for
  monocular direct sparse odometry.
\newblock In {\em European Conference on Computer Vision}, 2018.

\end{thebibliography}
}

\null
\vskip .375in
\twocolumn[{%
  \begin{center}
    \textbf{\Large Supplemental Material : \\ A Structure-Aware Method for Direct Pose Estimation}
  \end{center}
  \vspace*{24pt}
}]

\makeatletter

\newcommand{\tic}[1]{\includegraphics[width=25mm]{images/maskcomparisons/#1}}\textbf{}

\begin{figure*}
  \includegraphics[trim=160 10 140 30,clip,width=\linewidth]{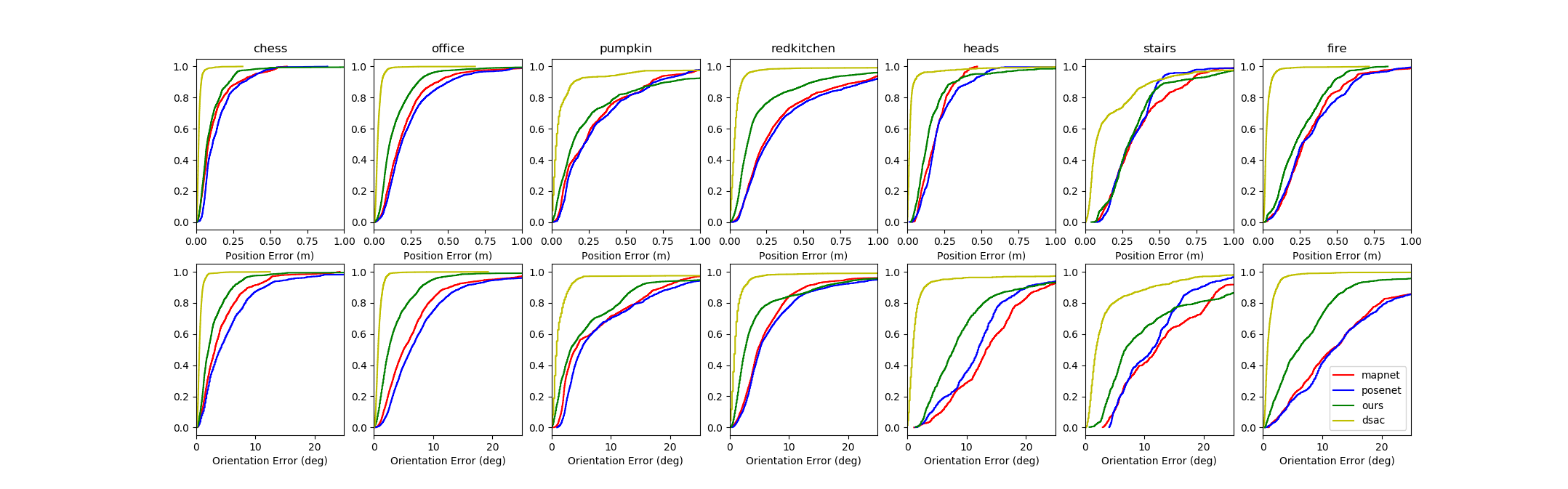}
  \caption{Individual cumulative histograms of error for each scene from 7Scenes for several methods, truncated to 1 meter and 25 degrees error.}
  
  \label{fig:histograms}
\end{figure*}

\begin{figure*}
  \centering
  \begin{tabular}{c|cc|cc}
    & \multicolumn{2}{c|}{Training With Mask} & \multicolumn{2}{c}{Training Without Mask} \\
    Input & Depth & SC &  Depth & SC \\ 

  \tic{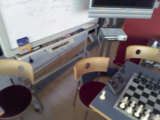} & \tic{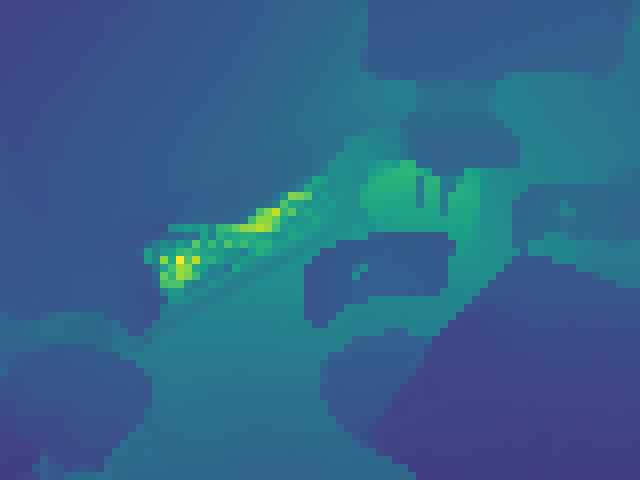} & \tic{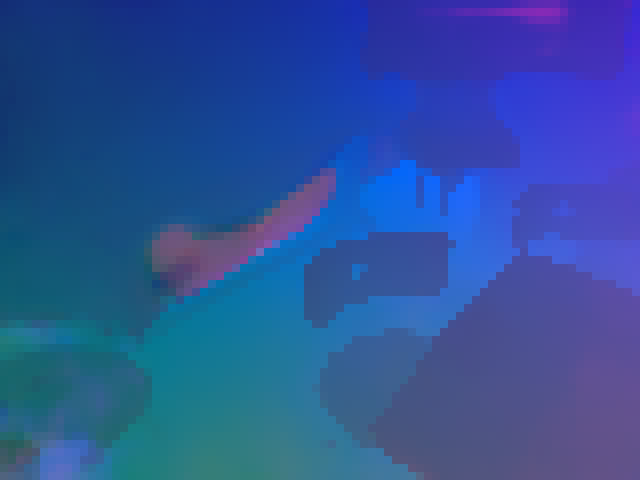} & \tic{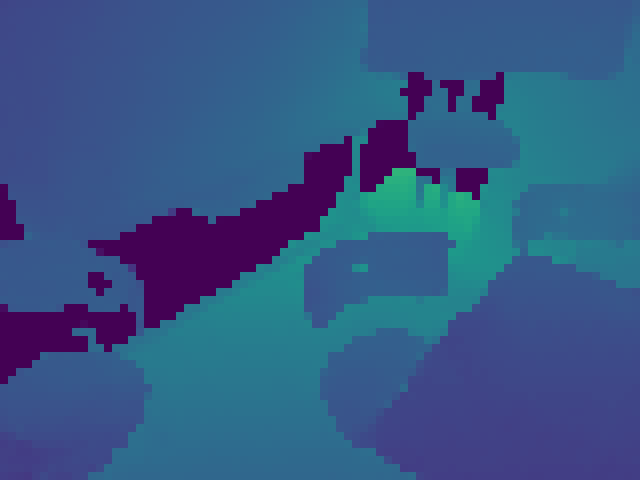} & \tic{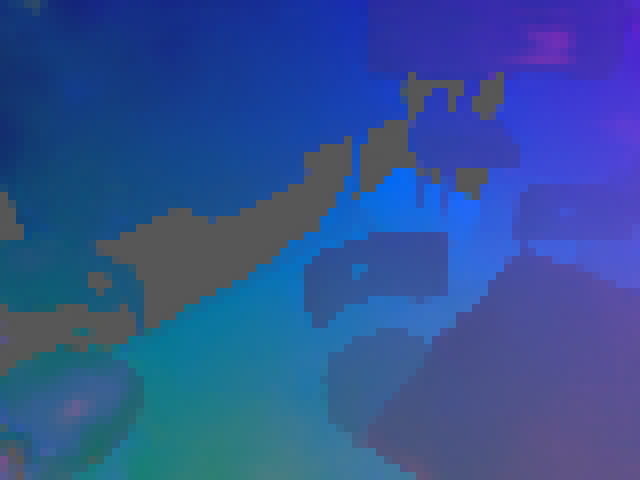}\\
  \tic{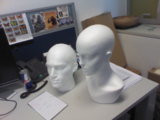} & \tic{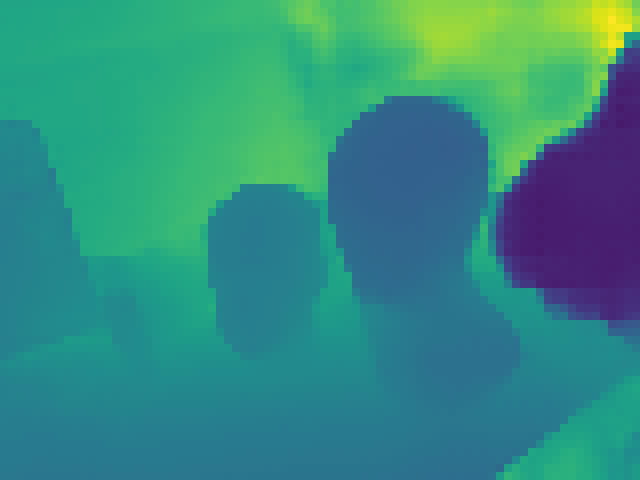} & \tic{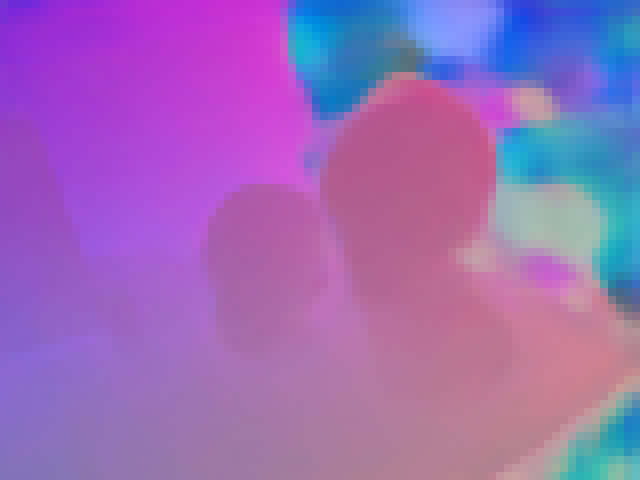}  & \tic{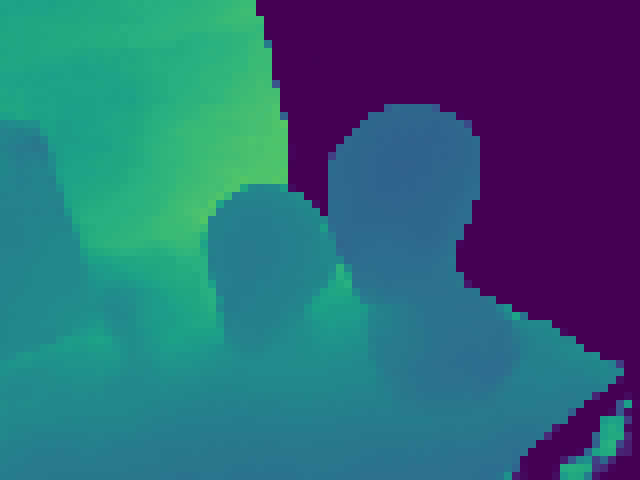} & \tic{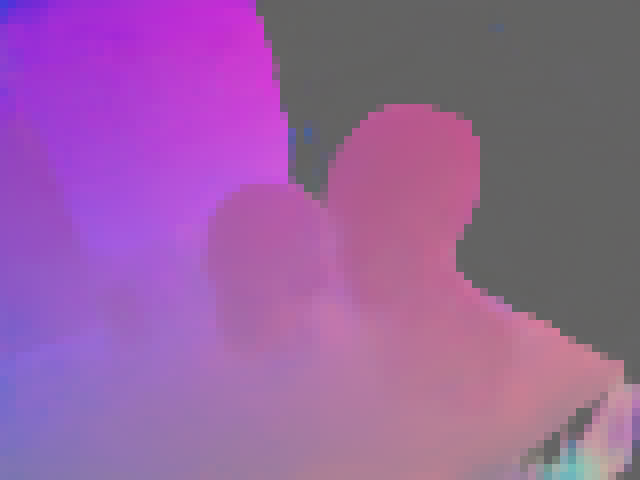}  \\ 

  \end{tabular}
  \caption{Qualitative results from training with or without a validity mask: (left) input image, (middle) output depth and scene coordinates (SC) trained using the validity mask, and (right) outputs trained without this mask. Note that these are not ground-truth labels. Training without the mask forces the network to recognize unknown areas.}
  \label{fig:maskornomask}
\end{figure*}

\section{Additional Results on Indoor Scenes}

We show per-scene histograms of error for the 7Scenes dataset in \figref{histograms}. We also show the result of training with or without the validity mask in $L_{geom}$ in \figref{maskornomask}.

We show qualitative results on images from the 12Scenes dataset in \figref{12scenes1} and \figref{12scenes2}. Compared to the 7Scenes dataset, it is less common for images to contain large regions that have no depth information during training. As such, the weighting CNN produces more varied weights in general because it can no longer depend on obviously incorrect points. This can result in seemingly strange results, such as row 5 in \figref{12scenes1} and row 12 in \figref{12scenes2}. Note, however, that even with these strange correspondence weights, the final re-projection error is typically low.

\section{Results on Outdoor Scenes}

We show quantitative results on common outdoor scenes from the Cambridge Landmarks dataset~\cite{posenet} in \tblref{landmarks}. While our approach does not work as well in this scenario as it does for the indoor scenes from the main paper due to the low quality depth labels, it still performs competitively on all scenes, and is the best method for the Hospital scene by a large margin. On average our method is best for position error, but the ResNet based PoseNet~\cite{posenet2} performs best on orientation error. This is surprising since PoseNet was not competitive even against other similar methods on 7Scenes. This shows the difficulty of this dataset for direct pose estimation methods. Note that this is among the most challenging scenarios for our method because of the poor quality of the depth labels generated from  rendering from sparse structure-from-motion (SfM) keypoints. Due to the explicit nature of our method, utilizing a better SfM tool to generate more accurate depth images will directly lead to better performance. Examples of depth labels found in the dataset are shown in \figref{labels}. While there are many areas that have missing depth (depth=0), this is not an issue for our method as we can ignore these pixels during training. However, there are a large number of sky pixels which are incorrectly labeled with depth, as well as erroneous depth values in general. These errors are an issue for our method because they result in incorrect supervision during training. However, as mentioned earlier, even with these labels our method performs well, and there is a clear path to improvement from better label generation alone.

Due to the poor depth quality and fewer training examples per scene, we use the masked version of $L_{geom}$ and train for more epochs compared to indoor scenes. For the geometry optimization phase, we train for 100 epochs with an initial learning rate of $1e^{-4}$ and reduce the learning rate by a factor of $0.5$ every 40 epochs. For the pose optimization phase, we train for 20 epochs with a learning rate of $1e^{-3}$ on the weighting CNN parameters and $1e^{-4}$ on the depth and scene coordinate CNN parameters. The higher learning rate on the geometry prediction parameters is similar to the the re-projection error optimization phase of DSAC++~\cite{dsac2} due to the error in ground-truth scene coordinate labels.

We show visualizations of network outputs on several Cambridge Landmarks inputs in \figref{examples}. Notice that even in areas where depth and scene coordinate predictions seem good, the predicted weights tend to focus on a smaller area. This is apparent mostly in the Kings College scene, examples of which are shown in rows 2 and 8. Also, in row 6 we can see a difficult case where most of the image is a tree, leading to bad predictions. This is reflected in the weights as all predicted correspondence weights for this example are very low. Overall, even with the noisy depth labels, the weighting mechanism is able to capture which points are more reliable for final pose computation.

\section{Depth Accuracy}

A key part of our method is the choice of depth estimation network. While we could have chosen a large network and trained it for generic depth estimation, we instead chose a more shallow network and trained on a per-scene basis for pose estimation. \tblref{depth_error} shows the average depth error for each scene. We compare the depth estimation accuracy in the case of 1) training with a single scene, 2) training with all scenes, and 3) holding out the scene. This last case tests the potential ability of our depth networks to transfer to other scenes. We report mean absolute error, as well as depth accuracy values for different error thresholds. As expected, we typically observe a gradual decline in depth estimation quality as we move from the single scene case where much of the scene structure can be memorized, to the held out case, where no information about the scene was observed during training. Also, we show that a high percentage of pixels have a depth error of less than 0.125 meters, so we believe our simple network is sufficient for this task.

\section{Architecture Details}

We use a standard ResNet-34~\cite{resnet} as the backbone feature extractor for our network. We use the implementation from the Pytorch~\cite{pytorch} torchvision library. The ResNet feature extractor has 5 major components: a single convolution and max pooling (ConvBlock) followed by 4 residual blocks (ResBlock). The general ResNet architecture is given in \tblref{resnet}. We only use the convolutional and residual layers, not the final linear layer.

We provide detailed architecture components for the other components of our network, namely the scene coordinate regression network (\tblref{scenecoords}), the depth network (\tblref{depth}), and the weighting network (\tblref{weighting}). These networks make use of intermediate outputs from the shared ResNet backbone which are given in the ``Input'' column. For convolutions, we list it as Conv2d(in channels, out channels, kernel size, stride, padding).

\clearpage

\newcommand{\figurescenes}[1]{\includegraphics[width=28mm]{images/12scenes/#1}}

\begin{figure*}
  \centering
  \begin{tabular}{cc|ccc|c}
  &  & \multicolumn{3}{c|}{Network Outputs} & Computed \\
 &    Input &  Depth &  Scene Coords &  Weights & Endpoint Error\\ 
   1&  \figurescenes{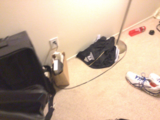} & \figurescenes{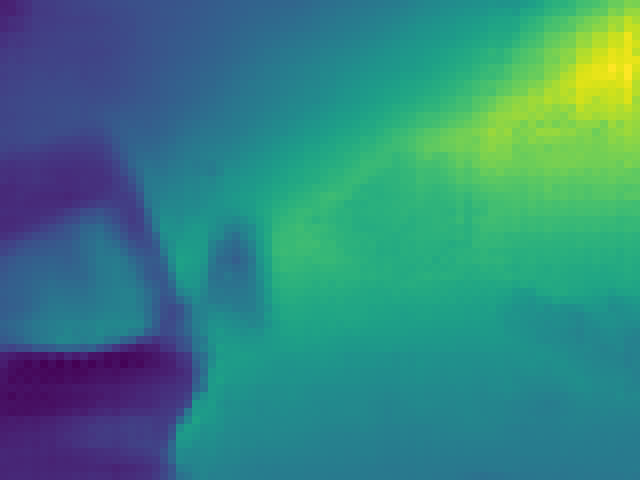} & \figurescenes{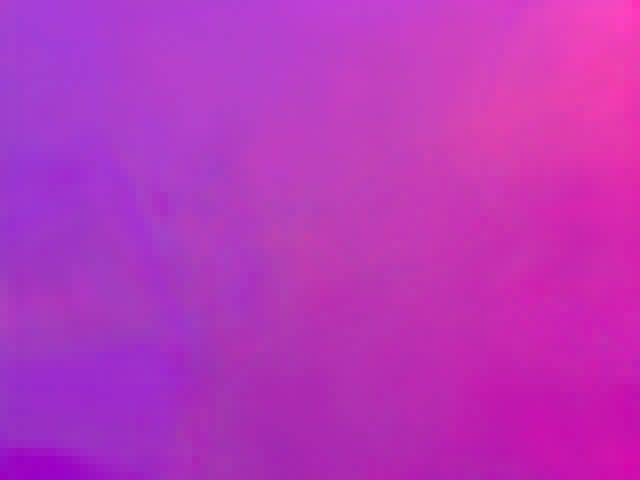} & \figurescenes{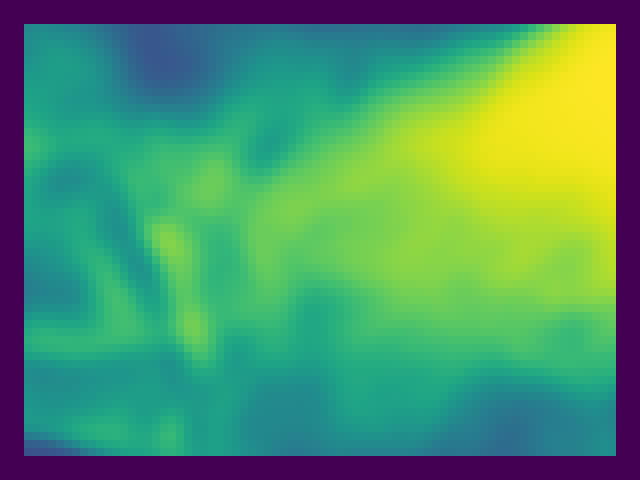} & \figurescenes{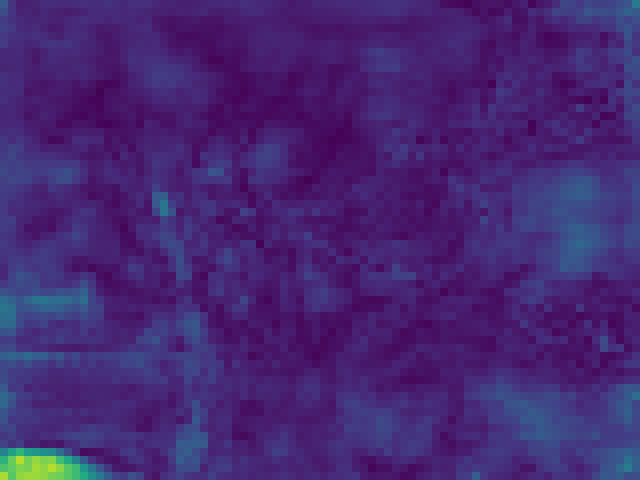}  \\
   2&  \figurescenes{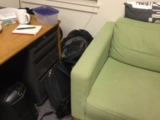} & \figurescenes{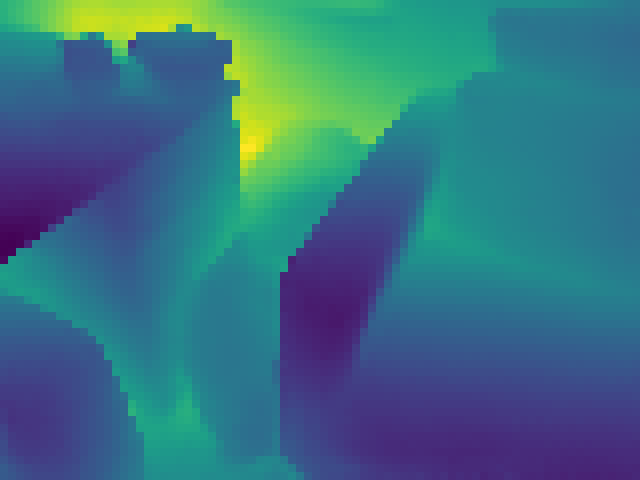} & \figurescenes{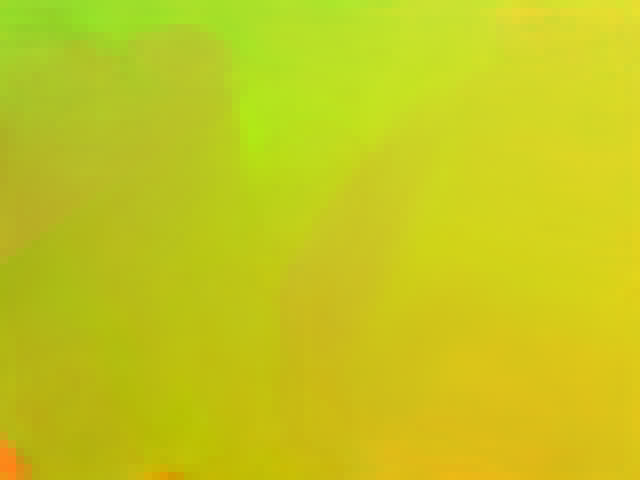} & \figurescenes{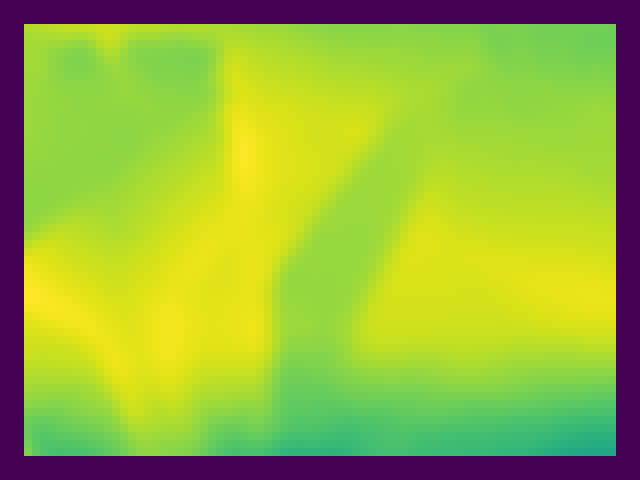} & \figurescenes{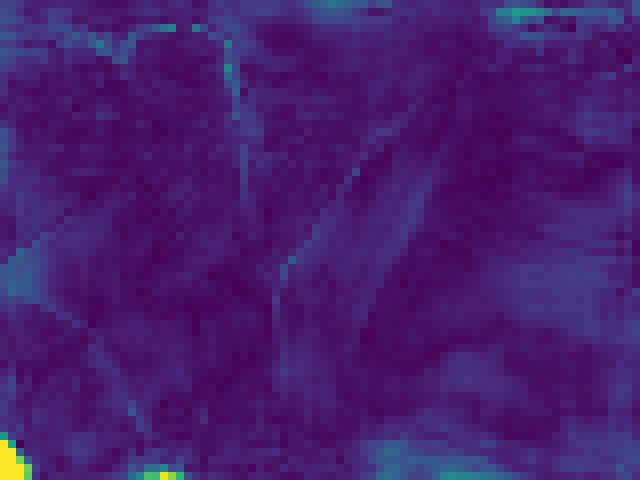}  \\
  3&  \figurescenes{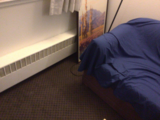} & \figurescenes{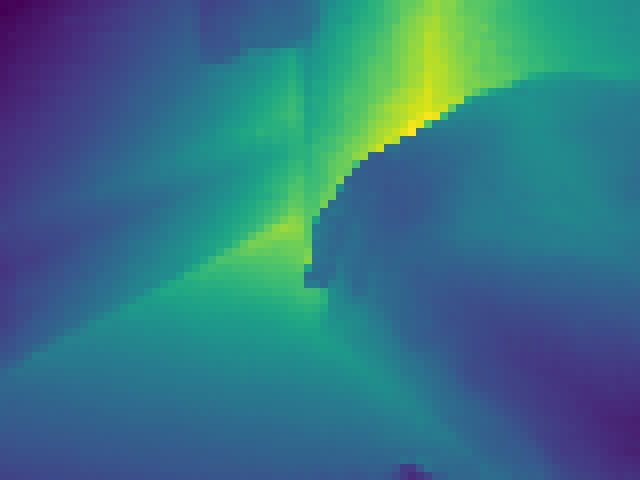} & \figurescenes{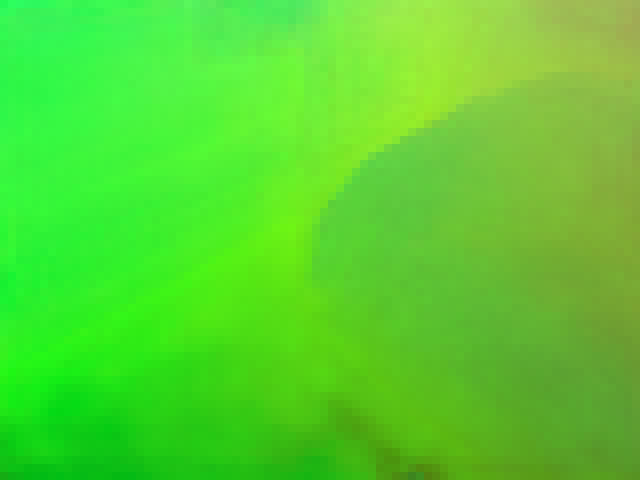} & \figurescenes{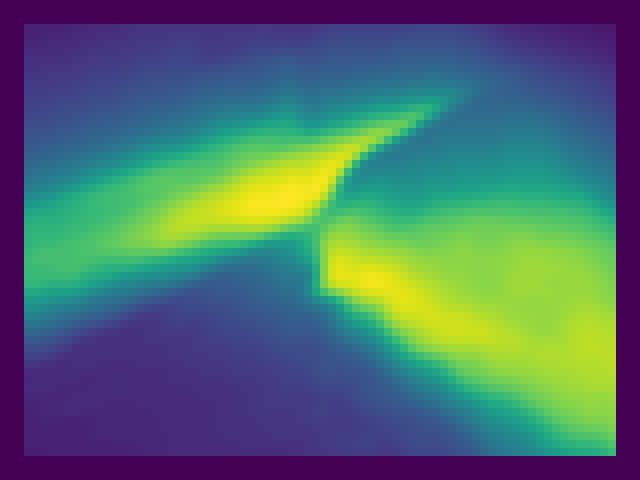} & \figurescenes{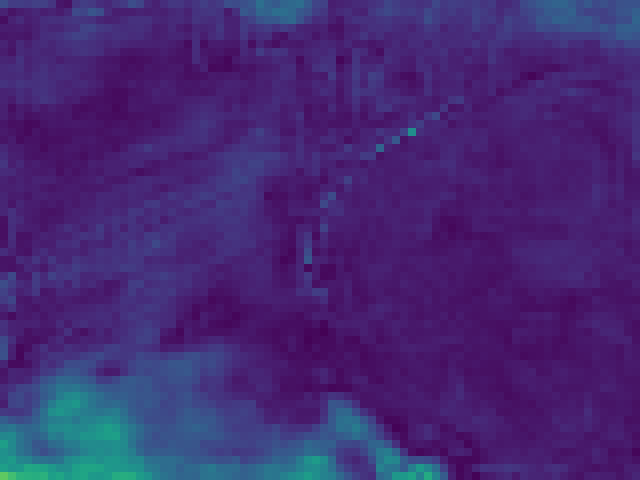}  \\
  4&  \figurescenes{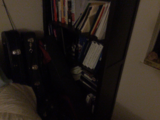} & \figurescenes{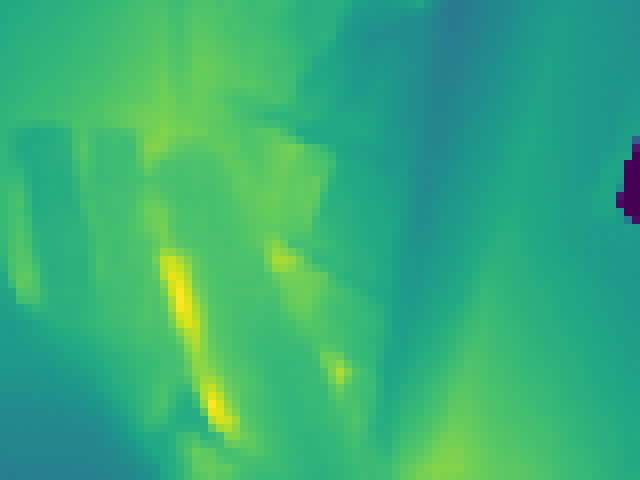} & \figurescenes{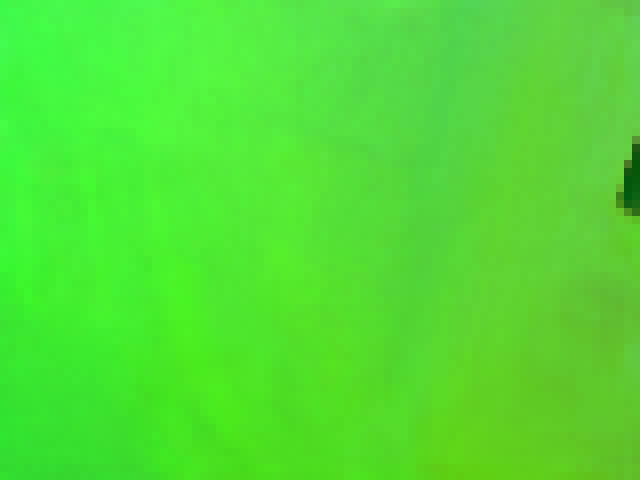} & \figurescenes{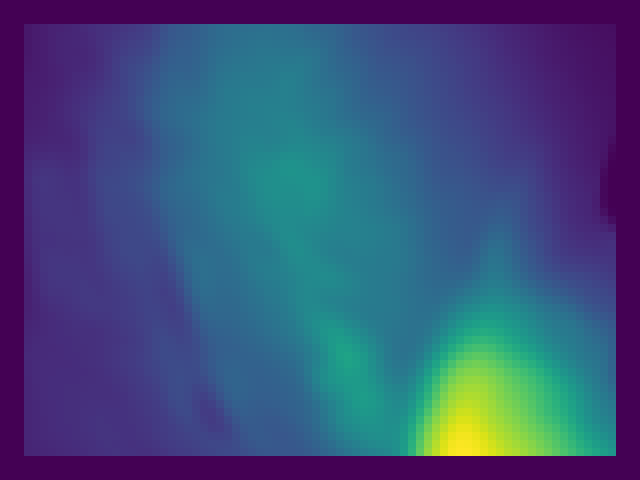} & \figurescenes{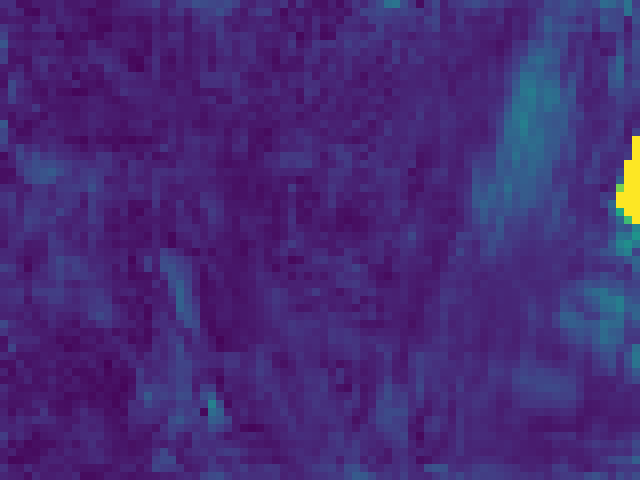}  \\
  5&  \figurescenes{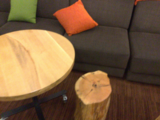} & \figurescenes{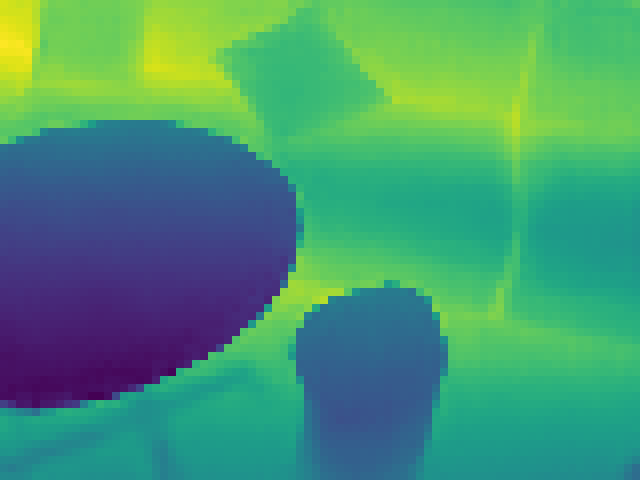} & \figurescenes{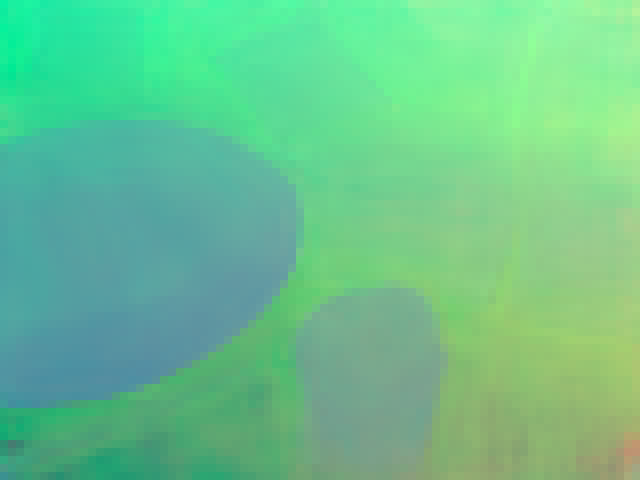} & \figurescenes{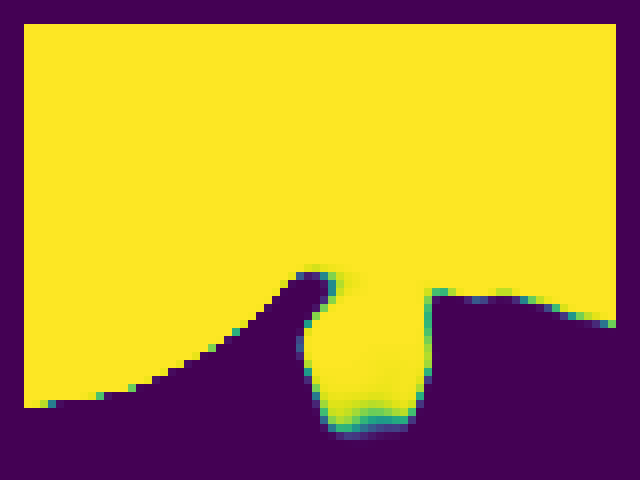} & \figurescenes{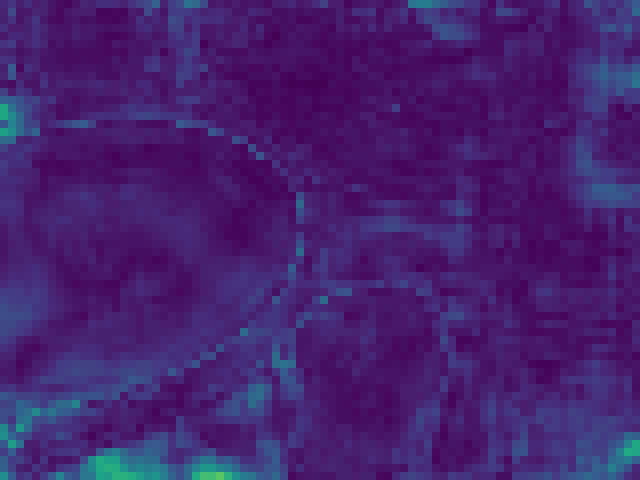}  \\
  6&  \figurescenes{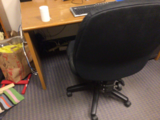} & \figurescenes{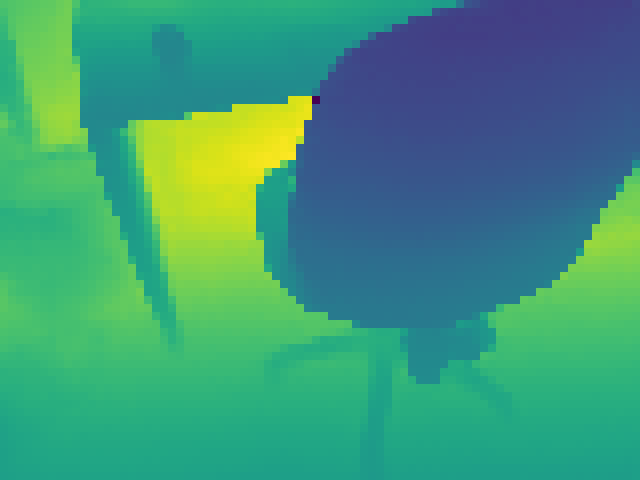} & \figurescenes{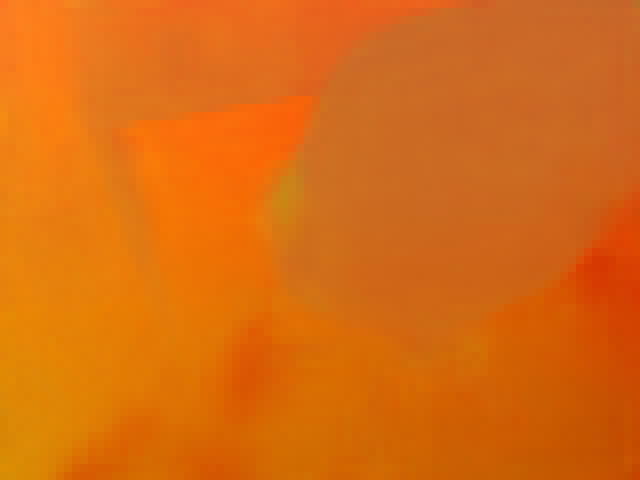} & \figurescenes{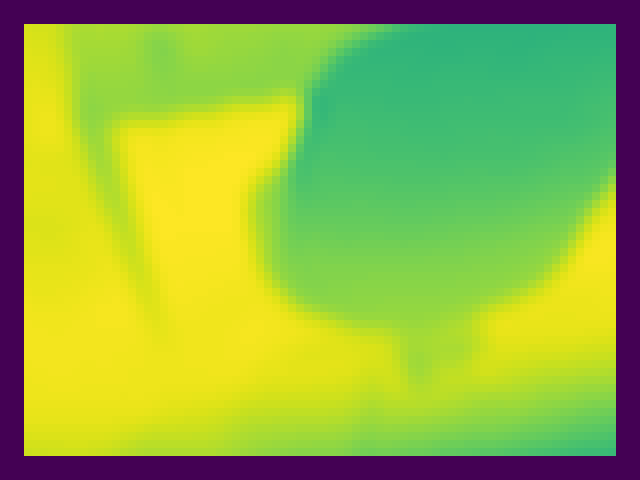} & \figurescenes{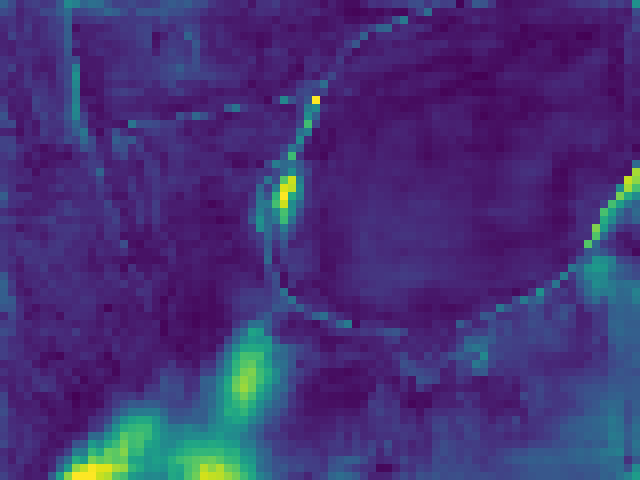}  \\
  7&  \figurescenes{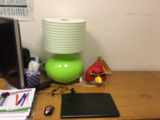} & \figurescenes{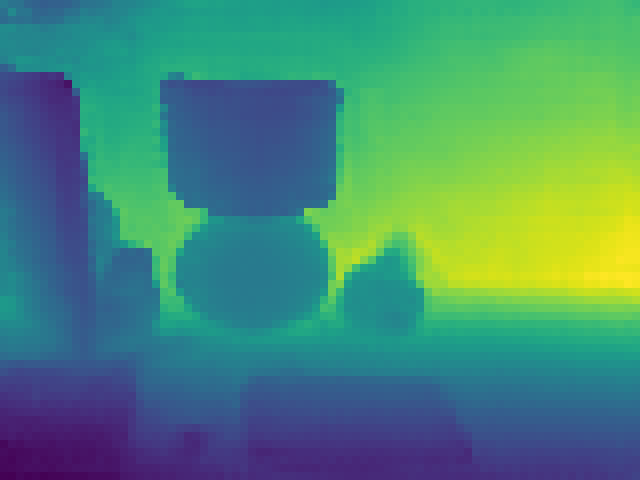} & \figurescenes{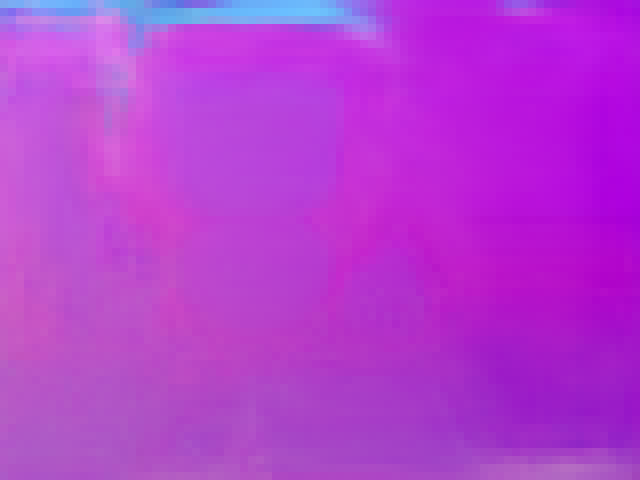} & \figurescenes{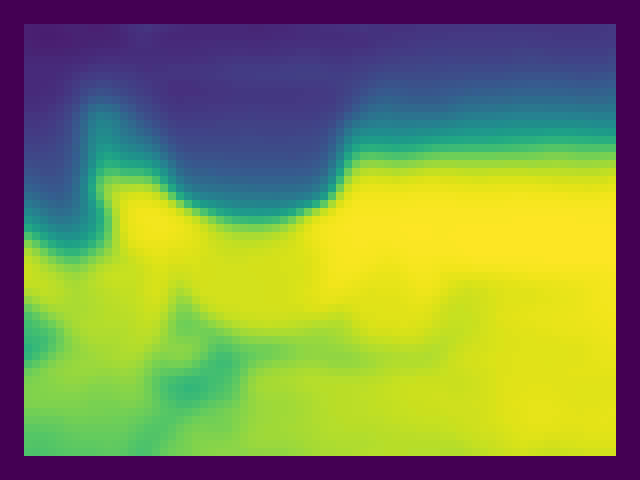} & \figurescenes{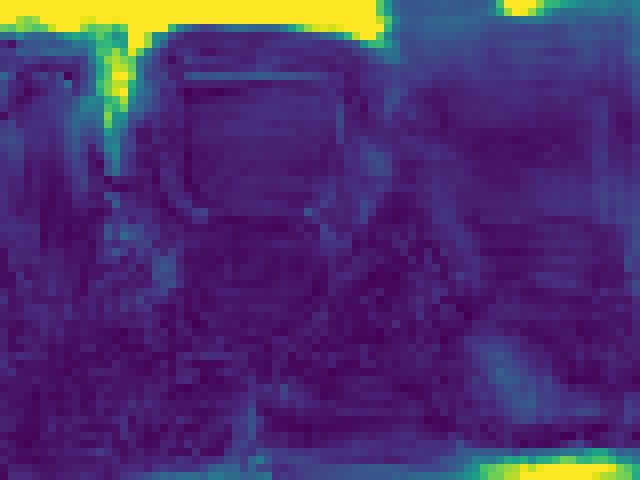}  \\
  8&  \figurescenes{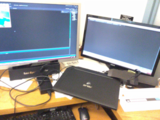} & \figurescenes{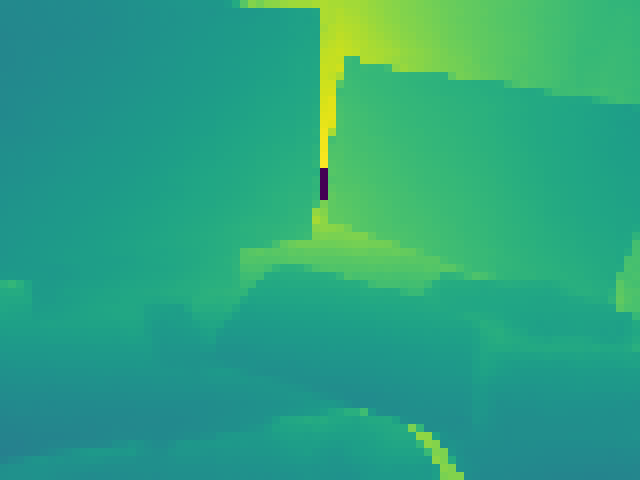} & \figurescenes{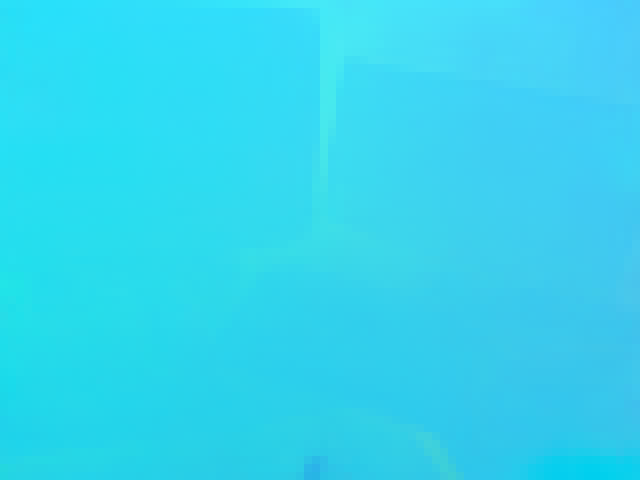} & \figurescenes{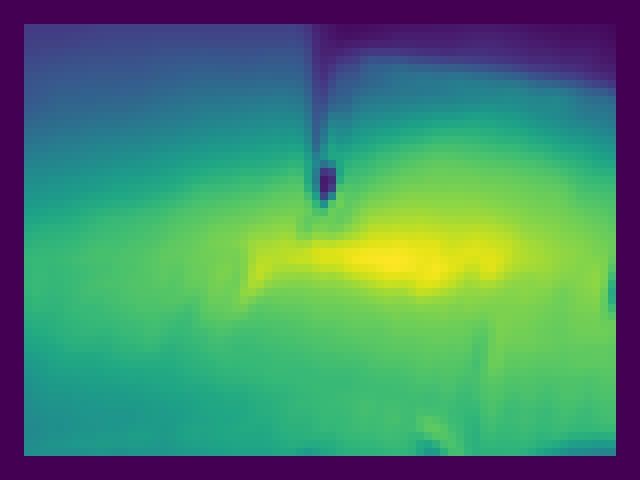} & \figurescenes{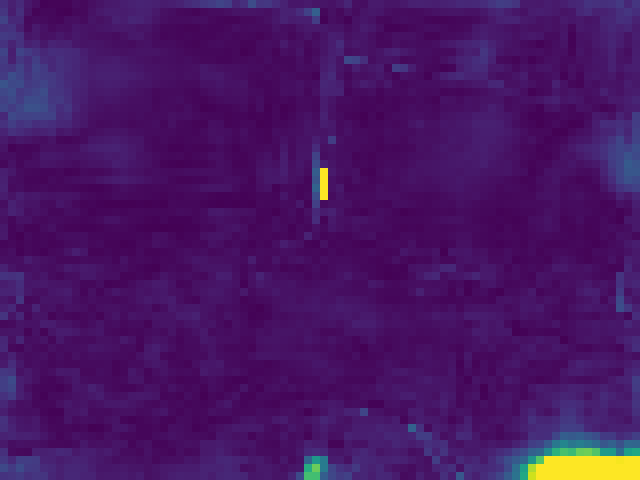}  \\
  9&  \figurescenes{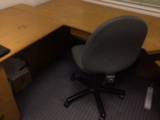} & \figurescenes{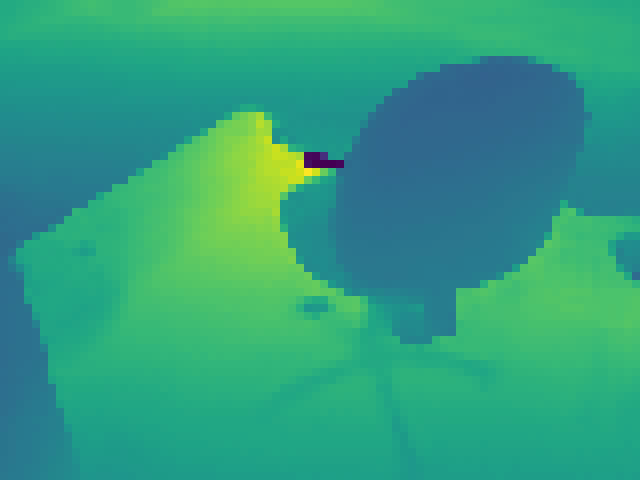} & \figurescenes{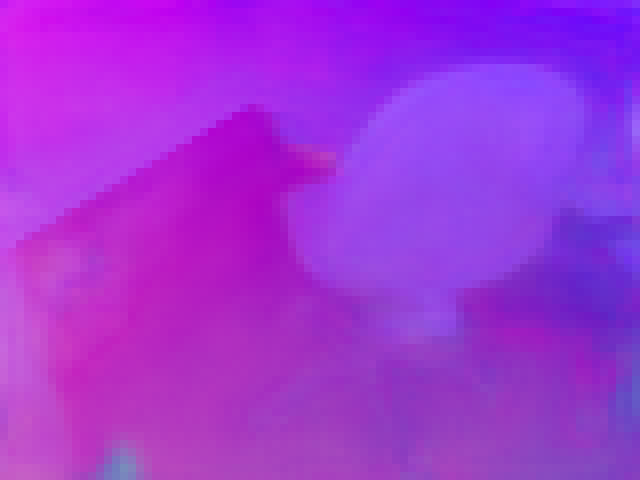} & \figurescenes{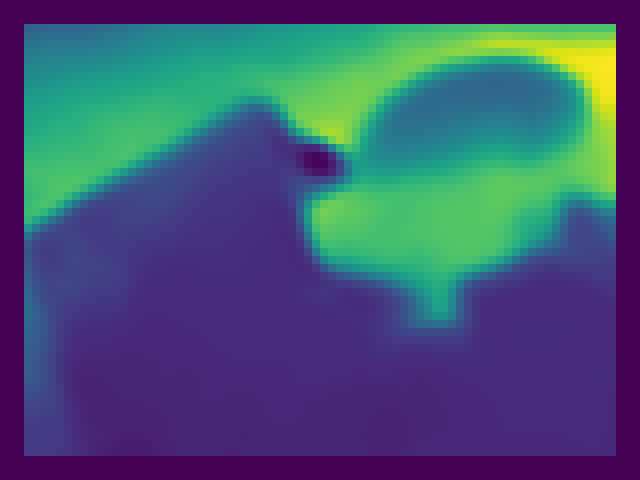} & \figurescenes{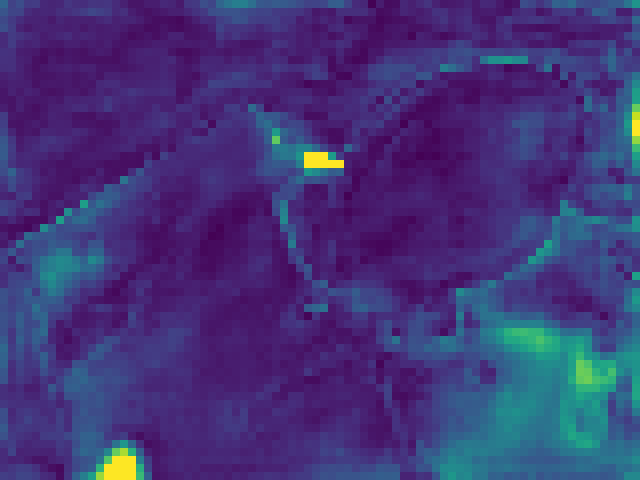}  \\

  \end{tabular}
  \caption{Example network outputs for 12Scenes images. Depth, scene coordinates, and weights are direct outputs of our network. The endpoint error is computed by applying the regressed pose to the scene coordinates and clamped to 1 for visualization. For depth, weights, and endpoint error, brighter means a higher value. Row labels are shown on the left for referencing in text.}
  \label{fig:12scenes1}
\end{figure*}

\begin{figure*}
  \centering
  \begin{tabular}{cc|ccc|c}
  &  & \multicolumn{3}{c|}{Network Outputs} & Computed \\
 &    Input &  Depth &  Scene Coords &  Weights & Endpoint Error\\ 
  10&  \figurescenes{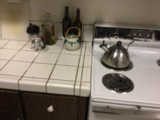} & \figurescenes{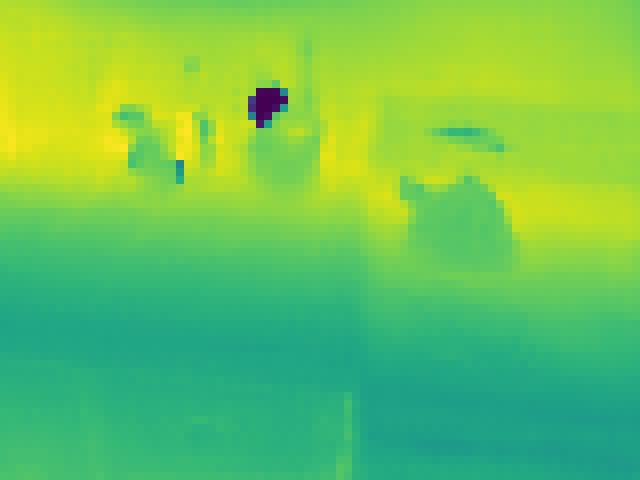} & \figurescenes{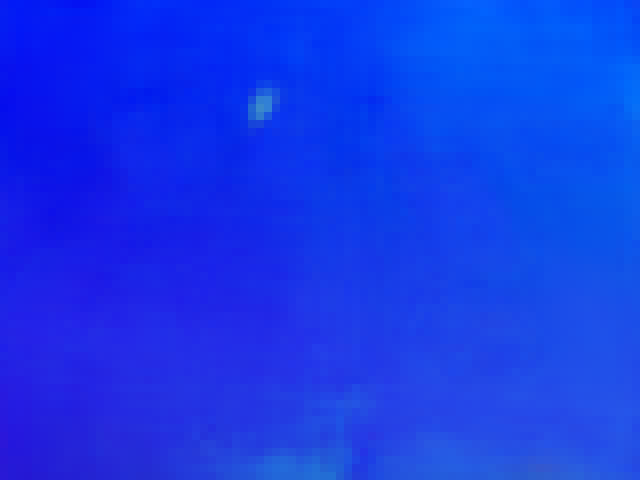} & \figurescenes{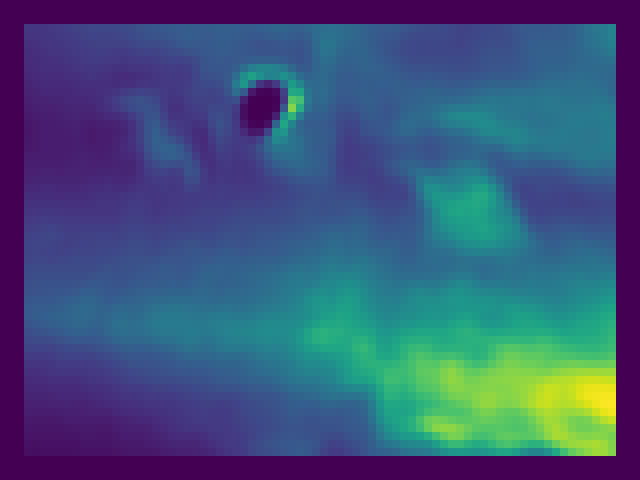} & \figurescenes{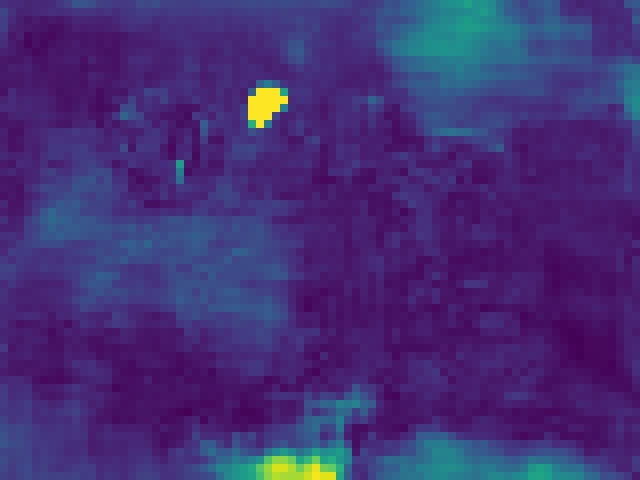}  \\
  11&  \figurescenes{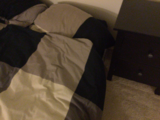} & \figurescenes{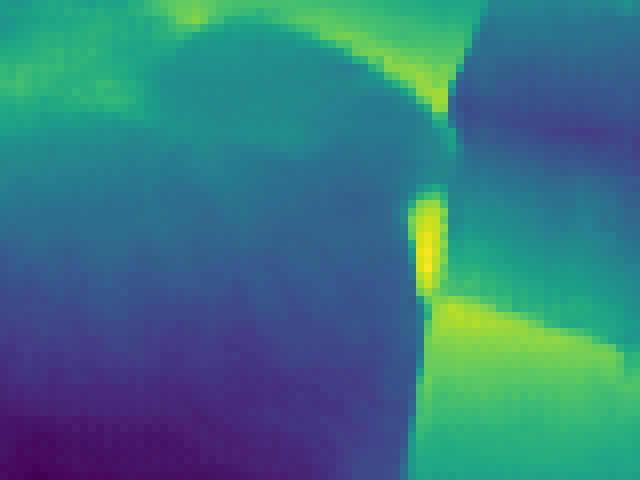} & \figurescenes{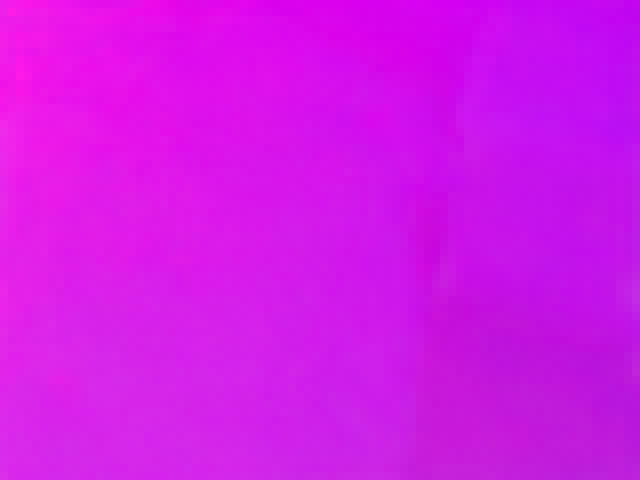} & \figurescenes{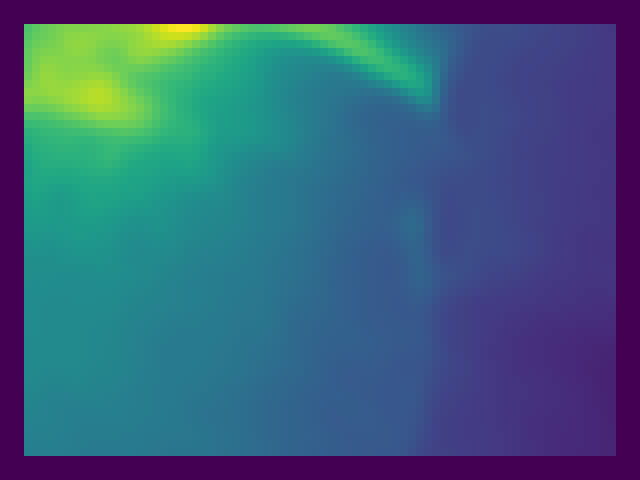} & \figurescenes{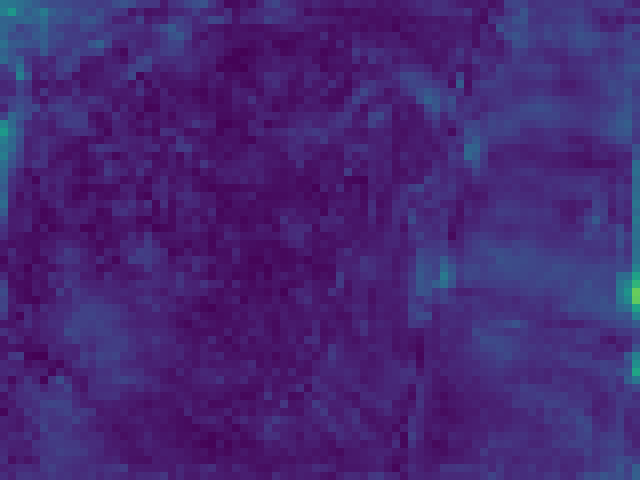}  \\
  12&  \figurescenes{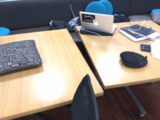} & \figurescenes{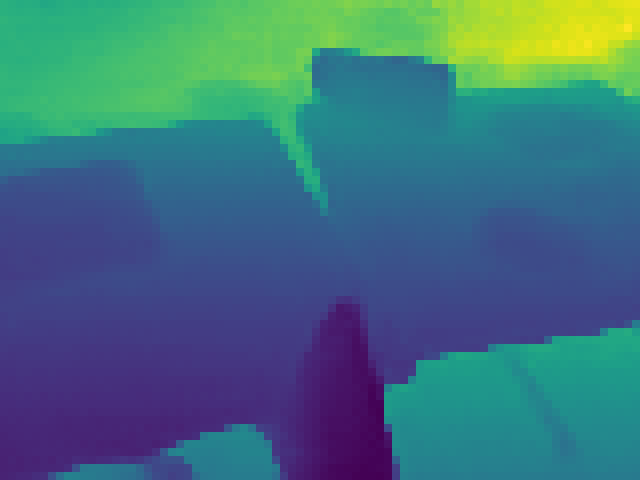} & \figurescenes{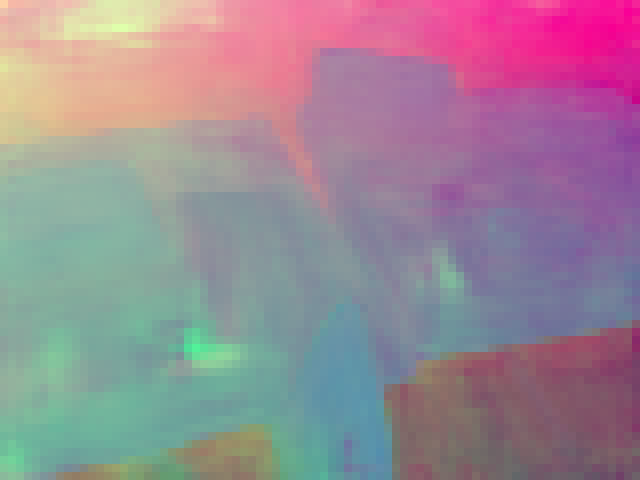} & \figurescenes{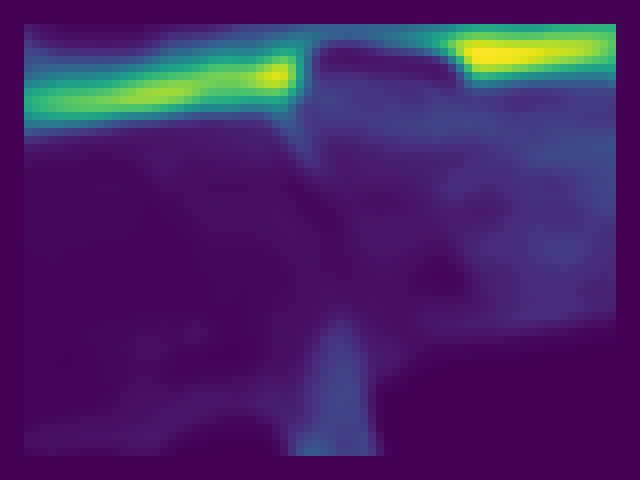} & \figurescenes{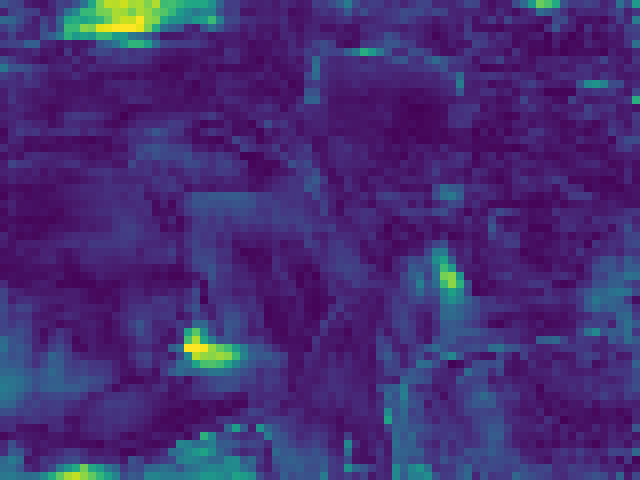}  \\
  13&  \figurescenes{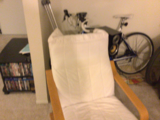} & \figurescenes{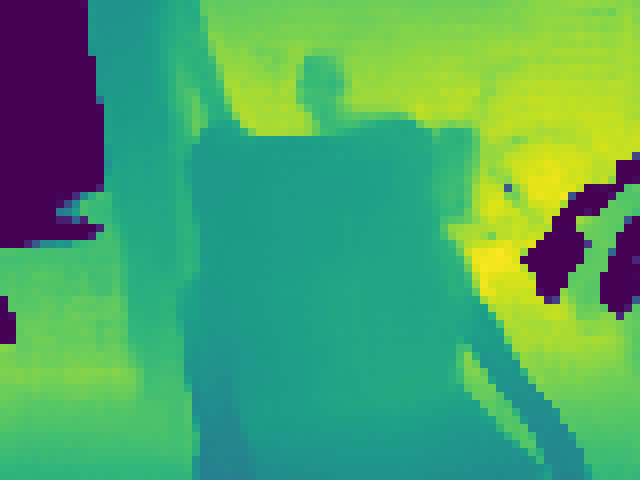} & \figurescenes{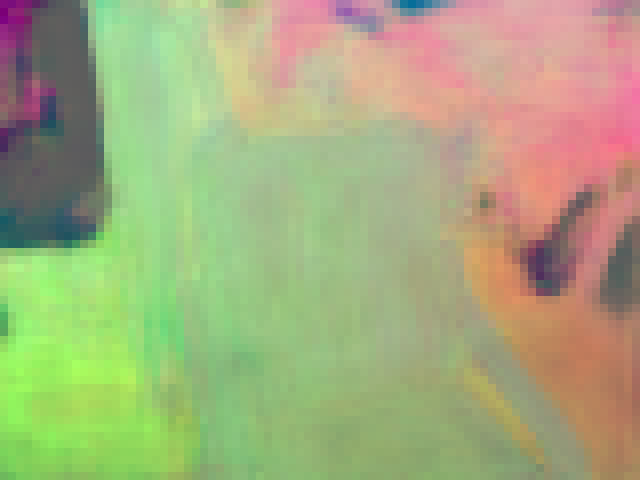} & \figurescenes{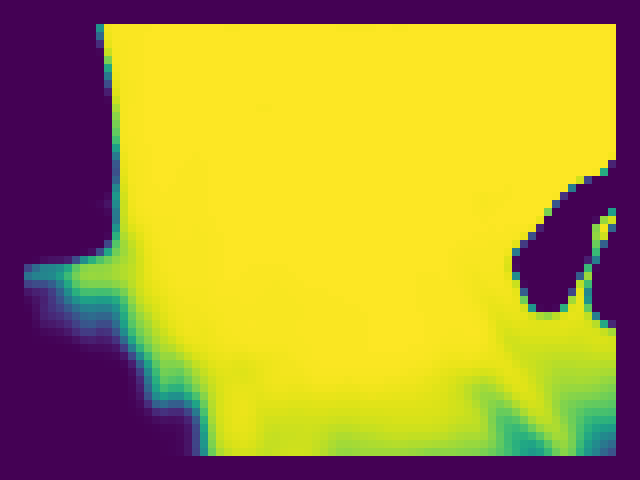} & \figurescenes{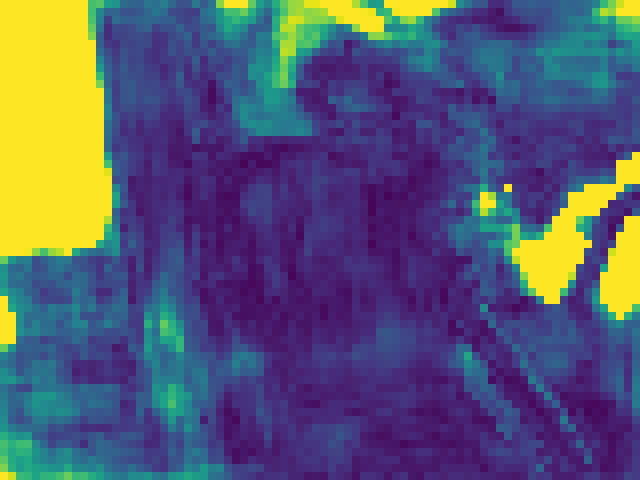}  \\
  14&  \figurescenes{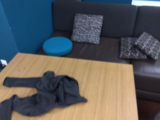} & \figurescenes{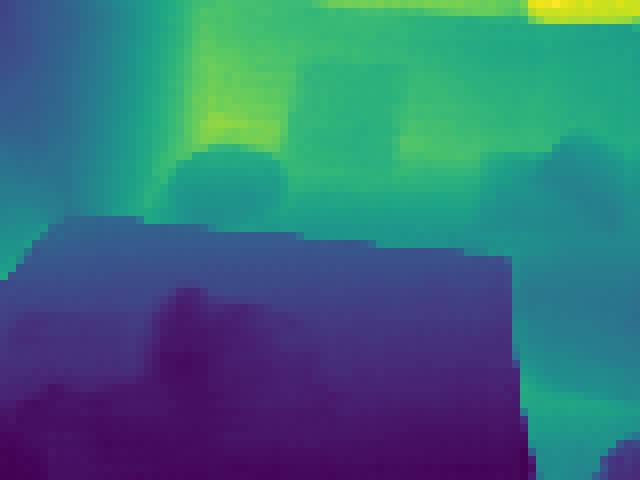} & \figurescenes{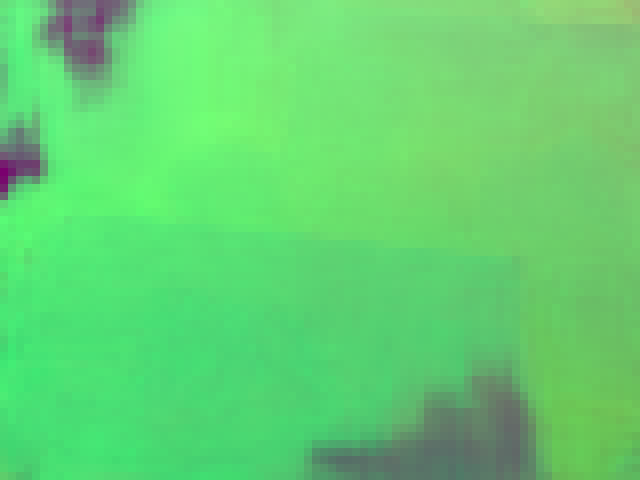} & \figurescenes{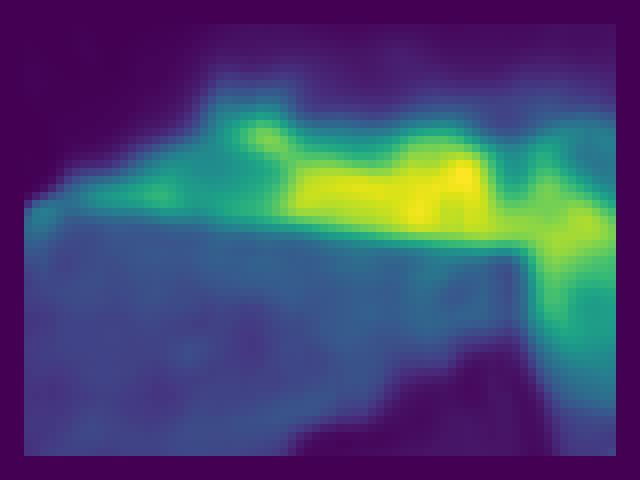} & \figurescenes{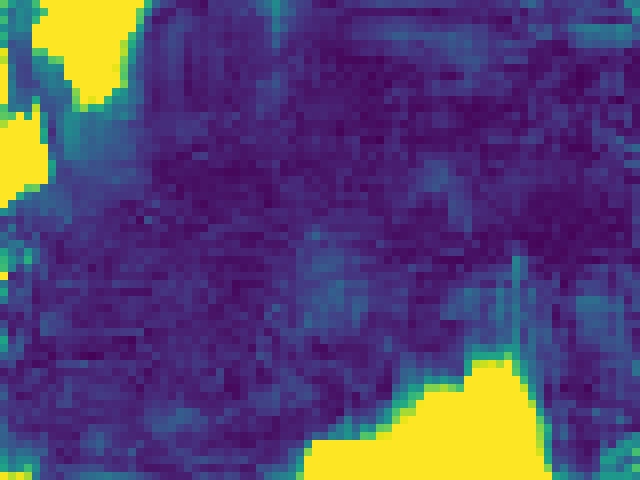}  \\
  15&  \figurescenes{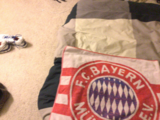} & \figurescenes{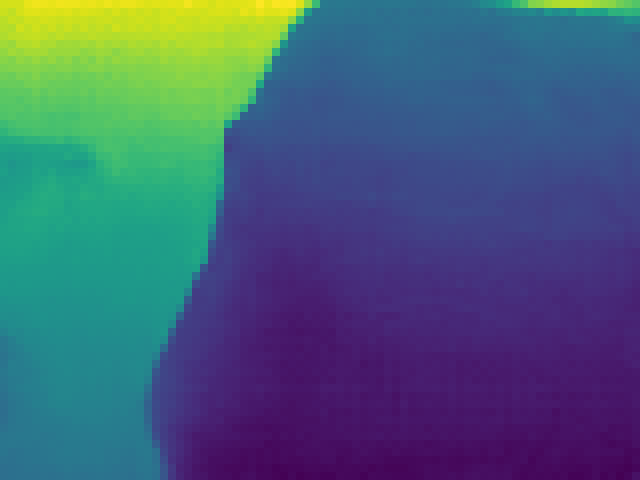} & \figurescenes{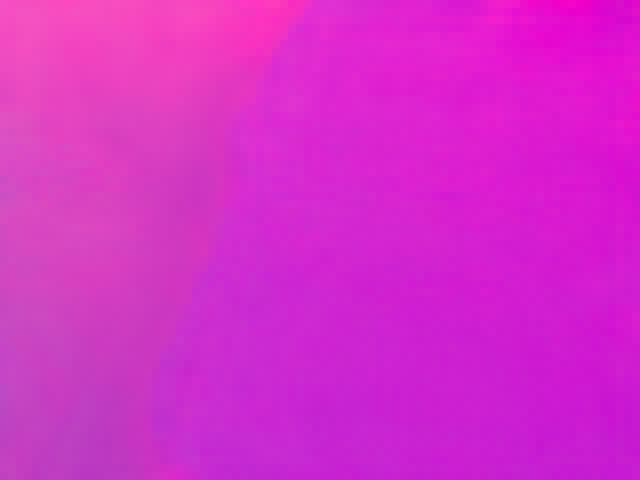} & \figurescenes{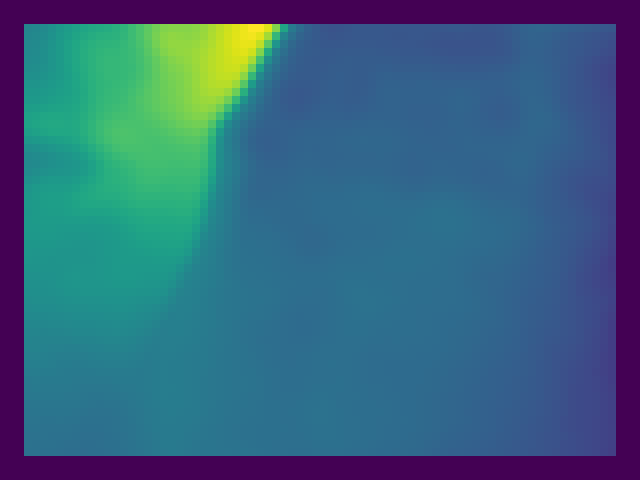} & \figurescenes{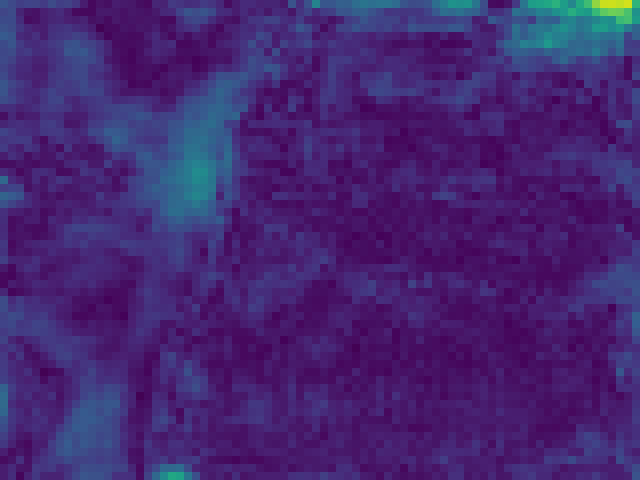}  \\
  16&  \figurescenes{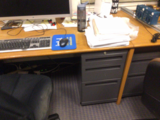} & \figurescenes{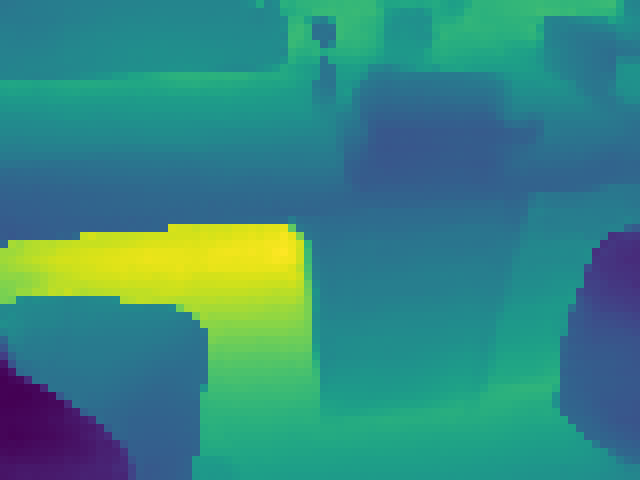} & \figurescenes{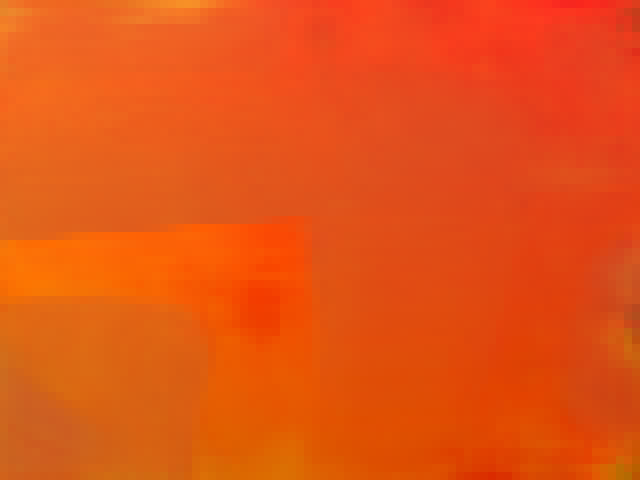} & \figurescenes{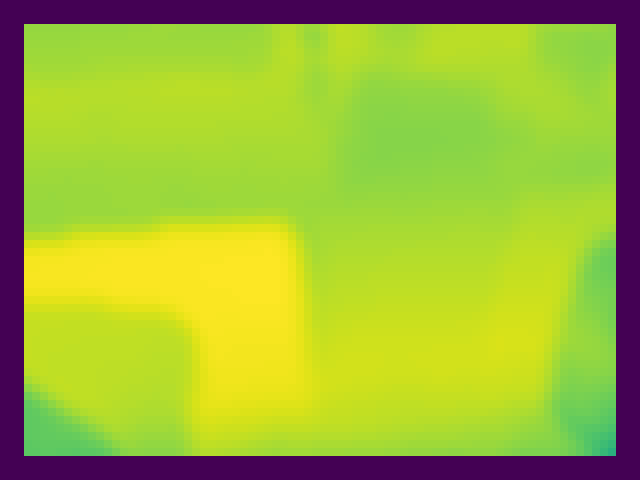} & \figurescenes{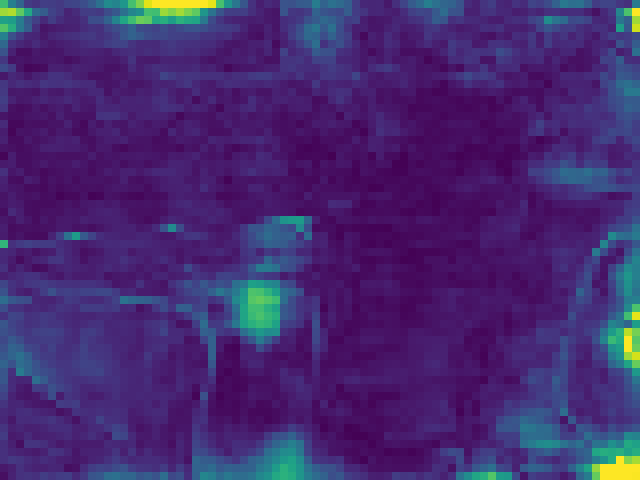}  \\
  17&  \figurescenes{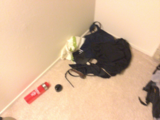} & \figurescenes{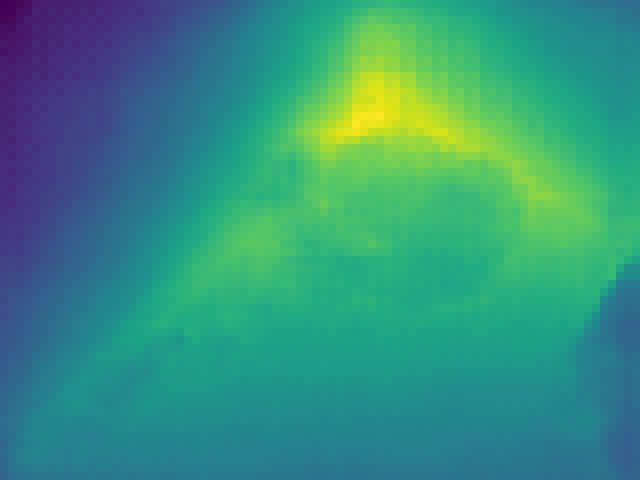} & \figurescenes{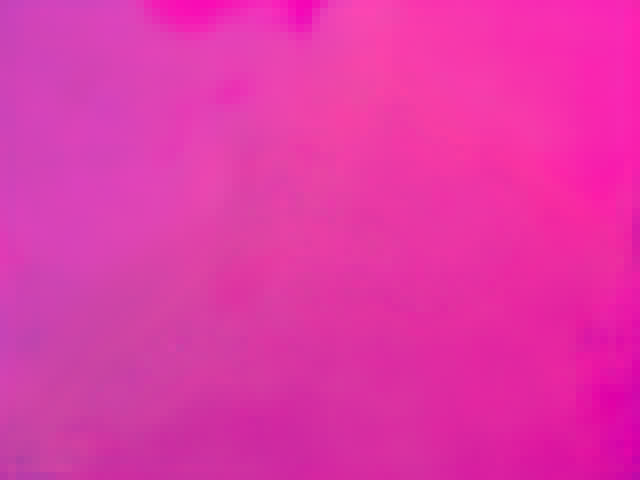} & \figurescenes{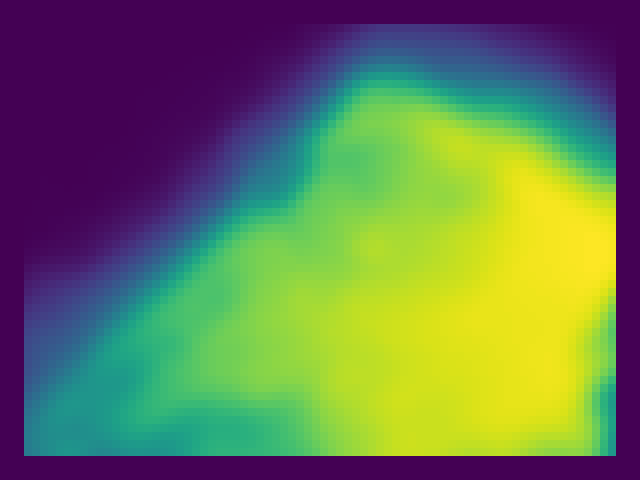} & \figurescenes{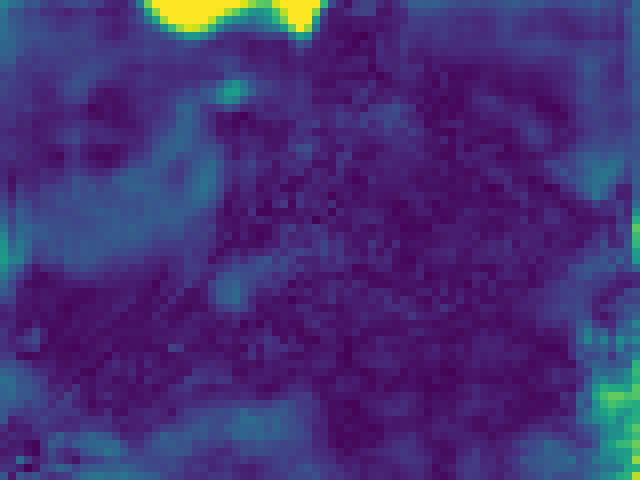}  \\
  18&  \figurescenes{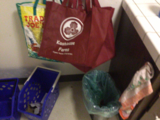} & \figurescenes{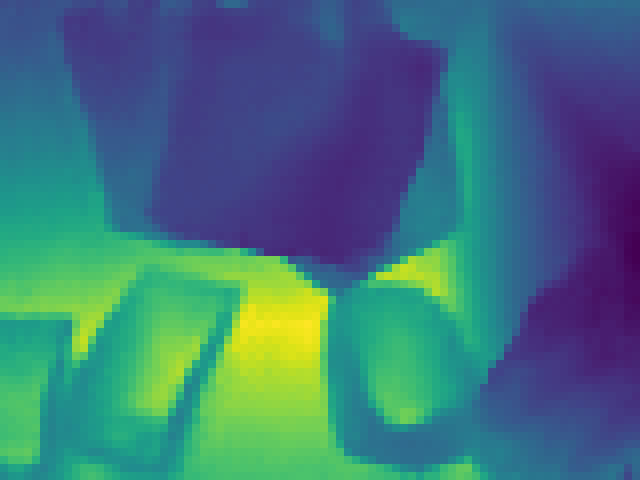} & \figurescenes{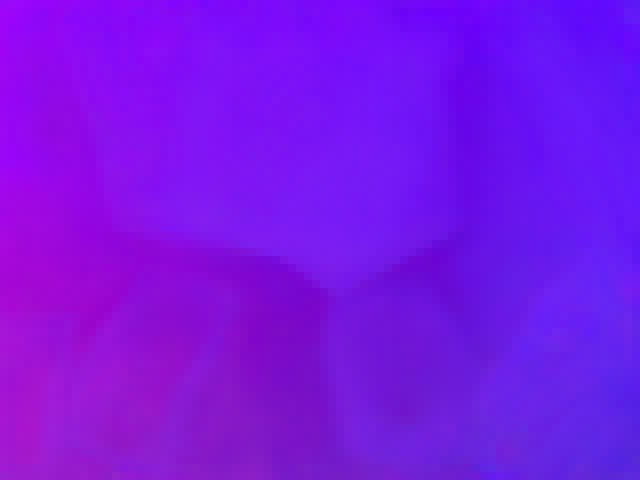} & \figurescenes{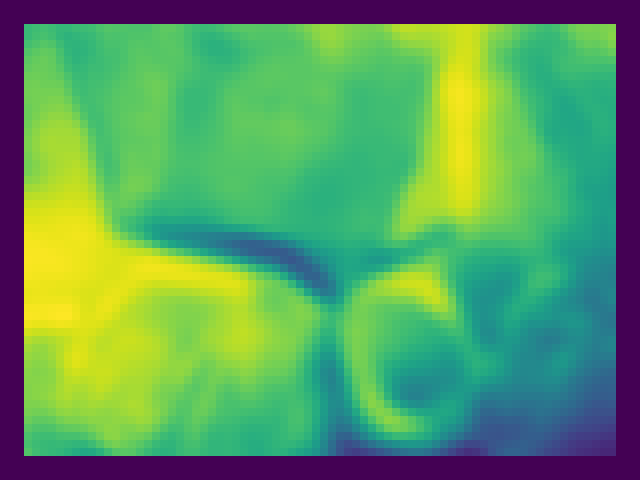} & \figurescenes{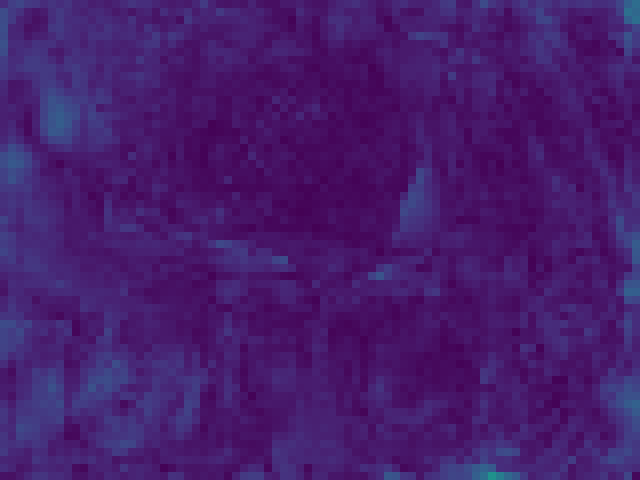}  \\

  \end{tabular}
  \caption{More example network outputs for 12Scenes images. Depth, scene coordinates, and weights are direct outputs of our network. The endpoint error is computed by applying the regressed pose to the scene coordinates and clamped to 1 for visualization. For depth, weights, and endpoint error, brighter means a higher value. Row labels are shown on the left for referencing in text.}
  \label{fig:12scenes2}
\end{figure*}

\newcommand{\til}[1]{\includegraphics[width=28mm]{images/cambridge/labels/#1}}

\begin{figure*}
  \centering
  \begin{tabular}{ccccc}
     \til{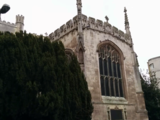} & \til{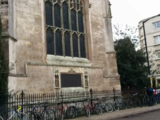} & \til{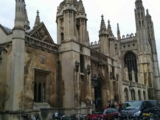} & \til{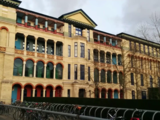} & \til{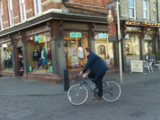}  \\
     \til{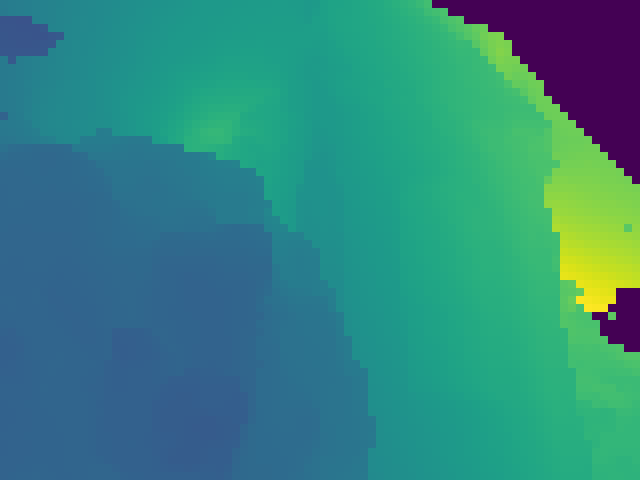} & \til{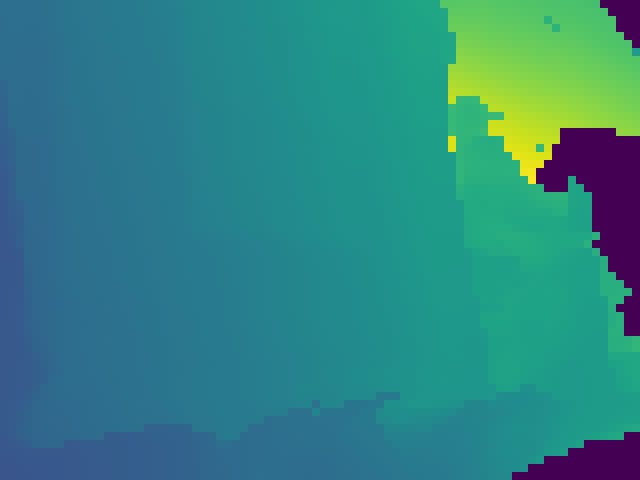}  & \til{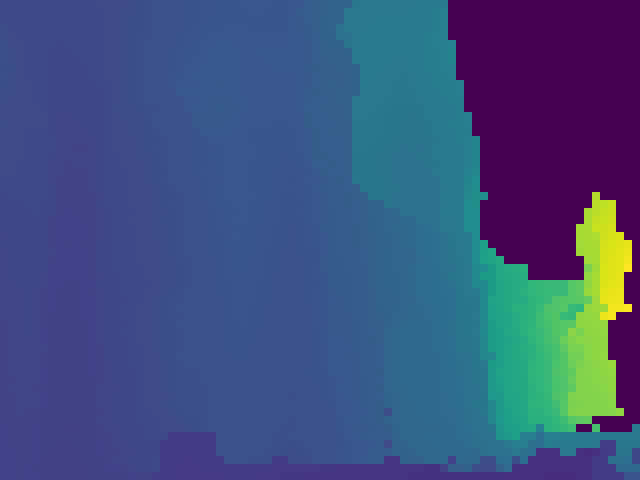} & \til{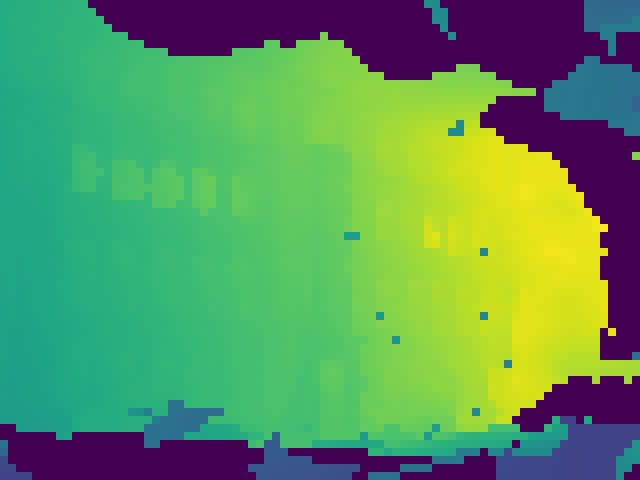} & \til{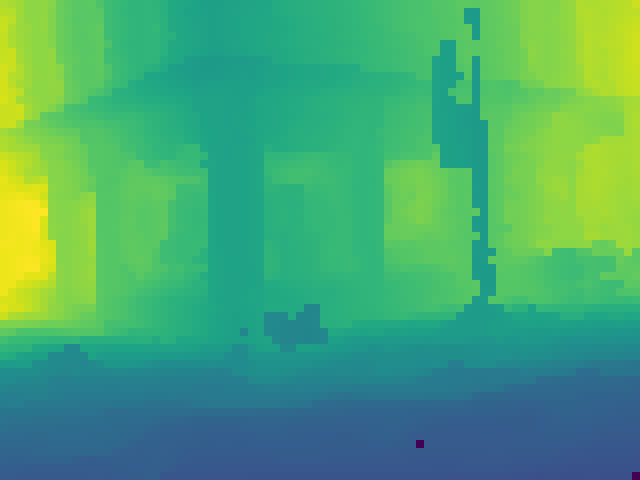}\\

  \end{tabular}
  \caption{Examples of errors in depth labels for Cambridge Landmarks images. Many areas that should have invalid depth are assigned depth values. There are many incorrect depth assignments in general. Dark areas are regions with no depth label (depth=0).}
  \label{fig:labels}
\end{figure*}

\begin{table*}
  \centering
  \ra{1.3}

    \begin{tabular}{llllll}
    \toprule
     & \multicolumn{4}{c}{Sequence} \\
    Method & College & Hospital & Shop & Church & Avg \\ \midrule
    PoseNet~\cite{posenet}                  & 1.92/5.40 & 2.31/5.38 & 1.46/8.08 & 2.65/8.48 & 2.08/6.83\\
    PoseNet Learned Weights~\cite{posenet2} & 0.99/1.06 & 2.17/2.94 & 1.05/3.97 & 1.49/3.43 & 1.43/\textbf{2.85}\\
    Geo PoseNet~\cite{posenet2}             & \textbf{0.88}/\textbf{1.04} & 3.20/3.29 & 0.88/\textbf{3.78} & 1.57/\textbf{3.32} &  1.63/2.86\\
    LSTM PoseNet~\cite{lstmposenet}         & 0.99/3.65 & 1.51/4.29 & 1.18/7.44 & 1.52/6.68 & 1.30/5.51\\
    GPoseNet~\cite{gposenet}                & 1.61/2.29 & 2.62/3.89 & 1.14/5.73 & 2.93/6.46 & 2.08/4.59\\
    SVS-Pose~\cite{svspose}                 & 1.06/2.81 & 1.50/4.03 & \textbf{0.63}/5.73 & 2.11/8.11 & 1.32/5.17\\
    MapNet~\cite{mapnet}                    & 1.07/1.89 & 1.94/3.91 & 1.49/4.22 & 2.00/4.53 & 1.62/3.64\\
    \midrule
    Ours                                    & 1.19/2.16 & \textbf{1.11}/\textbf{1.92} & 0.95/6.82 & \textbf{1.37}/4.45 & \textbf{1.16}/3.84\\
    \bottomrule

    \end{tabular}
  \caption{Direct pose estimation results on Cambridge Landmarks compared to other methods (median position in meters/median rotation in degrees).}
  \label{tab:landmarks}
 
\end{table*}

\newcommand{\incamb}[1]{\includegraphics[width=28mm]{images/cambridge/#1}}

\begin{figure*}
  \centering
  \begin{tabular}{cc|ccc|c}
  &  & \multicolumn{3}{c|}{Network Outputs} & Computed \\
 &    Input &  Depth &  Scene Coords &  Weights & Endpoint Error\\ 
  1&  \incamb{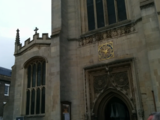} & \incamb{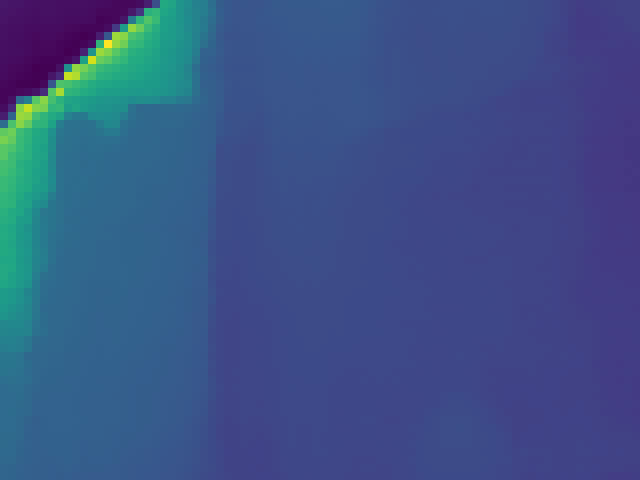} & \incamb{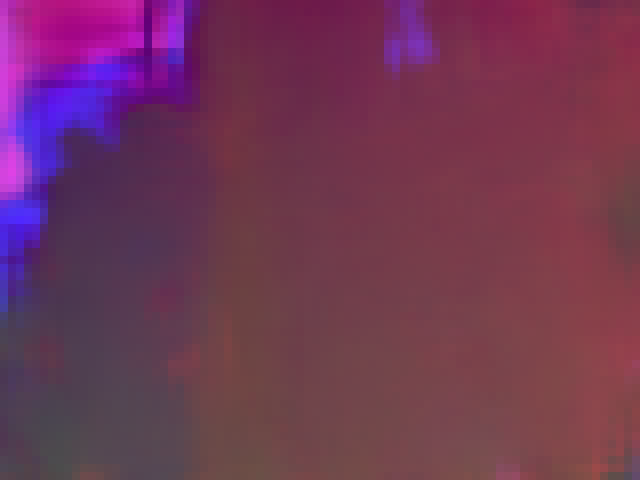} & \incamb{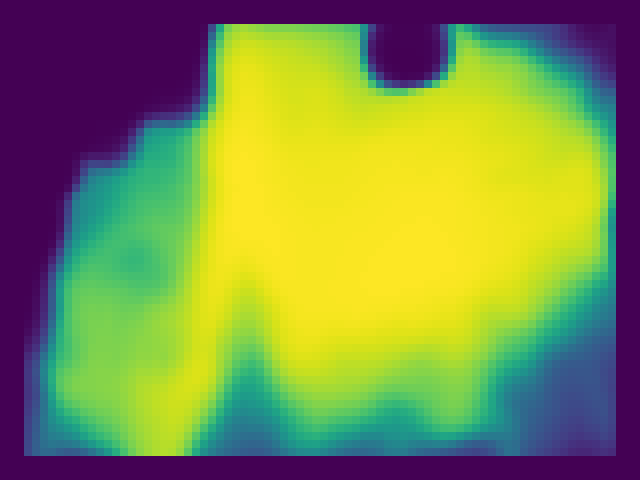} & \incamb{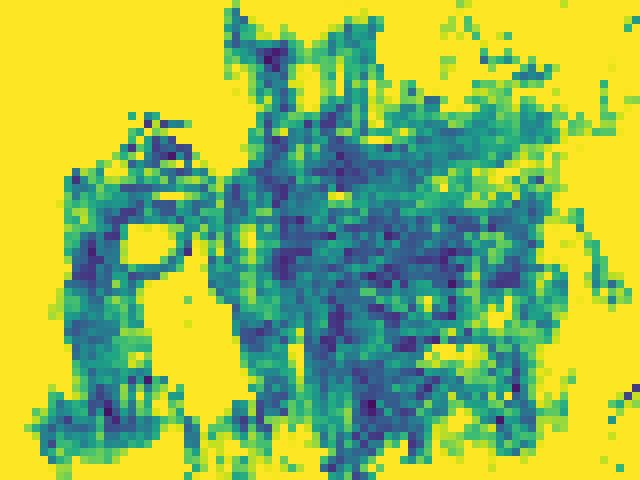}  \\
  2&  \incamb{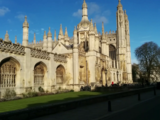} & \incamb{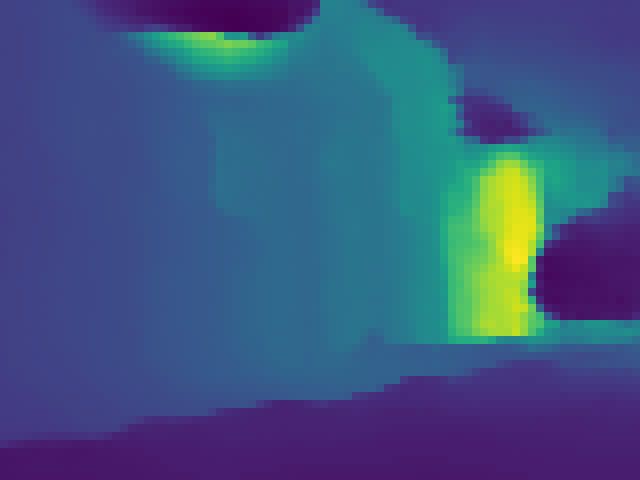} & \incamb{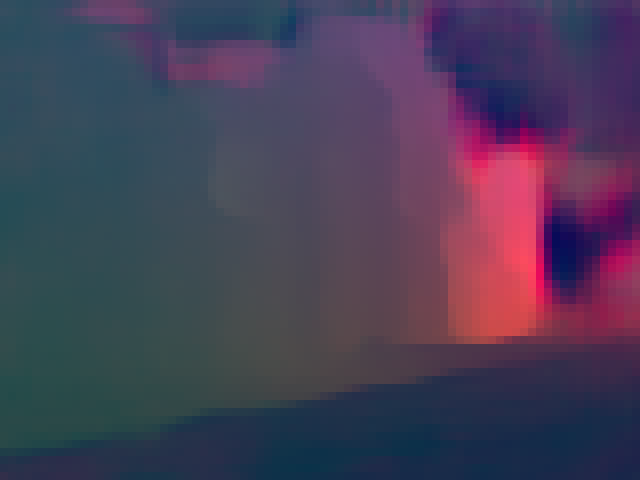} & \incamb{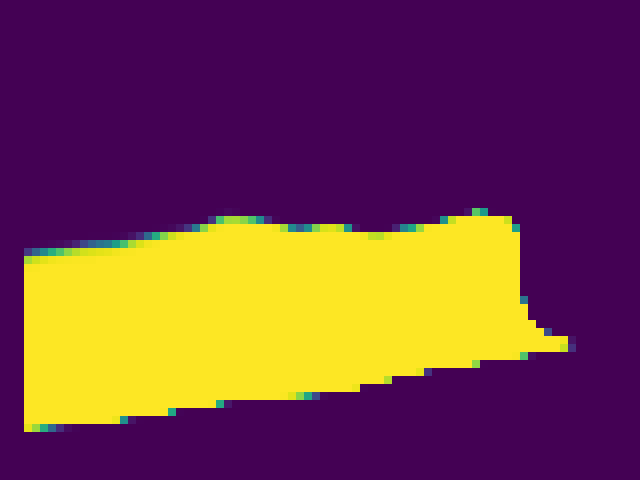} & \incamb{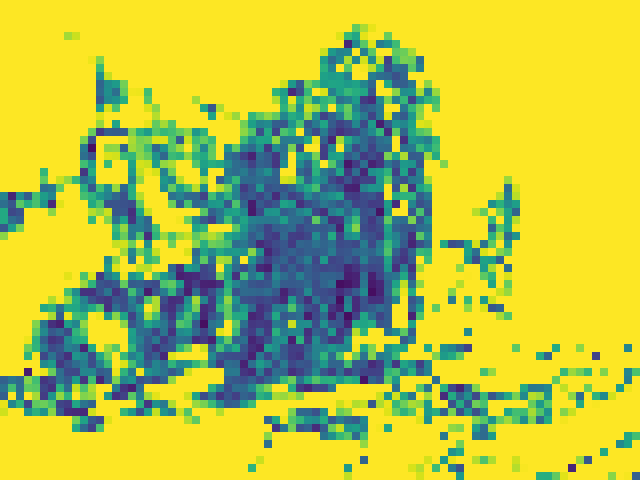}  \\
3&    \incamb{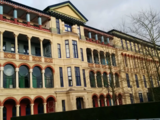} & \incamb{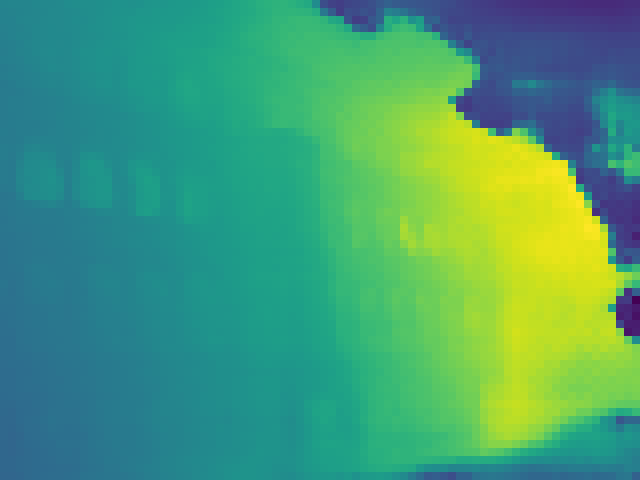} & \incamb{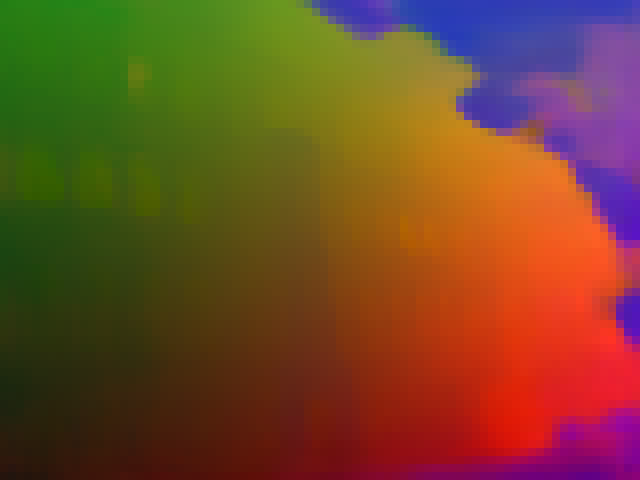} & \incamb{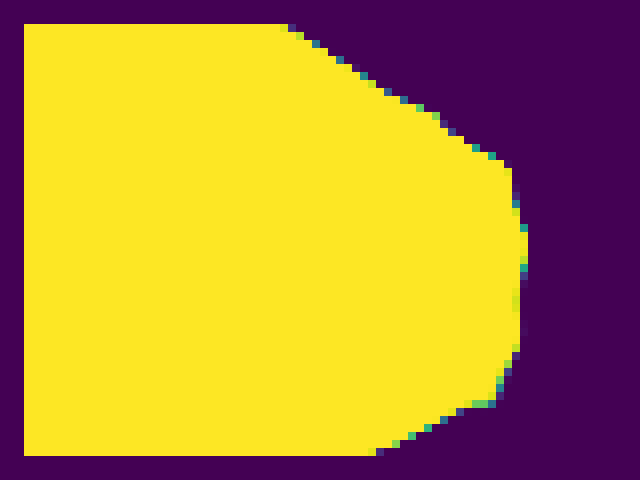} & \incamb{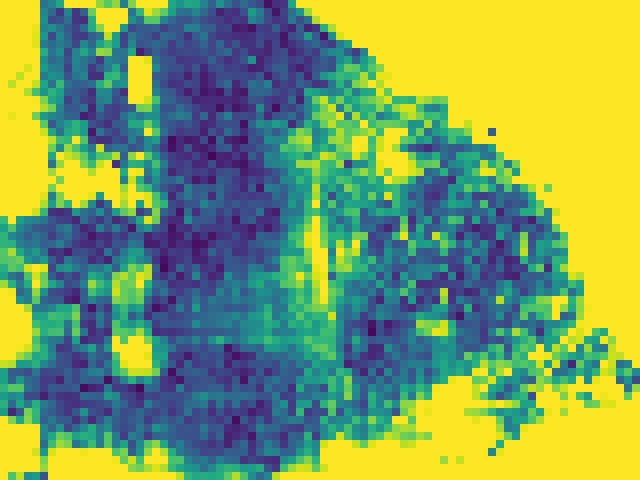}  \\
  4&  \incamb{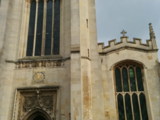} & \incamb{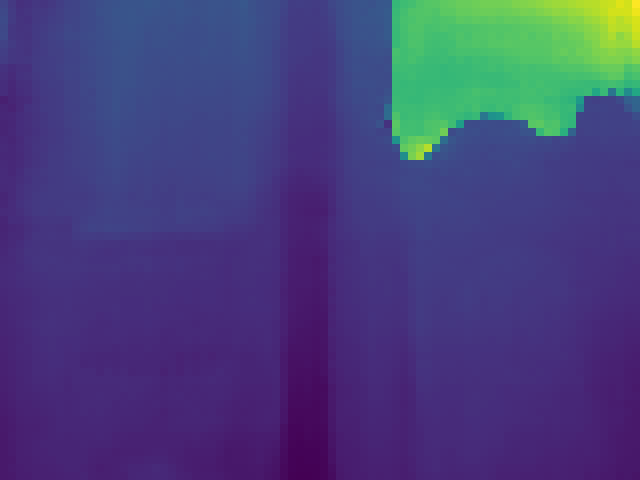} & \incamb{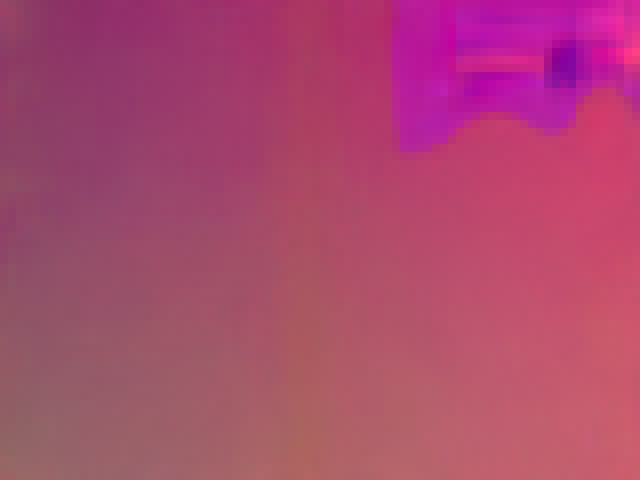} & \incamb{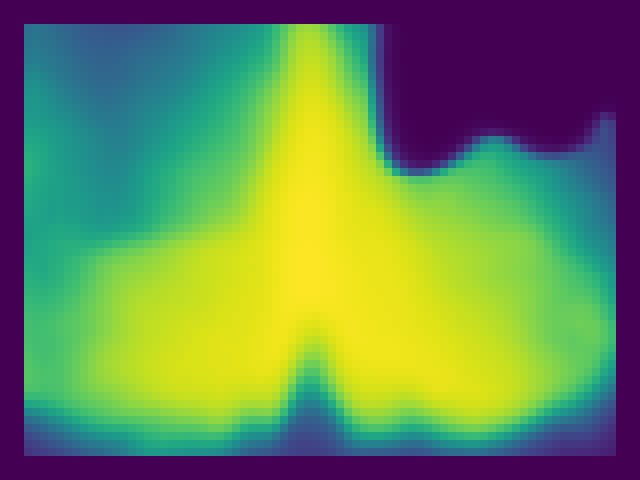} & \incamb{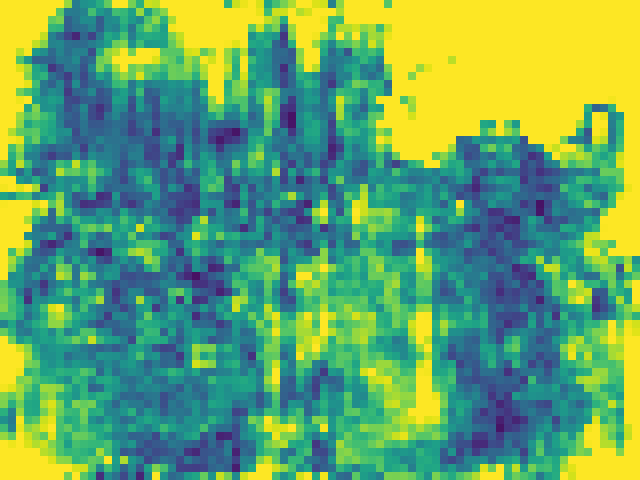}  \\
5&    \incamb{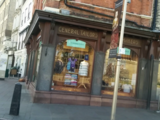} & \incamb{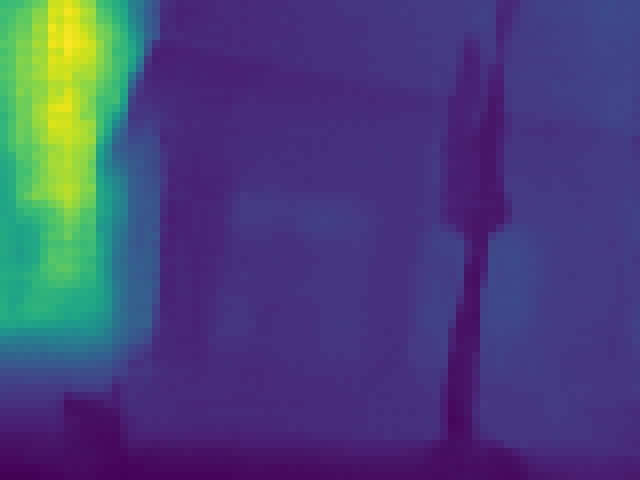} & \incamb{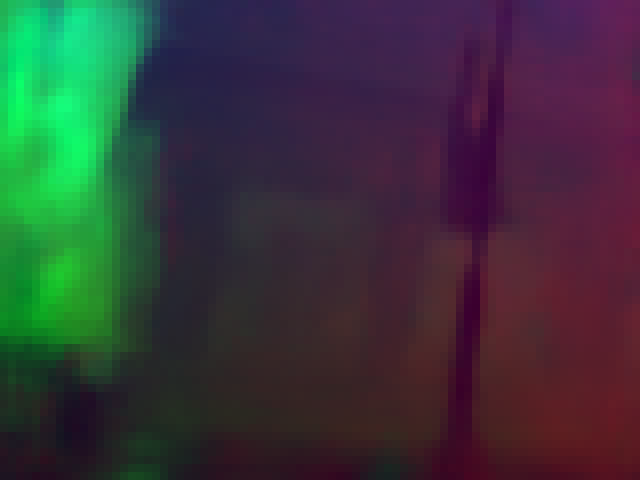} & \incamb{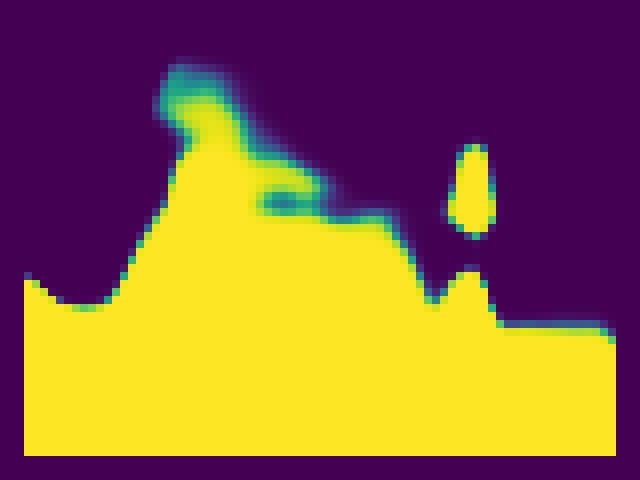} & \incamb{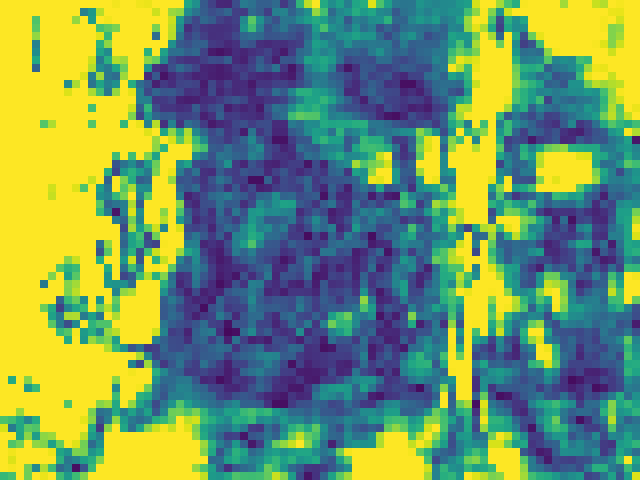}  \\
6&    \incamb{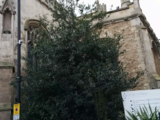} & \incamb{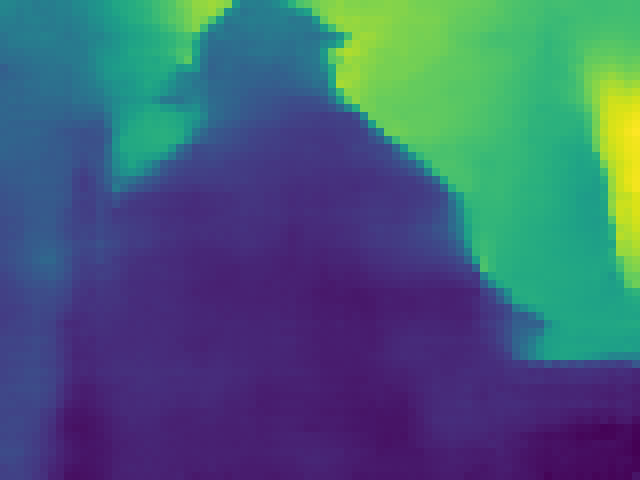} & \incamb{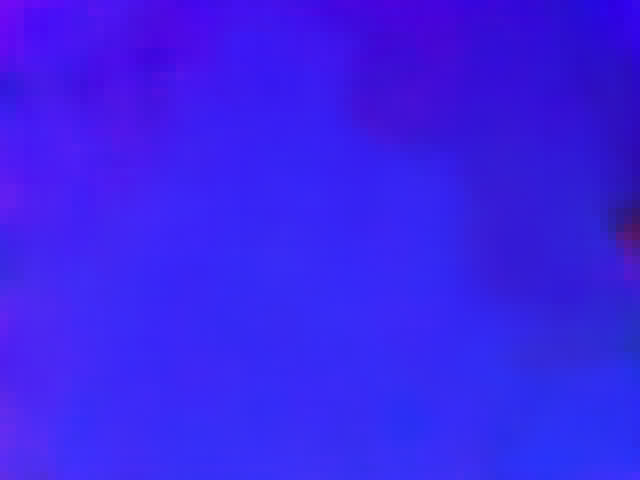} & \incamb{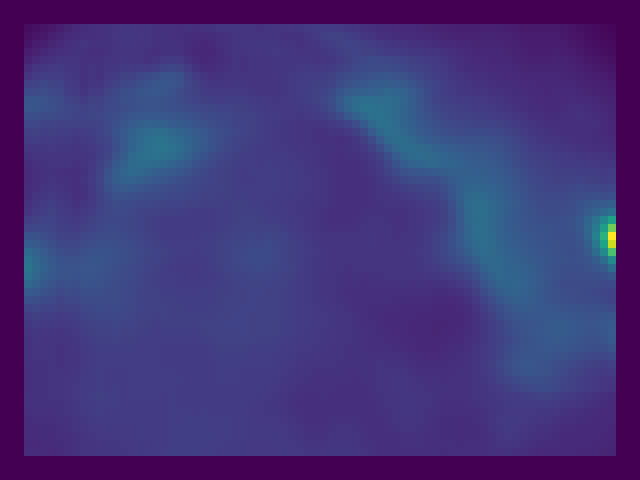} & \incamb{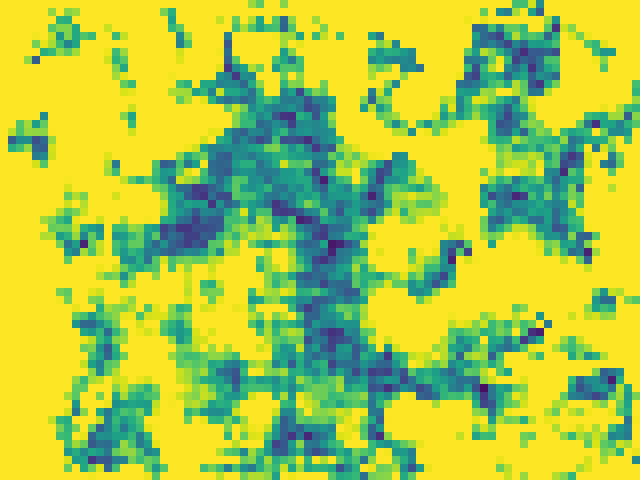}  \\
 7&   \incamb{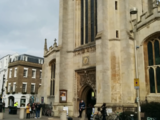} & \incamb{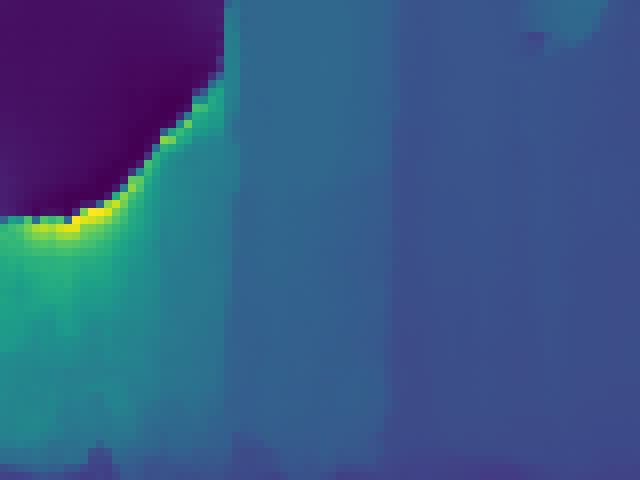} & \incamb{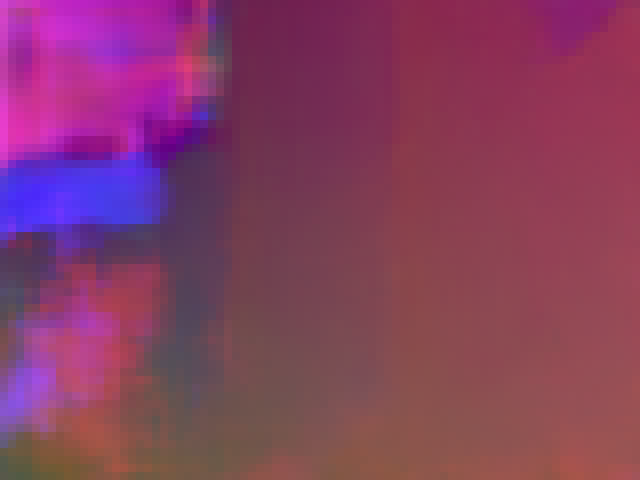} & \incamb{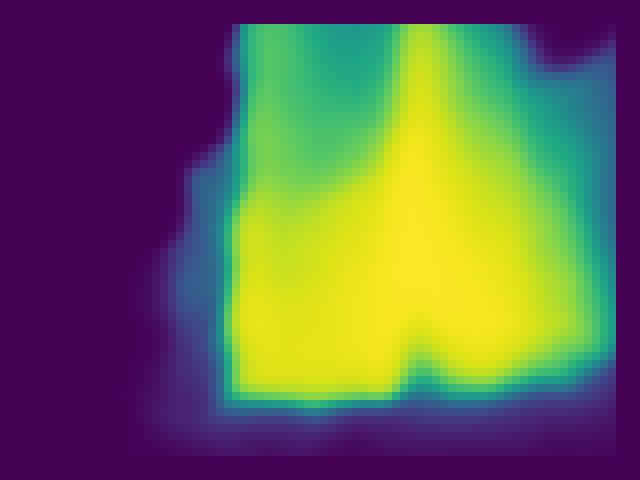} & \incamb{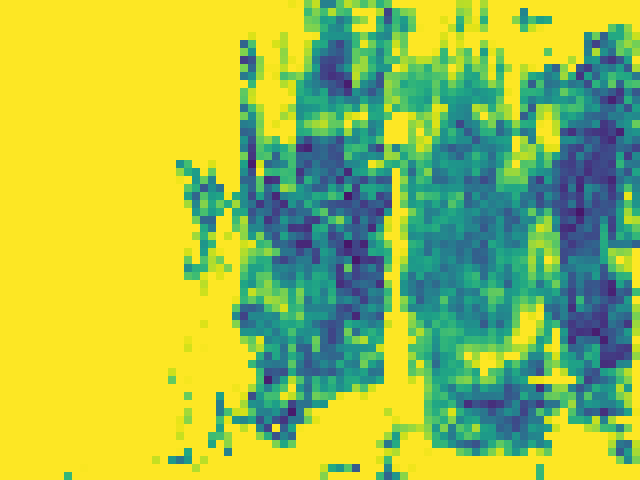}  \\
 8&   \incamb{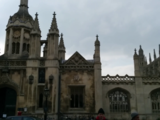} & \incamb{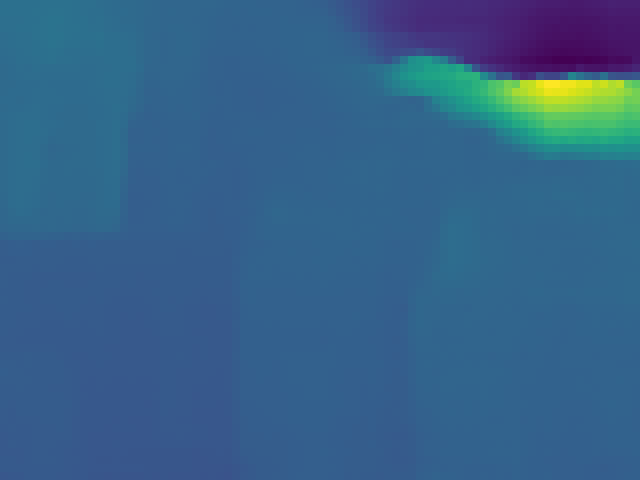} & \incamb{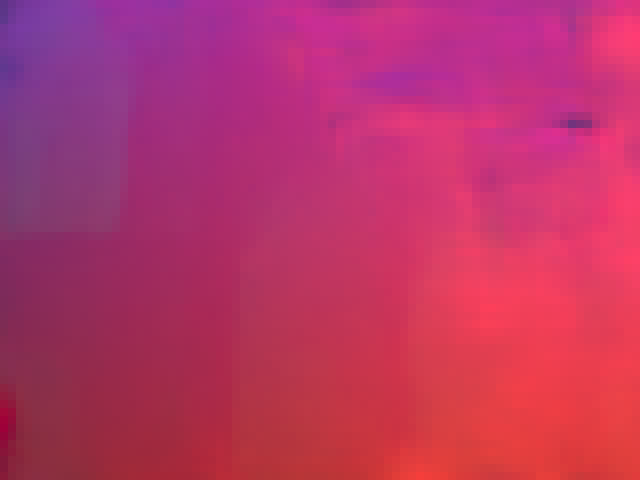} & \incamb{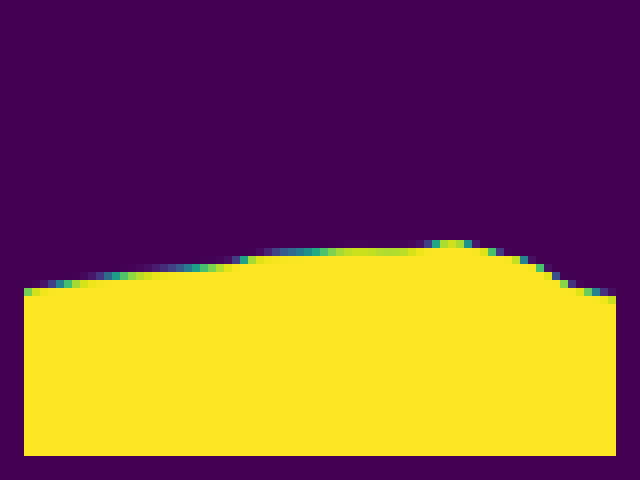} & \incamb{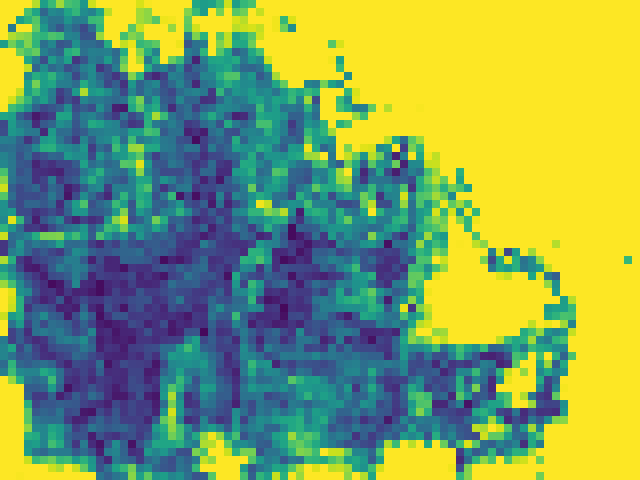}  \\

  \end{tabular}
  \caption{Example network outputs for Cambridge Landmarks images. Depth, scene coordinates, and weights are direct outputs of our network. The endpoint error is computed by applying the regressed pose to the scene coordinates and clamped to 1 for visualization. For depth, weights, and endpoint error, brighter means a higher value. Row labels are shown on the left for referencing in text.}
  \label{fig:examples}
\end{figure*}

\begin{table*}
  \centering
  \ra{1.3}
 % \resizebox{.7\linewidth}{!}{
    \begin{tabular}{llllllll}
    \toprule
     & \multicolumn{7}{c}{Sequence} \\
    Method & Chess & Fire & Heads & Office & Pumpkin & Kitchen & Stairs \\ \midrule
    \textbf{Single Scene} \\
    Abs Error & 0.2318 & 0.1954 & 0.1491 & 0.2367 & 0.2244 & 0.2570 & 0.3515 \\
$\delta < 0.5^3$ & 0.4883 & 0.4743 & 0.6206 & 0.3651 & 0.4164 & 0.3616 & 0.3260 \\
$\delta < 0.5^2$ & 0.7482 & 0.7461 & 0.8371 & 0.6554 & 0.7030 & 0.6349 & 0.5610 \\
$\delta < 0.5$ & 0.8968 & 0.9364 & 0.9385 & 0.9011 & 0.9261 & 0.8870 & 0.7959 \\

    \midrule
    \textbf{All Scene} \\
Abs Error & 0.2271 & 0.1842 & 0.1730 & 0.2325 & 0.2591 & 0.2654 & 0.4374 \\
$\delta < 0.5^3$ & 0.4909 & 0.4945 & 0.5817 & 0.3876 & 0.3599 & 0.3580 & 0.2559 \\
$\delta < 0.5^2$ & 0.7499 & 0.7710 & 0.7842 & 0.6694 & 0.6607 & 0.6280 & 0.4535 \\
$\delta < 0.5$ & 0.8980 & 0.9493 & 0.9216 & 0.9029 & 0.9044 & 0.8787 & 0.7196 \\

     \midrule
     \textbf{Held Out} \\

Abs Error & 0.2768 & 0.2157 & 0.2425 & 0.3333 & 0.3549 & 0.3386 & 0.4800 \\
 $\delta < 0.5^3$ & 0.3759 & 0.4060 & 0.3014 & 0.2212 & 0.1996 & 0.2615 & 0.2316 \\
 $\delta < 0.5^2$ & 0.6372 & 0.6893 & 0.5940 & 0.4526 & 0.4479 & 0.4814 & 0.4173 \\
 $\delta < 0.5$ & 0.8666 & 0.9156 & 0.9119 & 0.8048 & 0.8251 & 0.7910 & 0.6766 \\

    \bottomrule
    
    \end{tabular}
%  }
  
  \caption{Depth error on each scene in three scenarios: training with only on scene, training with all 7 scenes, and holding one scene out during training. Results are reported in meters.}
  \label{tab:depth_error}
 
\end{table*}

\begin{table*}
  \centering
  \ra{1.3}
    \begin{tabular}{l}
    \toprule
     ConvBlock1 \\
     ResBlock1 \\
     ResBlock2\\
     ResBlock3\\
     ResBlock4\\
     Linear Layer \\
    
    \bottomrule

    \end{tabular}
  \caption{ResNet}
  \label{tab:resnet}
 
\end{table*}

\begin{table*}
  \centering
  \ra{1.3}
    \begin{tabular}{lll}
    \toprule
    Name & Operation & Input \\
    \midrule
    conv1 & Conv2d(256,256,1,1,0) & ResBlock2 \\
    act-conv1 & ReLU & conv1 \\       
    conv2 & Conv2d(256,256,1,1,0) & act-conv1 \\
    act-conv2 & ReLU & conv2 \\  
    scene-coords & Conv2d(256,3,1,1,0) & act-conv2 \\
    
    \bottomrule

    \end{tabular}
  \caption{Scene Coordinate Regression CNN}
  \label{tab:scenecoords}
 
\end{table*}

\begin{table*}
  \centering
  \ra{1.3}
    \begin{tabular}{lll}
    \toprule
    Name & Operation & Input \\
    \midrule
    conv1 & Conv2d(512, 128, 1,1,0) & ResBlock4\\
    bn-conv1 & BatchNorm & conv1\\
    act-conv1 & ReLU & bn-conv1 \\
    tp-conv2 & ConvTranspose2d(128, 128, 3,2,1) & act-conv1\\
    bn-tp-conv2 & BatchNorm & tp-conv2\\
    act-tp-conv2 & ReLU & bn-tp-conv2 \\
    conv3 & Conv2d(128, 256, 1,1,0) & act-tp-conv2\\
    bn-conv3 & BatchNorm & conv3\\
    act-conv3 & ReLU & bn-conv3 \\

    conv4 & Conv2d(256, 64, 1,1,0) & ResBlock3+act-conv3\\
    bn-conv4 & BatchNorm & conv4\\
    act-conv4 & ReLU & bn-conv4 \\
    tp-conv5 & ConvTranspose2d(64, 64, 3,2,1) & act-conv4\\
    bn-tp-conv5 & BatchNorm & tp-conv5\\
    act-tp-conv5 & ReLU & bn-tp-conv5 \\
    conv6 & Conv2d(64, 128, 1,1,0) & act-tp-conv5\\
    bn-conv6 & BatchNorm & conv6\\
    act-conv6 & ReLU & bn-conv6 \\

    \midrule
    Half Res Depth Net \\
    half-depth-conv1 & Conv2d(256,256,1,1,0) & ResBlock3+act-conv3 \\
    act-half-depth-conv1 & ReLU & half-depth-conv1 \\       
    half-depth-conv2 & Conv2d(256,256,1,1,0) & act-half-depth-conv1 \\
    act-half-depth-conv2 & ReLU & half-depth-conv2 \\  
    half-depth-conv3 & Conv2d(256,1,1,1,0) & act-half-depth-conv2 \\
    hald-depth & Sigmoid() & half-depth-conv3 \\
    
    \midrule
    Depth Net\\

    depth-conv1 & Conv2d(128,128,1,1,0) & ResBlock2+act-conv6 \\
    act-depth-conv1 & ReLU & depth-conv1 \\       
    depth-conv2 & Conv2d(128,128,1,1,0) & act-depth-conv2 \\
    act-depth-conv2 & ReLU & depth-conv2 \\  
    depth-conv3 & Conv2d(128,1,1,1,0) & act-depth-conv2 \\
    depth & Sigmoid() & depth-conv3 \\
    
    \bottomrule

    \end{tabular}
  \caption{Depth CNN}
  \label{tab:depth}
 
\end{table*}

\begin{table*}
  \centering
  \ra{1.3}
    \begin{tabular}{lll}
    \toprule
    Name & Operation & Input \\
    \midrule
    conv1 & Conv2d(6,64,3,1,0) & cat(Scene Coordinates, Camera Coordinates) \\
    act-conv1 & ReLU & conv1 \\       
    conv2 & Conv2d(64,128,3,1,0) & act-conv1 \\
    act-conv2 & ReLU & conv2 \\  
    conv3 & Conv2d(128,512,3,1,0) & act-conv2 \\
    act-conv3 & ReLU & conv3 \\  
    weights & Conv2d(512,1,1,1,0) & act-conv3 \\
    
    \bottomrule

    \end{tabular}
  \caption{Weighting CNN}
  \label{tab:weighting}
 
\end{table*}

\end{document}